\documentclass[a4paper,UKenglish]{dagman}

\usepackage{microtype} 

\usepackage{comment}

\newcommand{\blue}[1]{\textcolor{blue}{#1}}

\subject{Manifesto from Dagstuhl Perspectives Workshop 22282}
\title{Current and Future Challenges in Knowledge Representation and Reasoning}

\author[1]{James P. Delgrande}
\author[2]{Birte Glimm}
\author[3]{Thomas Meyer}
\author[4]{Miroslaw Truszczynski}
\author[5]{Frank Wolter}
\authorrunning{J.\ P.\ Delgrande, B.\ Glimm, T.\ Meyer, M.\ Truszczynski, and F.\ Wolter}
\affil[1]{Simon Fraser University, CA, \texttt{jim@cs.sfu.ca}}
\affil[2]{Ulm University, DE, \texttt{birte.glimm@uni-ulm.de}}
\affil[3]{University of Cape Town, ZA, \texttt{tmeyer@cair.org.za}}
\affil[4]{University of Kentucky, US, \texttt{mirektruszczynski@gmail.com}}
\affil[5]{University of Liverpool, GB, \texttt{wolter@liverpool.ac.uk}}

\newcommand{\nop}[1]{}

\additionaleditors{
Vaishak Belle, Diego Calvanese, Jens Cla{\ss}en, Giuseppe de Giacomo, Wolfgang Dvo\v{r}\'ak, Anthony Hunter, Gerhard Lakemeyer, Marco Montali, Ana Ozaki, Francesco Ricca, Steven Schokaert,
Francesca Toni, Johannes Wallner and Stefan Woltran}

\subjclass{Computing methodologies $\rightarrow$ Artificial intelligence $\rightarrow$ Knowledge representation and reasoning}


\seminarnumber{22282}
\semdata{11.--15.~July, 2022 -- \href{http://www.dagstuhl.de/22282}{www.dagstuhl.de/22282}}

\volumeinfo
  {James P. Delgrande, Birte Glimm, Thomas Meyer, Miroslaw Truszczynski, and Frank Wolter}
  {5}
  {Current and Future Challenges in Knowledge Representation and Reasoning}
  {1}
  {1}
  {1}

\begin{document}

\maketitle

\begin{abstract}
Knowledge Representation and Reasoning is a central, longstanding, and active
area of Artificial Intelligence.
Over the years it has evolved significantly; more recently it has been
challenged and complemented by research in areas such as machine
learning and reasoning under uncertainty.
In July 2022 a Dagstuhl Perspectives workshop was held on Knowledge
Representation and Reasoning.
The goal of the workshop was to describe the state of the art in the field, including its
relation with other areas, its shortcomings and strengths, together with
recommendations for future progress.
We developed this \emph{manifesto} based on the presentations, panels, working
groups, and discussions that took place at the Dagstuhl Workshop.
It is a declaration of our views on Knowledge
Representation: its origins, goals, milestones, and current foci;
its relation to other disciplines, especially to Artificial Intelligence;
and on its challenges, along with key priorities for the next decade.
\end{abstract}

\section*{Executive Summary}
Knowledge Representation and Reasoning (KR) is a  field of Artificial
Intelligence (AI) that deals with explicit, symbolic, declarative representations of
knowledge along with inference procedures for deriving further, implicit
information from these 
representations.
Even though KR is one of the oldest and best-established areas of AI, it
continues to grow and thrive.
Most of the original research areas have evolved significantly, and have
matured from the discovery and exploration of foundations, to the development
and analysis of systems for emerging or established applications.
Other areas, such as 
answer set programming and argumentation, arose much more recently and are now thriving areas of KR. 

While progress in KR has been steady and often impressive, it has not kept pace
with the recent successes in AI in the use of statistical techniques and
machine learning (ML).
As a result, much of the work in AI, and much of the public perception of AI,
centres today on machine learning and on statistical applications.
Nonetheless, we take it as given that KR is a vital, essential area of AI,
and that research and development in KR remains crucial for the overall
development of AI and for 
its already wide and growing range of applications.
Indeed, despite the unquestionable successes in machine learning and
statistical techniques, limitations of these approaches are now emerging
that, we believe, can only be overcome with advances in KR.
Indicative of this are, for instance, the recent interest in \emph{Explainable AI}, which requires
a reference to declarative structures and reasoning over such structures, as well as the need for enhancing current AI systems with commonsense reasoning capabilities.
Thus, in common with the majority opinion in AI, cognitive science, and philosophy, we 
espouse the position that declarative representations of knowledge are essential for any ultimate, general theory of intelligence.

For all of these reasons, a reassessment of the area of Knowledge
Representation is timely.
The Dagstuhl Perspectives Workshop 22282 ``Current and Future Challenges in Knowledge Representation and Reasoning'' had this as its main objective.
During the workshop, the participants assessed the current state of KR along
with future trends and developments.
Further, the workshop offered ideas for developing an innovative agenda
for the next phase of KR research.
Key proposals targeted supporting a synergistic relationship with other
subareas of the rapidly-changing field of AI, and of computer science as a
whole.
The workshop further identified research areas for emphasis, assessed prospects for practical application of techniques, and considered how KR may address limitations of statistical techniques and machine learning.

This manifesto, while representing the views of the authors and several other researchers who offered input and comments, to a large degree is a reflection of discussions at the Dagstuhl workshop.
%
%
Its first section outlines the field of Knowledge Representation,
briefly characterising the area and explaining why a \emph{knowledge-based}
approach is important to Artificial Intelligence, arguably \emph{necessary} for the
development of general intelligent agents.
The following section reviews areas seen to be in KR, or related to it. Some of these areas fall squarely within KR. Some other have a strong overlap with KR, while for yet others there is a strongly felt need for their overlap with KR to grow. For each of these areas we give a snapshot of their state (as it relates to KR), along with their research issues and challenges. Where appropriate we indicate how KR may contribute to these areas, and how they may in turn contribute to KR.
The third section considers research challenges to the field of KR as a whole:
what these challenges are and how they may be met.
The fourth section discusses ways of promoting KR, including expanding the
field and enhancing its visibility.
Section~\ref{sec:conclusions} provides a brief conclusion.
While the history of KR is interesting and instructive, we feel that a detailed
history would overly lengthen the manifesto;
we have however included a brief history in the appendix.

\tableofcontents

\section{Introduction to Knowledge Representation and Reasoning}
\label{sec:intro}

\subsection{What is Knowledge Representation and Reasoning?}
\label{sec:KRdef}

Knowledge Representation and Reasoning (KR) is a field of Artificial
Intelligence (AI) that deals with explicit, symbolic, declarative representations of
information along with inference procedures for deriving further, implicit
information from these representations.
Typically, these symbolic representations encode information on some domain of
application.
They may also encode other forms of information. For instance, when describing knowledge of an agent, they may encode that agent's
goals and preferences, beliefs about other agents, etc.
This collection of symbolic expressions is called a
\emph{knowledge base (KB)}.
The key idea is that the information in a KB describes the domain, without specifying how information is to be used. Accompanying a KB is a \emph{knowledge base manager} consisting of a collection
of procedures that perform inferences on the knowledge base, possibly
interacting with the external world via sensors and actuators, and possibly
modifying the knowledge base. Together, the KB and the KB manager form a \emph{knowledge-based system}. 
This general paradigm of building systems based on knowledge is also referred to as the \emph{declarative} approach to system building, in contrast with a \emph{procedural} paradigm, in
which a system consists of a collection of specific procedures or commands that determine its scope of functionality (as in an operating system, for example).

\subsection{Why Knowledge Representation and Reasoning?}

The case for declarative representations was made early on in the history of AI, motivated by analogies with how humans address a wide range of problems they encounter in everyday life. A similar case was also made in other areas, including  databases, information systems, and logic programming. The arguments put forward in those areas provide a strong support for declarative approaches to knowledge representation. We review some of them below.

First, a symbolic framework allows for a principled, formal account to both representation and reasoning.
The meaning of sentences in a KB can be defined via a semantic account, while
specialised inference procedures can be developed for the given semantics, typically so that they are 
\emph{sound} and \emph{complete}, the characteristics that establish their formal adequacy for reasoning.\footnote{A \emph{sound} procedure is one that will derive true consequences from true
premises, while a \emph{complete} procedure is one that can derive all true
consequences of the premises.}

Second, in the declarative paradigm, the information in a KB represents \emph{what} an agent knows about
a domain but does not constrain \emph{how} to reason with this information: what types of reasoning tasks one might want to perform and how to execute them. For example, in a simple KB system in the form of a relational database, we have a declarative specification of the knowledge given by the set of ground atoms in the language determined by the relational schema. This representation leaves open what reasoning tasks (queries) the user might want to execute, and how they are actually executed once posed to the system.

An important point about this separation of information from reasoning tasks and their implementation, is that a KB can easily be updated (new information added, information no longer relevant removed, and erroneous information corrected), without requiring any changes to inference procedures.
For example, in a planning domain, information about a particular planning problem (scenario) can be modified, independently of how a plan is constructed.

Next, a knowledge-based system is, in principle, able to explain and justify its
behaviour, based on its KB and the inference steps it took to arrive at a conclusion.
For example a KB may indicate that a specific drug was not prescribed for a
patient, even though the drug is effective against the patient's ailment,
because the patient is allergic to the class of drugs to which the proposed
drug belongs.

Another important consideration concerns the type of information one may want to include in a knowledge-based system. Often it is factual information pertaining to a particular application domain. However, a KB may include \emph{any} additional pertinent information.
This can include the agent's (user's) knowledge, goals, obligations, and preferences, or that of other agents. This can include hypothetical or counterfactual information, and assertions regarding the past or future, and also information about actions available to the agent (or agents).


At the same time, KB systems pose numerous challenges (in other words, there is no free lunch). First, both the representation and reasoning problems in a KB may be extremely challenging. A notable example is in commonsense reasoning, where simple assertions such as
``birds normally fly'' but ``penguins normally do not fly''  have been the impetus for much research, but where arguably we still do not know how to fully represent and reason with such assertions.

Second, reasoning required by knowledge-based approaches is often computationally inherently hard. This stands in contrast with procedural representations which, when available, are generally more efficient (for example, to find a shortest path, an agent is better off using a specialised shortest-path algorithm than inference based on the declarative specification of the concept of a shortest path).

Next, as a matter of fact, information humans have about the world is imperfect. The same is true about the information stored in knowledge-based systems. Often this information is incomplete, imprecise, inconsistent, inaccurate, or otherwise incorrect. Thus, a major challenge is to design inference procedures to handle such information.
But in many cases, an agreement has not yet emerged on the appropriate choice of such inference mechanisms. And even in cases where there is agreement, reasoning tasks may be undecidable and, when decidable, often are intractable.

Given this, the desiderata of soundness and completeness for reasoning procedures can be
difficult or impossible to meet in practice.
Hence in the interest of efficient reasoning, a practical inference procedure
might be unsound or incomplete in general, or a representation language may be limited in expressibility (for example, in the case of first-order logic languages, to Horn clauses), in return for guaranteed bounds on inference.


A separate issue concerns the matter of acquiring relevant knowledge, and effectively structuring it
in a KB. In some cases (for example, description logics) substantial progress has been
made, but in others the complexity of the tasks remain a hindrance to the
development of practical applications.

The challenge then is to create practical knowledge-based systems for managing
complex, real-world tasks.
The fundamental problems of KR may be considered in two categories:
first, the development of sufficiently precise notations with which to
represent knowledge and, second, the development of effective procedures for
deriving further knowledge.
These have been referred to as, respectively, the \emph{epistemological} and \emph{heuristic} adequacy~\cite{McCarthy77} of a knowledge-based system.

We conclude this section by looking at what is often taken as the ultimate goal of AI, that is,
developing a general artificially-intelligent agent.
To this end, there is a stance, known as the \emph{KR Hypothesis}, that asserts
that a knowledge-based approach is \emph{essential} for the construction of
any generally intelligent agent.
This was articulated by Brian Smith~\cite{Smith82} as follows:
\begin{quote}
``Any mechanically embodied intelligent process will be comprised of structural
ingredients that
\begin{enumerate}
\item
we as external observers naturally take to represent a propositional account
of the knowledge that the overall process exhibits, and
\item
independent of such external semantical attribution, play a formal but causal
and essential role in engendering the behaviour that manifests that knowledge.''
\end{enumerate}
\end{quote}
That is, the KR Hypothesis claims that any intelligent agent will contain
symbolic structures (commonly referred to as knowledge base) where the symbols in the KB
\emph{mean} or \emph{represent} something.
Moreover, the manipulation of these symbols will play an essential role in
determining the behaviour of the system. This \emph{stance} (or \emph{thesis}) directly points to the key role of knowledge representation and reasoning for achieving general intelligence.
It is worthwhile to note, though, that some researchers, notably in the connectionist community, disagree. Thus, at present,
whether general intelligence is necessarily knowledge-based remains an open
question.

\section{Areas In and Related to Knowledge Representation and Reasoning}
\label{sec:areas}

KR is a broad area aiming to  establish foundations and general principles for representing knowledge in ways that make that knowledge usable by computers, and to develop effective automated reasoning mechanisms. It can be broken down into several subareas, some of a broad foundational nature, some concerned with specific formalisms and tools, and some focused on applications. Further, as one of the main areas in AI, KR is also related to several important subareas of AI. Finally, because of its significant use of logic and its focus on the concept of knowledge, KR is also related to mathematical and philosophical logic and, more generally, to philosophy. 

In this section we offer an overview of KR and its main research efforts, both past and current. 
These areas are presented in two rough groups.
First are those that fall squarely within KR, such as nonmonotonic reasoning.
Second are those areas that are distinct from KR but have  strong overlap;
these subsections have titles of the form ``KR and \dots''.

\subsection{Non-monotonic Reasoning}

Non-monotonic reasoning (NMR)~\cite{BrewkaEtAl08} is one of the original areas of KR and a driving force behind its early development. It remains a major area of KR today. Unlike other areas of KR, such as reasoning about action (Section~\ref{sec:areas:ReasoningAboutActionAndPlanning}), description logics and ontologies (Section~\ref{sec:areas:DLsOntologies}), or argumentation (Section~\ref{sec:areas:Argumentation}), NMR is not focused on any particular knowledge representation language or any particular application domain. Rather, it concerns a major aspect of \emph{commonsense reasoning}, that of reasoning (and acting) based on incomplete information by ``filling in the gaps'' of this missing information in some fashion. In such cases, when additional information becomes available, it is natural that some conclusions that have been reached earlier must be withdrawn. 

This type of reasoning cannot be modelled in concise, natural ways in classical inference systems such as first-order logic.\footnote{However, by expanding the underlying formal system, for instance to allow time-stamping of each formula or moving to a second-order logic, we can capture some non-monotonic behaviour within classical systems.} Classical inference systems are designed to derive new information only when the reasoning that leads to them is iron-clad and cannot be invalidated when new facts become known. More formally, classical reasoning is \emph{monotone}, that is, a conclusion derived from a set of facts remains a conclusion when this set of facts is enlarged. \emph{Defeasibility} of human commonsense reasoning when complete information is not available means that new information may render invalid some of the conclusions obtained in its absence. Therefore, commonsense reasoning requires a new type of logic as its natural formal expression, a logic whose inference mechanism lacks the monotonicity property that classical inference systems have. Several such logics, commonly referred to as \emph{non-monotonic logics}, were proposed and studied for their capacity to account for ways in which one may draw \emph{non-monotonic} (or defeasible) conclusions from incomplete or imprecise information.

The \emph{closed world assumption}~\cite{Reiter78} expressed the notion that a ground atomic fact could be assumed to be false if it could not be demonstrated to be true. (This, of course, is how negation is handled in standard relational databases.) \emph{Default logic}~\cite{Reiter80} generalised the intuition behind the closed world assumption. It augmented classical logic theories (\emph{base} theories) by a set of rules, called \emph{defaults} whose applicability depended on a consistency condition. 
\emph{Answer set programming} (Section~\ref{sec:areas.ASP}) in its most basic form can be seen as a fragment of default logic, with its semantics directly traceable to that of default \emph{extensions}~\cite{BidoitF87,MarekT89,BidoitF91}. \emph{Circumscription}~\cite{McCarthy80}, another major attempt at formalising reasoning with incomplete information, expressed the notion that when faced with such situations, humans make the extension of incompletely specified predicates as small as possible. (For example, a circumscriptive interpretation of ``birds fly'' would have the set of non-flying birds be as small as consistently possible.)
Yet another line of research of non-monotonic inferences was set in the language of modal logics. %
Two prominent examples of such non-monotonic logics 
are the modal logics S4F~\cite{SchwarzT1994} and KD45~\cite{Moore1984}. The former has been shown to be equivalent to default logic. The latter, broadly known as \emph{autoepistemic logic}~\cite{Moore1985}, formalised reasoning of an introspective agent based on what she knew and what she did not know.
Over the years, a major research effort was extended to establish fundamental properties of non-monotonic logics and their interrelations~\cite{MarekT1993}, and to understand the computational complexity of reasoning with such logics~\cite{Gottlob1992,EiterG1993}.

A different attempt to provide a direction for non-monotonic formalisms was provided by Kraus, Lehmann, and Magidor~\cite{KrausLehmannMagidor90}.
In that work, the authors formulated axioms against which any non-monotonic consequence relation could be tested, thus establishing the so called \emph{KLM approach} to non-monotonic reasoning. Seeking semantic accounts of non-monotonic inference relations satisfying selected sets of \emph{KLM postulates}, Kraus, Lehmann and Magidor proposed the so called \emph{preferential semantics} and developed fundamental characterisation results. Although that work focused on non-monotonic consequence for classical propositional logic, it was subsequently extended to study non-monotonic consequence over conditional propositional logics \cite{LehmannMagidor1992}. This extension has given rise to a flurry of activity in designing non-monotonic reasoning systems for versions of conditional logics going beyond propositional logic, especially description logics~\cite{GiordanoEtAl2015,Bonatti2019,BritzEtAL2020} and other fragments of first-order logic. 

Over the years NMR has been taken to include, or apply to, commonsense reasoning, answer-set programming, reasoning about action, query-answering in database systems, description logics, argumentation, and belief change. Indeed following the early development of domain-independent approaches, most work on non-monotonic reasoning has focused on addressing non-monotonic reasoning aspects arising in the areas just mentioned.
Current research in NMR may be seen as further mapping out the phenomenon of NMR, investigating its applications and relations to other areas of AI and information systems, and developing further applications.

While research on non-monotonic forms of reasoning contributed substantially to progress in commonsense reasoning, the key question of how to ensure that the correct inferences are drawn still remains a challenge~\cite{Delgrande11}.
At the centre of this challenge is the question of the link between NMR and cognitive science with its two facets: potential influence of NMR research to topics in cognition, and conversely, the extent to which work in cognitive science can benefit NMR, whether by providing benchmarks or phenomena for modelling~\cite{SpiegelEtAl2019,RagniEtAl2020}.

Another major area where NMR can make significant contributions is machine learning (Sections~\ref{sec:areas:ML} and~\ref{sec:challenges.ML}). It is broadly recognised that a major limitation of machine learning in applications is that they often require, but currently lack, common sense and an ability to adjust to unanticipated or novel situations. As noted above, non-monotonic reasoning seems to be a key tool to formalise and implement such abilities. However, this cuts both ways. A successful KR system will require large amounts of real-world information, including defeasible inference rules such as defaults. There is a growing evidence that machine learning may be a primary tool for acquiring these rules.

For further information of current issues in NMR we refer to the materials of a Dagstuhl Perspectives workshop in 2019 dedicated to non-monotonic reasoning~\cite{HunterEtAl19}.

\subsection{Answer Set Programming}
\label{sec:areas.ASP}

The \emph{stable model semantics} that provided the theoretical underpinning to Answer Set Programming (ASP) was introduced in the late 1980s~\cite{GelfondL1988}. The view of programs as models of search problems was verbalised about ten years later in the papers by Marek and Truszczynski~\cite{MarekT1999} and Niemel\"a~\cite{Niemela1999}. 
Supported by the first computationally-promising program-processing tools \emph{lparse} and \emph{smodels}~\cite{NiemelaS1997,SyrjanenN2001} it quickly attracted the interest of researchers dealing with computational solutions to problems in knowledge representation and constraint satisfaction. In the twenty years since its inception, ASP has grown into one of the most vibrant areas of research in KR. A good introduction and an overview of the field can be found in the survey paper by Brewka et al.~\cite{BrewkaET2011} and in the issue of the AI Magazine dedicated to ASP~\cite{BrewkaET2016}.

ASP continues to attract researchers. On the one hand, the theoretical foundations of ASP abound in challenging problems motivated by its intended use as a tool for modelling knowledge. On the other hand, the demonstrated power of ASP in addressing problems from a broad spectrum of application domains draws those interested in tool development as well as those who are interested in applying the tools. Below we enumerate some of the current research directions that we find particularly interesting or important.

A thorough understanding of why ASP has proved to be so effective still eludes us. Recent work by Hecher, comparing ASP restricted to normal logic programs with propositional logic, provides some insights~\cite{Hecher2022}. That work and several other related papers bring the concept of tree width to the forefront of studies of the expressive power, computational complexity and algorithms for computing answer sets, promising a more nuanced understanding of the inherent computational properties of the stable model semantics. A recent paper by Fandinno and Hecher is a good representative of this line of research~\cite{FandinnoH2021}. 

Another important line of research concerns extensions of the language of ASP with aggregates. Such extensions are necessary to facilitate both knowledge modelling and program processing. But they invariably bring with them the challenge of defining the ``right'' semantics. A good overview of this work can be found in the survey paper by Alviano et al.~\cite{AlvianoFG2021}. Another perspective on that matter was proposed by Vanbesien et al.~\cite{VanbesienBD2022}. Finally, extensions of the language by more general classes of constraints by integrating ASP and constraint programming are also being actively researched~\cite{Lierler2021}. 


An important emerging theme in ASP is to develop methods for counting answer sets of programs. Recent papers by Fichte et al.~\cite{FichteHN2022}, Kabir et al.~\cite{KabirESHFM2022}, and Pajunen and Janhunen~\cite{PajunenJ2021} are good examples of this line of work. The work is motivated by possible applications of ASP in settings requiring probabilistic reasoning, a theme that is increasingly attracting attention given the current emphasis in AI on probabilistic and neuro-symbolic approaches~\cite{LeeY2017,YangIL2020}.

The success of ASP stems to a large degree from the availability of excellent computational tools. Most notable here is the suite of tools developed under the \emph{Potassco} project~\cite{potassco}. This project is ongoing with the effort focused on improvements to its flagship programs \emph{gringo}, \emph{clasp} and \emph{clingo}. However, increasingly, the developers address requirements of emerging extensions of ASP: \emph{plingo} for probabilistic reasoning, \emph{aspirin} for reasoning with qualitative preferences, \emph{telingo} for temporal reasoning, and \emph{plasp} for planning. The main challenge of this line of work lies in developing fast programs for model generation, or ASP solvers. Successful research efforts, besides the \emph{Potassco} project, include \emph{wasp}~\cite{wasp} and \emph{dlv}~\cite{dlv}. 

In some applications, a significant portion of the overall processing time is spent in grounding. Recent efforts to improve grounding tools involve the development of grounders that aim at minimising the size of the ground program~\cite{CalimeriFPZ16}, and lazy grounders~\cite{WeinzierlTF2020} that delay grounding of some parts of the program until those parts are needed for solving to proceed. Related efforts are aimed at establishing formal foundations of grounding~\cite{HarrisonLY2014,KaminskiS2021}. 

A crucial aspect of making ASP easier to use and more effective is to develop tools for program development and verification. The tutorial paper by Kaminski et al.~\cite{KaminskiRSW2020} offers an overview of the contributions of the \emph{Potassco} project in this area. Other recent notable work on tools for ASP include the development of IDEs~\cite{FebbraroRR2011,BusoniuOPST2013}, debuggers~\cite{DodaroGRRS2019} and visualisation software~\cite{HahnSSS2022}. These and other related projects are discussed in a survey paper by Lierler et al.~\cite{LierlerMR2016}.

A major role in the recent advances in ASP program development and processing tools has been played by the language standard ASP-Core-2 adopted by the community \cite{CalimeriFGIKKLM2020}. This effort must be continued in view of the recent expansions of ASP discussed above.

Thanks primarily to the availability of high quality grounders and solvers, ASP has proved to be an effective technology in addressing problems arising in a wide range of application domains. This continues to be a vibrant area of research. Papers by Erdem et al.~\cite{ErdemGL2016} and Falkner et al.~\cite{FalknerFSTT2018} provide good overviews of applications of ASP in solving problems of practical interest. Recent interesting examples of applications of ASP concern train scheduling~\cite{AbelsJOSTW2021}, course scheduling~\cite{BanbaraIKOSSTW2019}, the stable roommates problem~\cite{ErdemFMP2020}, and robotics~\cite{ErdemFMP2020,RizwanPE2020}.

\subsection{Belief Revision}\label{Sec:BR}

\emph{Belief revision} began as an area of philosophy with the seminal
work of Alchourr\'{o}n, G{\"{a}}rdenfors, and Makinson~\cite{AGM1985,Peppas08}
called, after its founders, the \emph{AGM approach}. There are two facets to this approach, somewhat analogous to the proof theory and semantics of a classical logic.
On the one hand, there is a set of belief change postulates that arguably any intelligent agent should follow in changing its set of beliefs. On the other hand, there are various formal constructions for specifying a belief change operator;
these include orderings over formulas in a language, called an \emph{epistemic entrenchment ordering}, or orderings over possible worlds.
The belief change postulates and belief operator constructions are related via a \emph{representation theorem}, showing that an operator that satisfies the postulates can be specified via a formal construction, and vice versa.
These approaches guided the design of specific operators for implementing belief change.
The primary operators addressed were belief \emph{revision} in which an agent consistently incorporates a new belief into its belief corpus, and belief \emph{contraction} in which an agent loses belief in a formula without necessarily believing its negation.
Given that belief revision deals broadly with how an agent may change its beliefs in the face of new information, it is not surprising that belief revision attracted attention in the KR community. Today, it is the KR community that is responsible for most research in the area, including a comprehensive 
exploration of the original AGM framework, and proposed modifications and extensions to other forms of belief change such as belief \emph{merging}~\cite{Konieczny2000,KoniecznyPinoPerez02}. 

In a related development, researchers investigated interesting connections between belief revision and non-monotonic reasoning. Firstly, G{\"a}rdenfors and Makinson~\cite{GardenforsMakinson1994} showed that belief revision and the KLM style of non-monotonic reasoning \cite{KrausLehmannMagidor90} are in a certain formal way so similar that they can be viewed as two sides of the same coin. More recently, Casini et al.~\cite{DBLP:conf/kr/CasiniMV20} proposed a framework for belief revision and belief contraction within a non-monotonic formalism. That the latter even makes sense was somewhat surprising, since one of the basic tenets in belief change used to be that it needs to be based on a monotonic logic. 

The original AGM approach had nothing to say about how an agent's \emph{epistemic state} changes following a belief change operation, and it considered the result of a single change operation only.
This issue was addressed by Darwiche and Pearl~\cite{DarwichePearl1997} who considered the problem of \emph{iterated} revision by developing a qualitative counterpart to Spohn's~\cite{Spohn88} \emph{ordinal conditional functions}.
In their approach, an agent's epistemic state was represented by a total preorder over possible worlds (with the agent's beliefs characterised by the least set of worlds in the preorder).
Then, belief revision could be expressed in terms of the modification of the total preorder.
Jin and Thielscher~\cite{JinThielscher07}, Booth and Meyer~\cite{DBLP:journals/jair/BoothM06}, as well as Nayak et al.~\cite{NayakPagnuccoPeppas03} subsequently expanded this line of research.
In an orthogonal direction, the AGM approach to revision has also been shown to be applicable in any logic~\cite{DelgrandePeppasWoltran18}, including those weaker than classical propositional logic~\cite{DBLP:journals/jair/BoothMVW11,DelgrandeWassermann13,Delgrandepeppas15}.
As a result, the original AGM approach has been substantially broadened in recent years.

Nonetheless, the area suffers from a number of limitations.
Currently, the dominant framework by far is the AGM approach, so much so that on occasion approaches that violate one or another of the AGM principles are seen as being problematic or not \emph{really} about belief change.
This is in contrast with non-monotonic reasoning (NMR), where a wide range of approaches, based on differing underlying intuitions or motivation, have been developed.
While there are alternative approaches, such as based on distance between models or on syntactic considerations, these approaches have attracted little attention, and it remains an open question whether there are alternative, compelling accounts of belief change to complement the AGM approach.

Perhaps the most significant problem with the area, at least from a computer science perspective, is that research has been almost entirely theoretical in nature.
The area largely lacks compelling motivating examples, such as those that drove early work in NMR, and it lacks benchmark problems or domains.
While there have been some prototypical implementations, these implementations are usually limited, do not scale well, and again lack a compelling application.
Thus, arguably, the major challenge facing the area is to develop useful applications and implementations.
Such applications would of course be valuable in their own right.
Further, they would provide a driver toward the development of new formalisms, or the elaboration of existing ones.
To this end, work on belief change in description logics~\cite{DBLP:conf/aaai/MeyerLB05,DBLP:conf/dlog/LeeMPB06,ZhuangWangWangQi16}, or in reasoning about action~\cite{ShapiroEtAl11,ClassenDelgrande21} show promise in embedding belief change in potentially practical areas. In a similar vein, it is worth pointing out that the subareas in description logic research  known as \emph{axiom pinpointing}~\cite{DBLP:series/ssw/Penaloza20} and \emph{ontology repair}~\cite{DBLP:conf/kr/BaaderKNP18} are closely related to belief change. Roughly speaking, axiom pinpointing is concerned with identifying the statements needed to draw a conclusion, and can be used as the basis for performing belief contraction. Similarly, ontology repair has as its aim the removal, or weakening of description logic axioms in order to regain consistency, and can therefore be viewed as a form of belief revision.   

Lastly, belief change can be seen as a qualitative means of dealing with uncertainty and the acquisition of new knowledge;
consequently it would be very interesting to consider belief change in the context of current work on uncertainty or in the use of machine learning techniques toward the learning of contingent, qualitative information.

\subsection{Reasoning about Action and Planning}
\label{sec:areas:ReasoningAboutActionAndPlanning}

Reasoning about action and change is one of the original areas in AI and KR, and it remains an active and growing area of research within KR. 
Some of the most influential early AI work concerned planning using the STRIPS system~\cite{FikesN1971}.
Planning in general, and planning to address the needs of autonomous agents robots remains one of the core areas of AI. 
Many issues in planning involve representing knowledge about the world, and reasoning about the world as it changes. 
Nevertheless, today planning and KR are two largely separate areas of AI.
Consequently, a major challenge remains that of (depending on how one looks at things) generalising planners to incorporate the generality of KR theories of reasoning about action or, alternatively, coming up with computational models of reasoning about action that are the same order of efficeincy as current planning systems.

Within KR, an interesting trend is the convergence of reasoning about actions and the foundations of planning in AI on the one hand, and model checking, games on graphs, and synthesis in formal methods, on the other.
In particular, the rise of linear temporal logic on finite
traces~\cite{DBLP:conf/ijcai/GiacomoV13,DBLP:conf/ijcai/GiacomoV15},
which is particularly well-behaved computationally and applicable in practice,
has shed a new light on the interrelations among logic, automata, and games,
well-known in formal methods. It also provided a fertile ground for a new kind of
advanced research on reasoning about actions and strategic reasoning for
autonomous agents, including
planning with temporally extended goals~\cite{DeGiacomoRubin18},
self-programming agents~\cite{DeGiacomoEtAl22}, and strategy logic and
strategic reasoning~\cite{FijalkowEtAl22}.

Another trend, as exhibited by a corresponding series of
workshops,\footnote{\texttt{http://www.genplan.ai/resources/}} is
research on {\em generalisation in planning}, where one is interested
in general solutions for classes of problems, instead of just solving
single instances. In particular, instead of relying on plans in the
form of sequences of actions, solutions to generalised planning
problems can be programs (policies, controllers) that include control
structures such as branches and loops. Two representative examples of this research effort are contributions by De Giacomo
et al.~\cite{DBLP:journals/ai/GiacomoFLPS22}, who explore controller
synthesis in manufacturing scenarios based on first-order
representations in the situation calculus, and by Cui et
al.~\cite{DBLP:conf/ijcai/Cui0L21}, who present a framework for generalised
planning based on abstraction.


The trend to combine symbolic and machine learning approaches also shows in the area of planning and reasoning about action.
One particular issue is that approaches to planning and reasoning
about action and change rely on representations of action models,
which are traditionally crafted by hand. Instead, one wants to be able
to learn actions, affordances, and game rules from data or
observations. Asai and Muise \cite{DBLP:conf/ijcai/AsaiM20} present an
architecture for end-to-end learning of STRIPS representations from
images. Geffner \cite{DBLP:conf/aaai/Geffner22} argues more generally
that symbolic target languages are to be preferred over inductive
biases for learning to act and plan.

One form of machine learning that lends itself particularly well to integration with reasoning about action is that of {\em reinforcement
  learning}, where the goal is to train an agent to act in an
environment based on rewards administered after the execution of each
action. Here, symbolic representations can be used for representing
the learned policies (for example, in methods based on {\em policy gradient
  search}), or for the purpose of encoding reward functions more
succinctly, which can help to learn an optimal policy in a more
sample-efficient manner. Icarte et
al.~\cite{DBLP:journals/jair/IcarteKVM22} explore the effectiveness of
this approach based on a finite state machine representation, while De
Giacomo et al.~\cite{DBLP:conf/aaai/GiacomoIFP20} study ``restraining
bolts'' expressed through temporal logics over finite traces.

Below, we briefly discuss other research directions in the broad area of reasoning about action and planning. 

\smallskip
\noindent
\emph{Goal Reasoning}.
Whereas classical planning is concerned with finding a sequence of
actions to achieve a given, fixed goal, dynamic multi-agent
environments require that agents are also able to reason about {\em which}
goals to pursue. Hofmann et al.~\cite{DBLP:conf/icaart/HofmannVGHNL21} elaborate on goal reasoning in
the context of the Robocup Logistics League (RCLL), where teams of
autonomous robots have to coordinate their actions in a simulated
factory environment. Roberts et al.~\cite{DBLP:conf/aaaifs/RobertsHCCJA16} present ActorSIM, a general
software framework for studying multi-agent goal reasoning in
simulated environments. Goal reasoning has also been addressed in the situation
calculus~\cite{DBLP:conf/atal/GiacomoL20}.

\smallskip
\noindent
\emph{Beliefs}. 
Uncertainty is often connected to some notion of belief.
It is helpful to explicitly encode what the agent believes and does not
believe, and consider how an agent's beliefs are affected by its acting and
sensing. In this instance, action formalisms are
extended by epistemic modalities, one prominent example being Dynamic
Epistemic Logic (DEL). For example, Bolander et
al.~\cite{DBLP:conf/kr/BolanderDH21} consider the application of DEL
for epistemic planning in the context of human-robot
collaboration. While these formalisms traditionally rely on
qualitative notions of belief and uncertainty, probabilistic variants
have been studied as well. For example, Liu and Lakemeyer
\cite{DBLP:conf/ijcai/LiuL21} study belief and meta-belief for a
probabilistic action logic based on a modal variant of the situation
calculus.

\smallskip
\noindent
\emph{Hybrid Reasoning}. 
While traditionally, most approaches made use of qualitative
representations, many real-world applications require the ability to
handle quantitative information as well. Notable quantitative aspects
include time as studied, for example, by Cabalar et
al.~\cite{DBLP:conf/lpnmr/CabalarDSS22}, who present an approach for
incorporating metric time into Answer Set Programming for the
purpose of planning and scheduling. Another aspect is probabilistic
uncertainty, where Zarrie{\ss} \cite{DBLP:conf/kr/Zarriess18}, for
instance, presents results on the projection of stochastic actions in a
probabilistic description logic.

\smallskip
\noindent
\emph{Infinite-State Systems and the Situation Calculus}. 
While most research in planning is based on finite-state systems, important
research has also been conducted on infinite-state systems, most notably, at
least in recent years, in the Situation Calculus~\cite{Reiter01}.
This includes decidable reasoning about actions through a bounded (FOL) state
assumption~\cite{DBLP:journals/ai/GiacomoLP16},
a non-Markovian version of the situation
calculus~\cite{DBLP:conf/aaai/GiacomoLT20},
research on abstraction in the situation
calculus~\cite{DBLP:conf/ijcai/BanihashemiGL18}, and
controller/program synthesis in the situation calculus
\cite{DBLP:journals/ai/GiacomoFLPS22}.

\smallskip
\noindent
\emph{Ethics and Norms}.
Where embodied agents (such as mobile robots or autonomous cars) act
in physical environments that they share with human beings, it becomes
increasingly important that their actions are governed by social norms
and moral guidelines. The subarea of {\em normative (multi-)agent
  systems} \cite{chopra2018handbook} is concerned with the study of
such agents. Many approaches are based on integrating reasoning about
actions with some form of {\em deontic logic} to represent notions
such as obligation, permission, and prohibition. For instance, Horty
\cite{horty2019epistemic} studies obligations in the context of an
epistemic variant of {\em stit} (``see-to-it-that'')
semantics. Furthermore, Lindner et
al.~\cite{LindnerMattmuellerNebel2020} present an approach for
incorporating ethical reasoning into a planning system, distinguishing
act-based from goal-based deontological principles.

\subsection{Description Logics and Ontologies}
\label{sec:areas:DLsOntologies}

Description logics (DLs) were first introduced in the 1980s as an attempt to create a formal semantics for frames and semantic networks. Initially, DLs had limited expressive power and were designed for polynomial time reasoning. However, today they have evolved into a large family of languages that typically permit unary and binary relation symbols and some form of universal or existential quantification. While some DLs also permit second-order features like transitive closure or more general fixpoints, current DLs are mostly fragments of first-order logic and inherit its classical open world semantics. The research community generally agrees that DLs should be decidable in order to enable effective and robust reasoning support.

The DL community meets annually at the Description Logic Workshop,\footnote{\url{https://dl.kr.org/}} which had its 35$^{\text{th}}$ jubilee in 2022.  
It closely interacts with many other research communities, including the ASP and general non-monotonic reasoning community in efforts to combine the open world semantics of DLs with closed world features required in many applications.
As well, it interacts with the datalog and rule-based reasoning community to develop ontology-mediated data access. Because of its focus on decidable reasoning, the DL community also contributes to ongoing efforts of mapping out the boundary between decidable and undecidable fragments of first-order logic. 

Currently, the main view of DLs is as an underpinning language for ontologies, regarded in this document as logical theories that define the relevant classes, attributes, and relations for a domain of interest and that specify the relationships between them by means of logical axioms. Indeed, ontologies can be regarded as knowledge bases with a focus on terminological, schema, or conceptual knowledge. The design of ontology languages together with efficient reasoning engines has been a major research direction in DL.
This research has led to the World Wide Web Consortium (W3C) OWL ontology language \cite{McGuinnessVH2004} and its successor OWL~2 \cite{owl2} specifying three sublanguages, called \emph{profiles}, of OWL \cite{owl2-profiles}: OWL~2 EL, a fragment with polynomial time reasoning complexity used for large scale ontologies, OWL~2 QL, designed to support efficient access to data using ontologies, and the rule-based language OWL~2 RL. 

A basic research challenge for any ontology language is to understand the trade-off between expressivity and complexity of reasoning, and to develop, implement, and analyse efficient reasoning systems. In what follows, we distinguish between two rather different types of reasoning: \emph{terminological reasoning} that aims at extracting knowledge from an ontology by computing, for example, the induced concept hierarchy, and \emph{ontology-mediated querying of data} that supports access to data modulo an ontology. In the former case, ontologies are typically very large and often it is not the full logical theory underpinning the ontology that is needed in applications, but only the induced concept hierarchy or variants thereof. In contrast, in the latter case, ontologies are typically much smaller and their main purpose is to provide a schema for accessing, managing, and interpreting various types of data.  

Regarding terminological reasoning, amazing progress has been made over the past 15 years. There are now extremely powerful reasoning engines that compute the concept hierarchy induced by very large ontologies formulated in OWL~2 EL (for example, ELK~\cite{ELK-JAR}), in Horn extensions of OWL~2 EL, and even in very expressive description logics~(for example, HermiT~\cite{GHMS14a}, Pellet~\cite{SPGK06a}, or Konclude~\cite{DBLP:journals/ws/SteigmillerLG14}). A typical application of this type of reasoning is modelling support for SNOMED~CT,\footnote{\url{https://www.snomed.org/}} a comprehensive, multilingual clinical healthcare terminology with more that 350,000 concepts and used in more than 80 countries, which enables the consistent representation of clinical content in electronic health records. 
Reasoners for OWL~2 are, by now, so highly optimised that it is very challenging to implement novel optimisations into an existing reasoner. Equally challenging is the development of novel competitive reasoners based on new approaches. One possible direction is investigate a reasoner that translates to SAT, so that it can be computed by a state of the art SAT solver.

It remains a major research challenge to provide principled logic-based support for the development and maintenance of large-scale ontologies. 
Active research areas include the computation and presentation of \emph{explanations} of subsumptions (or, even more challenging, non-subsumptions) between classes~\cite{Horridge:PhD,DBLP:conf/rweb/GlimmK19}, automated support for \emph{modularisation and module extraction}~\cite{DBLP:series/lncs/5445}, \emph{versioning} and collaborative development of ontologies~\cite{JIMENEZRUIZ2011146,DBLP:conf/rweb/BotoevaKLRWZ16}, and \emph{schema manipulation} such as forgetting~\cite{KonevWW09,DBLP:conf/cade/ZhaoS18}.
All mainstream ontology languages in this area are fragments of first-order logic. Extensions of these languages to deal with defeasible knowledge and exceptions~\cite{DBLP:conf/dlog/MoodleyMS14,DBLP:series/ssw/BonattiPS20}, time~\cite{DBLP:conf/time/LutzWZ08}, and uncertainty~\cite{LUKASIEWICZ2008852} remain important research questions.

If ontologies are used to access data, the main reasoning problems often have a rather different flavour. There, reasoning becomes an extension of query answering over data, a problem 
originally addressed in database research. As data stored on the web and in data warehouses is often heterogeneous, distributed and only partially structured, it is often incomplete and even logically inconsistent. Dealing with such data requires complex and expressive data representation models and dedicated reasoning services to support data access, management, and interpretation. Hence the distinction between data and knowledge bases has become blurred and the insights gained in KR research for dealing with issues such as incomplete information and logical inconsistency have become directly relevant for mainstream database applications. 

A basic application of ontology-mediated query answering is to enrich an incomplete data source with background knowledge, in order to obtain a more complete set of answers to a query. Another application is data integration, where an ontology is used to provide a uniform view on multiple data sources abstracting from specialised  schemata and implementation details. Starting with DL-Lite~\cite{DBLP:journals/jar/CalvaneseGLLR07},  great progress has been made in understanding the complexity of answering queries under DL ontologies~\cite{DBLP:journals/ai/CalvaneseGLLR13,DBLP:journals/tods/BienvenuCLW14,DBLP:conf/rweb/BienvenuO15} and under existential rules~\cite{DBLP:journals/ai/CaliGP12}. Various types of systems such as Ontop and Vadalog have been developed that support ontology-mediated query answering~\cite{DBLP:conf/ijcai/XiaoCKLPRZ18,DBLP:journals/pvldb/BellomariniSG18,DBLP:conf/semweb/XiaoLKKKDCCCB20}. 

Many research challenges related to ontology-mediated data access remain to be solved, however. Active research areas include dealing with inconsistency in ontology-mediated querying-answering~\cite{DBLP:journals/ki/Bienvenu20}, temporal ontology-based data access~\cite{DBLP:journals/jair/ArtaleKKRWZ22,DBLP:journals/jair/WalegaZG23}, and reverse engineering of queries from data examples~\cite{DBLP:conf/www/ArenasDK16,DBLP:journals/ai/JungLPW22}. Moreover, research into crucial issues such as privacy and data provenance under ontologies has only just begun. 

\subsection{Knowledge Graphs}
\label{sec:areas:kgs}
In 2012 Google introduced the term \emph{knowledge graph} to refer to a large collection of facts accompanied by a basic ontology providing semantics. Since then, the term knowledge graph has been used to denote a wide variety of data models, which have in common that information is stored in a graph-like fashion, making use of labelled nodes and edges to represent objects and entities, facts
about them, and relationships between them (see the paper by Hogan et al.~\cite{10.1145/3447772} for a discussion of the basic concepts). These data models have become an important paradigm in both industry and research, with applications, for instance, in (web) search, as a backbone of Wikipedia through the structured knowledge in Wikidata \cite{10.1145/2629489}, and in personal assistants such as Alexa or Siri. Knowledge graphs are often of impressive size. For example, Google’s knowledge graph started with public sources such as Freebase, Wikipedia and the CIA World Factbook, but was quickly extended to contain more than 500 million nodes and 3.5 billion facts and relationships. In 2020, Google reported a size of 500 billion facts about five billion entities.\footnote{\url{https://blog.google/products/search/about-knowledge-graph-and-knowledge-panels/}} 
Microsoft, in its Bing search engine, also uses a knowledge graph.\footnote{\url{https://blogs.bing.com/search/2015/08/20/bing-announces-availability-of-the-knowledge-and-action-graph-api-for-developers/}} In 2015, that graph contained over one billion entities with more than 21 billion facts associated to them and over 5 billion relationships between them. 
Apart from numerous commercial knowledge graphs, Wikidata is an example of a large publicly maintained knowledge graph. 

Knowledge graphs give rise to difficult but exciting research challenges, both for the database and the knowledge representation community. From a database perspective, the design of new query languages that can navigate graphs in ways that go beyond first-order queries is ongoing~\cite{DBLP:journals/csur/AnglesABHRV17}. From a knowledge representation perspective, a principled approach to adding semantics to knowledge graphs without affecting the efficiency of query answering is an important but very hard problem. For instance, to enable the addition of logical rules to knowledge graphs, highly optimised rule systems such as RDFox~\cite{DBLP:conf/semweb/NenovPMHWB15} and VLog~\cite{DBLP:conf/semweb/CarralDGJKU19} have been developed. 
Further open research problems in this direction include dealing with meta annotations and with restrictions on relations. For example, in many knowledge graphs, relations are time-dependent; they are true for a certain amount of time, but not before or after. Modelling this in standard description logics requires expensive workarounds such as reification \cite{rdf11-semantics}. First steps
towards new modelling languages dealing with such issues 
are \emph{attributed description logics} presented by Krötzsch et al.~\cite{DBLP:conf/ijcai/Krotzsch0OT18}.
Another fundamental problem is support for automated updating and revising knowledge graphs, as discussed by Chaudhri et al.~\cite{KGs}.

Finally, one of the main practical problems in the area is the need for robust automated support for constructing and completing knowledge graphs. Standard ML approaches use embeddings of entities and relations into continuous vector spaces; see, for instance, the survey by Wang et al.~\cite{Wang2017KnowledgeGE}. These approaches, however, neither use nor respect the semantics that knowledge representation adds to knowledge graphs. Recent approaches to deal with this issue exploit textual knowledge within knowledge graphs in the form of
literals~\cite{DBLP:journals/access/AlamRNMAVL22} or include background knowledge in the embedding~\cite{DBLP:journals/pami/NayyeriXALY23}.

\subsection{Argumentation}
\label{sec:areas:Argumentation}

Argumentation is concerned with how arguments are supported or undermined by other arguments, and how these arguments and their components (claims, supporting or undermining claims) interact. It addresses with conflicts among arguments and in measures to evaluate arguments' plausibility. 

In practical settings, arguments are typically stated as logical formulas. 
Therefore, studying argumentation in such setting has had a steady presence in argumentation research. 
However, arguably the most consequential development in the theory of argumentation in KR was the emergence of abstract argumentation~\cite{Dung1995}. 
It does away with the structure of arguments and language-specific inference, and focuses instead on the basic relation between arguments, when one argument \emph{attacks} another. 
The resulting \emph{abstract argumentation framework} has attracted attention of researchers in several fields of AI, with deep connections to multi-agent systems, natural language understanding, machine learning and, more recently, explainable AI. 
%
Below, we outline some of the major, current research themes in argumentation. Howwever, our selection of themes is by no means meant to be comprehensive.

\smallskip
\noindent
\emph{Argumentation for explainable AI.} 
Decisions, choices and recommendations made by humans or by decision-support systems must be \emph{explainable}.
AI software systems, whether based on machine learning or logical rules and constraints, are often opaque and their outputs do not come with human-understandable explanations. 
\emph{Explainable} AI (XAI) aims to address this shortcoming. 
One of the promising directions in XAI is based on the concept of \emph{argumentative} explanations, and focusing on the human perspective~\cite{AntakiL1992}. 
It turns out that many forms of argumentative explanations can be cast in terms of current argumentation frameworks~\cite{CocarascuSCT2020,RagoCBT2020,PrakkenR2021}.  
This provides a setting for stating explanations and computational tools for establishing them. 
The approach applies both when explaining models in an argumentation framework formalism 
and when explaining machine learning models~\cite{CyrasRABT2021}. 
Argumentation-based XAI is in its early stages; one of its key challenges is to develop human-computer interaction tools that will effectively construct argumentative explanations in a human-understandable form.

\smallskip
\noindent
\emph{Argument mining}. This line of research studies techniques to extract, from text, arguments and their components such as premises and claims. 
It studies methods to arrange this data into abstract argumentation frameworks and applies inferential techniques to discover fallacies and inconsistencies in the original text. 
The area can be traced back to the work of Teufel et al.~\cite{TeufelSB2009}, as well as Mochales and Moenas~\cite{MochalesM2011}. The survey by Villata and Cabrio~\cite{VillataC2018} provides a good overview of the state of the art and a roadmap for the future. Recent work by Goffredo et al.~\cite{GoffredoHVCV2022} showcases the potential of argument mining by identifying fallacious arguments from the US presidential campaigns.

\smallskip
\noindent
\emph{Computational aspects of argumentation.} 
Another major theme in argumentation research concerns computational aspects, both regarding complexity of reasoning, and the design and implementation of argumentation systems. 
Initial research was limited to the domain of abstract argumentation and its plethora of semantics. 
A comprehensive account of the complexity landscape is presented by Charwat et al.~\cite{CharwatDGWW2015} and Cerutti et al.~\cite{CeruttiGTW2018}. 
However, it is also clear that arguments in practical settings have structure, including their claims and support not captured by abstract argumentation. 
Therefore, formalisms have been developed in which this structure is represented explicitly. 
These formalisms include assumption-based argumentation~\cite{BondarenkoDKT1997}, ASPIC$^{+}$~\cite{ModgilP2013,ModgilP2018}, and deductive argumentation~\cite{BesnardH2008,BesnardH2018}, among others. 
Recent papers include work by Lehtonen et al.~\cite{LehtonenWJ2021}, which provides a treatment of complexity and algorithms for assumption-based argumentation, and Dvo\v{r}\'ak et al.~\cite{DvorakGRW2021}, which deals with the instantiation-based approach~\cite{GorogiannisH2011}. 

\smallskip
\noindent
\emph{Applications.} 
From its inception, research in argumentation has been driven by practical applications. 
Currently, developing and implementing applications of argumentation is one of the most actively pursued research directions by the community. 
For instance, in medicine, argumentation is applied as a patient management tool \cite{CyrasOKT2021} and for persuasion to bring about changes in behaviour \cite{Hunter2018}; in shared governance, argumentation is used in collaborative decision support systems~\cite{SchneiderGP2013}. 
Law is another area that shows particularly strong connections with argumentation, both motivating research and benefiting from it~\cite{AtkinsonB21}. 
At present no argumentation-based systems aimed at legal reasoning have been deployed in practice, making law a particularly urgent and promising target area. 
Another emerging area of applications is \emph{judgmental forecasting} \cite{ZellnerABG2021}, a decision-making approach to situations when statistical methods are not applicable. 
A recently proposed variant of an argumentation formalism, a \emph{forecasting argumentation framework} \cite{IrwinRT2022} is guided by forecasting research and aims to support argumentation-based forecasting.

\subsection{Reasoning under Uncertainty}
\label{sec:areas:uncertainty}

The unification of logic and probability has always been a major concern in AI.   It is, therefore, not surprising that John McCarthy, who was first to suggest the use of logic for representing the knowledge of AI agents and the calculability of that knowledge, was also concerned about the role of probabilities. However, he was not quite convinced that there is any easy approach to adding probabilities to knowledge bases \cite{McCHay69}. His concerns notwithstanding, there is a great body of work on integrating knowledge representation and uncertainty, as years of engineering efforts in knowledge representation have shown us that there is pervasive uncertainty in almost every domain of interest. The upshot is that the “rigidity” (sentences always evaluate to \emph{true} or \emph{false}), “brittleness” (sentences in the knowledge base must be true in all possible worlds), and “discreteness” of classical logic have forced scientists to look at formalisms such as fuzzy and probabilistic logics. Below we enumerate some of the key developments and current research trends in this area. 

\smallskip
\noindent
\emph{Expressive probabilistic logics}. Starting with Nilsson’s probabilistic logic \cite{DBLP:journals/ai/Nilsson86}, and Halpern’s and Bacchus’ investigations on first-order logics of probability \cite{DBLP:journals/ai/Halpern90}, we now have a range of expressive modal propositional and first-order logics that allow us to reason about meta-knowledge, actions, plans and programs. In the recent years, there has been interest in epistemic planning, which can involve the modelling of mental states of multiple agents in domains with noisy effectors and sensors. 
    
\smallskip
\noindent
\emph{Tractable probabilistic logics}. Expressive probabilistic logics tend to be hopelessly undecidable, and so given the success of probabilistic graphical models, there has been a happy marriage of finite-domain relational logic and Bayesian and Markov networks. To go beyond essentially propositional models, there has been considerable recent work on probabilistic description logics,  as well as probabilistic logic programs. It is also worth noting that from the database community, a related formalism called \emph{probabilistic databases} has emerged.

\smallskip
\noindent
\emph{Learning of logical axioms}. There is a considerable body of work on learning propositional and relational formulas, and in the context of probabilistic information, learning weighted formulas, probabilistic automatons, and grammars. 

\smallskip
\noindent
\emph{Neural reasoning}. Given the increasing popularity of neural networks for low-level perception tasks, an emerging concern in both the neural and logical communities is how the two areas can be bridged. The resulting area of neuro-symbolic AI has now come to include approaches such as fuzzy logic, probabilistic logic, and differential statistical relational learning and inductive logic programming. 

For earlier work on reasoning under uncertainty, see the worrk by Pearl~\cite{Pearl1988}, while the Russell and Norvig text~\cite{RussellNorvig10} provides an extensive overview.
For a more recent overview and further details, see the work by Belle~\cite{Belle2017,Belle2022}.  

\subsection{KR and Machine Learning}
\label{sec:areas:ML}

The breakthroughs in \emph{machine learning} (ML), in particular the emergence of deep
neural networks, have brought about a new general trend in research on combining machine learning with symbolic or knowledge-based approaches. On the one hand, researchers expect KR methods to help tackle ML problems. On the other hand, they expect ML to help address some of the challenges of KR systems. Below we discuss how the two fields interact and list some of the key challenges that arise. A more through discussion can be found in the paper by Benedikt et al.~\cite{benedikt_et_al:DR:2020:11842}.

\smallskip
\noindent
\emph{KR for ML}. Despite the undeniable success of deep learning, the ability of neural networks to reason and to generalise in systematic ways has often been called into question~\cite{DBLP:journals/natmi/GeirhosJMZBBW20,DBLP:journals/jair/HupkesDMB20,DBLP:journals/corr/abs-2205-11502}. This view has led to the development of neuro-symbolic methods, which integrate deep learning architectures with explicit symbolic reasoning processes~\cite{DBLP:series/faia/342}. 
For instance, DeepProbLog~\cite{DBLP:journals/ai/ManhaeveDKDR21} uses probabilistic logic programs to reason about the predictions of a deep learning model. Similar combinations of neural networks with Answer Set Programming~\cite{YangIL2020} and Markov Logic Networks~\cite{DBLP:conf/uai/MarraK21} have also been proposed. 
More broadly, the usefulness of symbolic reasoning for machine learning has been studied within the context of Inductive Logic Programming~\cite{DBLP:journals/jlp/MuggletonR94}, and more recently, under the umbrella of Statistical Relational Learning~\cite{GetoorTaskar:book07}. Machine learning models can benefit from reasoning components in at least two different ways. First, the use of (rule-based) reasoning helps models to generalise in predictable and systematic ways. Second, training neuro-symbolic methods in an end-to-end fashion can help bridge the gap between the training data that is available, and the data that would be needed for training the individual components of the system in isolation. To illustrate, in DeepProbLog one can train a neural network   to recognise hand-written digits when provided only with training examples that specify the sum of two hand-written digits~\cite{DBLP:journals/ai/ManhaeveDKDR21}. 

Beyond reasoning, machine learning models can also benefit from symbolic knowledge in other ways. For instance, the so-called \emph{knowledge injection} methods use rules and constraints to regularise neural network models \cite{DBLP:conf/emnlp/DemeesterRR16,DBLP:conf/acl/LiS19,DBLP:conf/icml/XuZFLB18}. The underlying idea is essentially to discourage the model from making predictions that are incompatible with a given set of (soft) rules. Rules are also sometimes used as a form of weak supervision, to deal with a scarcity of labelled training examples \cite{DBLP:conf/iclr/AwasthiGGS20}. While rules and constraints encode specific semantic dependencies, simply knowing which concepts are relevant for a given domain can also be important. For instance, Concept Bottleneck Models \cite{DBLP:conf/icml/KohNTMPKL20} use such knowledge to ensure that the representation spaces that are learned by neural networks are semantically meaningful, and to some extent interpretable. Other approaches aim to discover semantically meaningful concepts, without prior knowledge, by designing models that learn vectors that can be interpreted as prototypes of concepts \cite{DBLP:conf/aaai/LiLCR18}. These prototypes can then be used for explanations.

Explainability and interpretability have become important topics within machine learning. Explanations can take many forms, including sets of input features \cite{DBLP:journals/kais/StrumbeljK14}, linear combinations of input features \cite{DBLP:conf/kdd/Ribeiro0G16}, or generated natural language sentences \cite{DBLP:conf/nips/WiegreffeM21}. Given that transparency is one of the key strengths of KR systems, it should come as no surprise that ideas from KR often play a central role in this context. For instance, one problem with generating explanations using language models is that such explanations are often not faithful \cite{DBLP:conf/acl/CamburuSMLB20}. One possible solution is to develop models that infer the answer by incrementally constructing the analogue of a proof tree \cite{DBLP:conf/emnlp/DalviJTXSPC21}. Somewhat related, large language models are capable of generating step-by-step derivations for answers that require reasoning (known as \emph{chains of thought} in this context), by generalising from a few examples of such derivations \cite{DBLP:journals/corr/abs-2201-11903}. However, the proof strategies that are implicitly employed by these models are rather primitive \cite{DBLP:journals/corr/abs-2210-01240}, which suggests that a hybridisation with KR methods would improve their capabilities. Specific KR frameworks can also be used more directly. For instance, techniques from computational argumentation have long been used for modelling explanations \cite{DBLP:conf/ijcai/Cyras0ABT21}. As another example, when explaining entire models, rather than individual predictions, default rules often allow for significantly more compact explanations than decision trees or traditional rules \cite{DBLP:conf/ijcai/KuzelkaDS16}. Another potential application of KR methods is to formally verify 
properties of ML models, such as their robustness against adversarial attacks~\cite{DBLP:conf/aaai/NarodytskaKRSW18,DBLP:conf/fmcad/Narodytska18,10.5555/3495724.3495875}.

\smallskip
\noindent
\emph{ML for KR}. The knowledge acquisition bottleneck is one of the core challenges in KR (see also Section~\ref{sec:challanges:knowledgeAcquisition}). KR systems need access to structured knowledge, encoded in some formalism, but such encodings are not readily available for most domains, and tend to be expensive to obtain. There are several ways in which ML techniques can be used to alleviate this issue.  ML models can be used, for instance, to identify and exploit statistical regularities in existing knowledge bases. 
This resulted in strategies for automatically extending knowledge graphs with plausible triples \cite{DBLP:conf/nips/BordesUGWY13,DBLP:journals/corr/YangYHGD14a,DBLP:journals/jmlr/TrouillonDGWRB17,DBLP:conf/iclr/SunDNT19,DBLP:conf/emnlp/BalazevicAH19} and methods for identifying plausible missing subsumptions in ontologies \cite{DBLP:conf/semweb/LiBS19,DBLP:journals/corr/abs-2202-09791}.
A large number of methods have been developed for converting knowledge expressed in text into a structured format \cite{DBLP:conf/aaai/CarlsonBKSHM10}. While most work has focused on knowledge graph triples, learning more expressive knowledge has also been considered \cite{PETRUCCI201866}. More recently, the focus has shifted to extracting knowledge directly from (large) language models \cite{DBLP:conf/emnlp/PetroniRRLBWM19,DBLP:journals/corr/abs-2206-14268,DBLP:journals/corr/abs-2301-12810}. The success of language models as knowledge acquisition tools, however, extends beyond knowledge graphs. For instance, Hwang et al.~\cite{DBLP:journals/corr/abs-2010-05953} use language models to capture social commonsense knowledge, while Jin et al.~\cite{DBLP:journals/corr/abs-2204-10176} and Zellers et al.~\cite{DBLP:conf/cvpr/ZellersLLYZSKHF22} induce script knowledge.

ML methods have also been proposed for improving reasoners, or even as an alternative to the use of classical reasoners. One strand of work is focused on approximating the entire reasoning process using neural networks, for instance to allow for efficient and inconsistency tolerant reasoning with ontologies \cite{DBLP:conf/aaai/AbboudCL20}, or to heuristically search for solutions to computationally intractable problems~\cite{DBLP:conf/aaai/PratesALLV19}. Another possibility is to use ML methods within traditional reasoners, for instance, for learning suitable heuristics \cite{DBLP:conf/sat/HaimW09,DBLP:conf/sat/LiangGPC16}.

\subsection{KR and Robotics}

A robot is a physical agent that carries out tasks by interacting with the
world via its sensors and effectors. Prominent examples are vacuum robots or robots operating in
warehouses, but also autonomous vehicles in urban environments.
Given that robots need to be given or to learn or acquire domain and general knowledge and to reason with this knowledge to solve problems such as building plans, the field of robotics would
appear to be a prime application of KR. Indeed, in one of the first robotics projects in the late
1960s, the robot {\em Shakey} already made use of explicit representations of knowledge and was able
to plan a course of actions to move objects between rooms. While the reasoning techniques developed
for Shakey have long been superseded by more advanced techniques, it is remarkable that the planning
language STRIPS~\cite{FikesN1971}, which also resulted from the Shakey project, has survived until today.
However, after Shakey, KR played almost no role in robotics for a long time, mainly because more immediate
problems like safe navigation in unknown environments needed to be solved first. A breakthrough in
this regard came about in the mid 1990s with the development of probabilistic robotics~\cite{ThrunBF05}. It resulted in 
impressive artefacts like the museum tour guide robots Rhino and Minerva~\cite{Rhino99,Minerva00},
and in the re-emergence of KR techniques like the action language Golog~\cite{Golog97}
controlling the high-level behaviour of Rhino.

Around the same time, Ray Reiter and colleagues established a brand of {\em
  Cognitive Robotics} where KR plays an essential role. To quote from a recent Dagstuhl report on the area:
\begin{quote}
  \em
  Cognitive Robotics is concerned with endowing robots or software agents with higher level
cognitive functions that involve reasoning, for example, about goals, perception, actions,
the mental states of other agents, collaborative task execution, etc. This research agenda
has historically been pursued by describing, in a language suitable for automated reasoning,
enough of the properties of the robot, its abilities, and its environment, to permit it to
make high-level decisions about how to act.
\end{quote}
A concise summary of some of the early work on Cognitive Robotics based on the situation calculus is given by Levesque and Lakemeyer~\cite{LakemeyerLevesque08}. Despite the advances of Cognitive Robotics, it is fair to say that this vision of placing KR at the heart of
robotics has not yet been fully realised.\footnote{Indeed, the robotics community has developed other
versions of Cognitive Robotics with much less emphasis on KR techniques as in the work of Cangelosi and Asada~\cite{CangelosiM22}.}

Nevertheless, KR plays an important role in controlling the high-level behaviour of many of today's robotic systems.
One of the best known KR frameworks for robots is the {\em KnowRob} system developed
by Tenorth and Beetz \cite{BeetzTenorth17}, combining rich ontologies with specialised reasoners
like temporal and qualitative spatial reasoners.
Planning techniques have also played a major role. While early work focused on domain-dependent planners like
IxTeT~\cite{GhallabLaruelle94} and TAL~\cite{KvarnstromDoherty00}, domain-independent PDDL planners like FF~\cite{HoffmannNebel01},
TFD~\cite{TFD12}, and ROSPlan~\cite{ROSPlan15} are now used routinely in robotic systems. As mentioned above, the action language Golog was
employed early in controlling the robot Rhino. More recently, Golog has also been integrated
with PDDL planners~\cite{ClassenRLN12}. Besides task-level planning, goal reasoning has also found
its way into the world of robotics. For example, Hofmann et al.~\cite{DBLP:conf/icaart/HofmannVGHNL21} consider
production logistics scenarios, where robots need to concurrently entertain multiple goals (requests
for products), which may need to be prioritised and sometimes abandoned due to timing constraints or failures.
An interesting direction specific to robotics is the combination of both task and motion planners (see~\cite{TaskMotionPlanning20} for a
survey), which can help produce more robust plans, especially in scenarios involving object
manipulation. Extending this idea to conditional planning~\cite{NoumanPE21} allows for
handling incomplete information and partial observability.

When a robot has decided on a course of actions, their execution needs to be constantly
monitored. An early KR approach to execution monitoring in the context of Golog programs was proposed by De Giacomo et al.~\cite{DBLP:conf/kr/GiacomoRS98}. More recently {\em continual
planning}~\cite{DBLP:journals/aamas/BrennerN09,DBLP:conf/aaai/HofmannNCL16} not only tightly
integrates planning and execution monitoring, but also allows for postponing plan refinement until
enough information is gathered at runtime.
Regarding the diagnosis of failures, model-based techniques with strong KR
foundations have also been developed~\cite{ZamanSMLU13,LeauteWilliams05}.
While execution failures cannot be avoided in general, efforts are being made to make robots safer,
more robust and trustworthy. In this regard, verification plays an important role with many
connections to KR. For a recent survey on verifying robotic systems, we refer interested readers to the work of Luckcuck et al.~\cite{LuckcuckFDDF19}.

Needless to say, many challenges remain when it comes to realising the vision of KR-based cognitive
robots. A list of many of these challenges can be found in the Dagstuhl report on cognitive robotics~\cite{HeintzLM23}.

\subsection{KR and Information Systems}
\label{sec:areas:infosys}

Ontologies and knowledge graphs (c.f.~Sections~\ref{sec:areas:DLsOntologies} and \ref{sec:areas:kgs}) are two key KR contributions to information systems research and practice, as we have discussed already. In Section~\ref{sec:areas:DLsOntologies}, we also considered ontology-mediated data access, which addresses fundamental questions originating in database research by utilising advanced KR techniques. Variants of datalog, description logics, and existential rules are its main underpinning logical formalisms.
 Closely related to ontology-mediated data access in terms of techniques and goals are declarative approaches to data exchange~\cite{DBLP:journals/tcs/FaginKMP05,DBLP:books/cu/ArenasBLM2014} (transform data structured under a source schema into a target schema using logical rules) and data integration~\cite{DBLP:conf/pods/Lenzerini02,DBLP:books/sp/18/GiacomoLLPR18} (combine data residing in different sources and provide users with a unified view of them).
Description logics have also been proposed as a logical underpinning of standard database design formalisms such as ER and UML diagrams~\cite{DBLP:conf/krdb/FranconiN00,DBLP:journals/dke/QueraltACT12}. Highly optimised description logic reasoners discussed earlier are then used to check their consistency and logical consequences. Recently, KR methods developed in non-monotonic reasoning and description logics have been used to further develop the W3C standard SHACL (Shapes Constraint Language)~\cite{DBLP:conf/rweb/Pareti021} for validating graph-based data against sets of constraints to enhance semantic and technical interoperability~\cite{DBLP:conf/www/AndreselCORSS20}.

These applications of KR methods to problems in information systems and databases have the common feature that they are concerned with structural and static aspects of information. In contrast, KR methods have only recently been applied when the dynamic behavior of information systems is taken into account. This is despite the fact that the integration of structural and behavioural aspects to 
capture how information systems dynamically operate over data is recognised as one of the main challenges in business process management (BPM) and, more generally, information systems engineering. Research communities addressing this challenge using KR methods include the AI4BPM workshop (co-located with the BPM conference) and the PM4AI workshop (co-located with IJCAI). Three aspects of BPM are particularly relevant for KR and should be considered when KR methods are applied:
\begin{enumerate}
\item Business processes are modelled, configured, executed, and continuously improved based on the so-called business process lifecycle, where every phase calls for reasoning support, ranging from model-driven verification to the combined analysis of models and event data tracing the actual process execution (known as process mining~\cite{DBLP:books/sp/Aalst16}).
\item Business processes vary depending on their complexity, predictability, and repetitiveness. The modelling paradigms used in BPM reflect these differences, ranging for instance from procedural to declarative approaches.
\item Business processes are specified in BPM using concrete modelling patterns. These should guide the use of modelling restrictions needed to achieve decidability or tractability of reasoning tasks.
\end{enumerate}
We briefly discuss the state of the art of KR applications in this field. 
First, the integrated modelling of processes and data yields infinite-state relational transition systems where each state comes with a first-order interpretation. The analysis of standard properties (such as reachability and safety), or more sophisticated properties expressed in variants of first-order temporal logics have been studied both under complete \cite{DBLP:conf/pods/CalvaneseGM13} and incomplete \cite{DBLP:conf/birthday/ClassenLZ19,DBLP:journals/jair/HaririCMGMF13} information over relational states, and also in the presence of read-only data \cite{DBLP:journals/siglog/DeutschHLV18}. 
Relevant KR techniques include reasoning about action~\cite{DBLP:journals/iandc/CalvaneseGMP18} and planning~\cite{DBLP:conf/aaai/Borgwardt0KKNS22}.

Second, KR methods are suitable for the analysis of knowledge-intensive processes, where flexibility is a key requirement \cite{DBLP:books/daglib/0030179}. Here methods developed in the context of decidable first-order temporal logic~\cite{gabbay2003many,DBLP:journals/tocl/BaaderGL12} and temporal conceptual modelling~\cite{DBLP:conf/birthday/ArtaleF09} are relevant. Applications of KR approaches in declarative process modelling, management, and mining are discussed by Di Ciccio and Montali~\cite{DBLP:books/sp/22/CiccioM22}.

Finally, KR methods should be utilised to tackle new reasoning problems emerging from process mining. This should facilitate  the formal definition of tasks and the exploration of their decidability and complexity status; and it should then also be possible to utilise automated reasoning techniques instead of ad-hoc algorithmic approaches. Notable examples of logic-based techniques employed already are planning~\cite{DBLP:conf/aaai/XuLZ17a}, SAT~\cite{DBLP:journals/computing/BoltenhagenCC21}, and SMT~\cite{DBLP:conf/bpm/FelliGMRW21}.

\subsection{KR and Logic/Philosophy}
Classical first-order logic was originally developed in an attempt to
formalise the foundations of mathematics, with the monumental work of Russel and Whitehead perhaps being the primary example \cite{WhiteheadRussell1927}.
A good representation of early papers has been collected by van Heijenpoort \cite{vanHeijenoort1967}, and an entertaining history of logic has been provided by Doxiadis and Papadimitriou \cite{DoxiadisPapadimitriou09}.
Later work focused on philosophical notions including strict vs.\ material implication, necessity vs.\ contingent truth~\cite{HughesCresswell68},
issues concerning naming and reference~\cite{Kripke80}, and the like.
So the use of logics -- whether propositional, first-order, modal, relevance,
or others -- to represent and reason about real-world domains can be regarded
as a radical shift in the application of logic in general.
Indeed, it can be noted that the KR Hypothesis~\cite{Smith82}, which asserts that a knowledge based approach is
\emph{essential} for the construction of any generally intelligent agent, inextricably links logic and
KR since it posits semantically meaningful declarative structures in any
generally-intelligent agent.

The relation between KR  on the one hand, and philosophy and logic on the other hand, has been a mutually beneficial one. The list of benefits drawn by both sides, so to speak, is long. The example of belief revision originating as part of philosophy (Section~\ref{Sec:BR}) is but one of many. Very broadly, KR (and indeed AI as a whole) has been a source of new problems and issues for logic, and through real-world applications has served to sharpen such issues.
On the other hand, formal logic has provided the tools and direction for addressing such problems and issues.
Indeed, much formal work in KR involves the development and elaboration of new and existing logics.

Here, we merely provide a few illustrative examples of such benefits and synergies.  
Interesting reasoning problems, such as the frame, qualification, and ramification problems have been identified and addressed~\cite{McCHay69,McCarthy77,Thielscher97,Reiter01}.
Areas originating in philosophy, such as belief revision (Section \ref{Sec:BR}) and deontic reasoning~\cite{GabbayEtAl13}, see much (if not most) of their progress coming from the KR community. Much current work in modal logic is driven by research in KR and computer
science more broadly. Work in non-monotonic reasoning has contributed significantly to ``extended'' logical reasoning, in particular fixed-point semantics~\cite{DeneckerMarekTruszczynski03}, and the preferential structures advocated for by Shoham~\cite{Shoham1987} and subsequently extended by others.  Work in description logics has provided a broad and deep analysis of useful fragments of first-order logic and attendant complexity properties~\cite{BaaderEtAl07}. In the other direction, the realization that many description logics are syntactic variants of modal logics~\cite{Schild1991} enabled the description logic community to draw on many of the theoretical and technical results that have been developed for modal logics over the years.

\subsection{Summary}

The areas described in the preceding subsections is to be understood as a starting point for a further, individual literature research; and, certainly, much further works exists within the described areas, between them, and also beyond them. 
For example, dealing with uncertainty (Section~\ref{sec:areas:uncertainty}) is an important issue in essentially all other above described areas, while there are further areas in or linked to KR such as in spatial and qualitative reasoning (see, for example, the survey by Chen et al.~\cite{DBLP:journals/ker/ChenCLWOY15}), epistemic logic (see, for example, the introductory text by van Benthem~\cite{10.2307/4321713}), or commonsense reasoning (see, for example, the work by Davis and Marcus~\cite{10.1145/2701413} or the recent book by Brachman and Levesque~\cite{BrachmanLevesque22}).







\section{Major Research Challenges for KR}
\label{sec:challenges}

In the previous section, we presented several areas within or related to KR. We included in that discussion research problems and challenges specific to individual areas. 
In this section, we discuss major overarching challenges for KR that cut across several areas. In doing so, it is useful to keep in mind the following questions that have always been fundamental to KR.  
\begin{itemize}
    \item What knowledge does a system need to have in advance, as opposed to what can be acquired by observations? 
    
    \item What is the language for representing and reasoning with the background knowledge and observations? 
    
    \item What kind of semantics governs the updating of knowledge, given new and possibly conflicting observations?
    
    \item What impact does continual learning have on the entailments of a  knowledge base?
    
    \item What is the mechanism for requesting “shallow” versus “deep” reasoning (for example, quickly reacting to a tiger versus reflecting on the nature of the universe)? 
    
    \item How does a system generalise from low-level observations to high-level structured knowledge? What effect might this have on the computational tractability of the overall system?
\end{itemize}

As is clear from the remainder of this section, much work remains ahead of us, and these are deeply challenging and exciting questions, where knowledge representation and symbolic approaches more generally will have a significant role to play.

\subsection{Commonsense Reasoning}
\label{sec:challenge:commonsense}


Commonsense reasoning can be defined as reasoning in which an agent has ``the ability to make effective use of ordinary,
everyday, experiential knowledge in achieving ordinary, everyday, practical
goals''~\cite{BrachmanLevesque22}.
Broadly speaking, this might be thought of as ``what a typical seven year old
knows about the world, including fundamental categories like time and space,
and specific domains such as physical objects and substances; \dots humans,
their psychology, and their interactions; and society at
large''~\cite{Davis17}.%
\footnote{Davis~\cite{Davis17} gives a comprehensive survey of issues in
commonsense reasoning from a KR perspective.
Several points below are drawn from this paper.}
Clearly it is something that humans possess and use effectively.
It is broadly relevant in many areas of AI, including natural language
understanding, constructing visual interpretations, planning, and interacting with the real world.
In KR, different aspects of commonsense reasoning have been addressed by means of concepts and techniques developed for non-monotonic
reasoning, temporal and spatial reasoning, approaches to action and change,
qualitative reasoning, and others.

However, no successful commonsense reasoning system of substantial
breadth has yet emerged out of KR research (efforts by projects such as CYC
notwithstanding) and major challenges remain.
First, many domains are not well understood, and it is not clear how they
may be formalised, both with respect to representing commonsense knowledge
and in defining effective inference mechanisms.
While substantial progress has been made in areas such as temporal reasoning, and in reasoning about action
and change, much remains to be done with regards to physical processes,
knowledge and mental attitudes, social mores and attitudes, and the like.

Second, in many formal settings such as mathematics, databases, information systems, and most forms of declarative programming, reasoning is deductive, and is
based on classical logic, or a fragment thereof, as in description logics.
However, in commonsense reasoning, conclusions are rarely deductive, but
rather are tentative and based on the best information (and perhaps lack of
information) available.
So reasoning for the most part is \emph{plausible}, and conclusions may be
retracted as other information is gleaned.
Again, much progress has been made, notably in non-monotonic reasoning on the
one hand, and reasoning under uncertainty on the other.
But there is no comprehensive account of plausible reasoning in general, nor any
substantial understanding of how different approaches relate to each other.

Third, commonsense domains are subject to the so-called \emph{long tail
phenomenon}.
While general rules may account for a large proportion of domain instances,
most often there is also a large (arguably unbounded) number of exceptional instances.
This crops up in, for example, the notion of exceptional individuals in
default reasoning, and in the \emph{qualification problem} in reasoning about
action, wherein it is often impossible to list all preconditions to an action.
As Brachman and Levesque~\cite{BrachmanLevesque22} note, even when 
an individual exceptional condition occurs very rarely, there will usually be enough of them that
\emph{some} exceptional circumstance will arise with reasonable likelihood.
The challenge for a commonsense reasoner is how to handle exceptional
circumstances that are impossible to foresee.

Last, it is a natural question to ask whether generative systems such as
ChatGPT, trained on vast amounts of human knowledge, possess common sense and
to what degree.
Thus, an interesting question arises as to how systems such as ChatGPT can be
analysed for their commonsense capabilities:
where they display comprehensive and convincing commonsense capabilities, and
where they are lacking.
Related to this is the question of how KR can be used to enhance the commonsense capabilities of such systems.
This emphasises another major challenge for KR of developing frameworks to
support integration of diverse systems, both knowledge-based and
non-declarative.

\subsection{Knowledge Acquisition and Maintenance}
\label{sec:challanges:knowledgeAcquisition}

One of the main challenges for knowledge representation and reasoning was and is the task of knowledge acquisition and maintenance, that is, the formulation of suitable domain knowledge by knowledge engineers and domain experts. While this challenge is not a new one, for instance, Studer et al.\ discussed the challenges already in 1998~\cite{STUDER1998161}, it is still a challenge today with a dedicated conference series, the International Conference on Knowledge Engineering and Knowledge Management.
In the area of ontologies, the presence of tools such as the Prot\'eg\'e ontology editor
certainly helped to promote ontologies. In other KR-related areas, for example, planning, the development of methods and tools for modelling support are currently starting to emerge (see, for example, the work of Lin et al.~\cite{Lin2023RepairingClassicalModels} for plan modelling support). The presence of tools for modelling support alone is, however, not sufficient to solve the knowledge acquisition and maintenance problem. 

Some large knowledge graphs are successfully built and maintained by a community, for example, Wikidata~\cite{10.1145/2629489}, or are based on semi-automatic or automatic extraction (see, for example, the work of Ji et al.~\cite{JiPCMY2022} for a recent survey on knowledge graph acquisition and use) and there are even some patents for knowledge acquisition in knowledge graphs.\footnote{For example, patent US20180144252A1 ``Method and apparatus for completing a knowledge graph'' by Fujitsu Ltd.} 
These examples, however, typically use rather lightweight schema languages and stay away from treating evolving knowledge as a belief revision task. 

There are also approaches to learn schema axioms/rules in more expressive languages \cite{KR2022-51} and first-order rule learning has been extensively studied in the area of Inductive Logic Programming \cite{ILP,10.1613/jair.1.13507}.
Such learned knowledge, however, may be inconsistent or it may come with a degree of uncertainty. In order to deal with this, a knowledge engineer could check and disambiguate the learned knowledge, which is difficult at least for some domains. Another option is better support for reasoning with uncertainty; we have discussed this area and its challenges in Section~\ref{sec:areas:uncertainty}. Depending on the domain, one might also need support for commonsense reasoning, which we discussed in Section~\ref{sec:challenge:commonsense}). Overall, better support for acquiring and maintaining large, complex, but still \emph{consistent} knowledge bases or better capabilities for reasoning with \emph{inconsistent} knowledge is a challenging but necessary task for demonstrating the use of KR techniques within applications. 
Work on these topics generally requires empirical evaluations, rather than establishing theoretical foundations.
The KR community needs to support such efforts more actively.

\subsection{KR and Hybrid Systems}
\label{sec:challenges.ML}
The paradigms that have been developed in KR and ML clearly have complementary strengths. Integrating approaches from both fields into hybrid systems, thus, is intuitively appealing. In practice, however, there are a number of important and non-trivial challenges that need to be overcome. One issue is that KR methods are typically designed to work with clearly defined and carefully formulated knowledge. The knowledge generated by ML methods, on the other hand, is inevitably noisy. Beyond the possibility of errors in the extracted knowledge, there are also issues related to the vagueness of natural language concepts. KR settings normally assume that predicates have well-defined and precise meanings, even if these meanings can be application-specific and sometimes somewhat arbitrary. 
When knowledge is obtained from text,  we have no guarantees that the concepts or predicates involved have the intended meaning. Similar problems arise in the context of ontology alignment, but the informal nature of extracted knowledge tends to make these problems more severe.  In terms of reasoning, neuro-symbolic methods essentially aim to reason about the output of a neural network model. This means that the learning and reasoning components may be only loosely coupled. 
It would be beneficial to have a tighter integration, where the steps of knowledge extraction and reasoning are intertwined.
There are also various challenges of a more practical nature.  
Work in this area would clearly benefit from more easy-to-use implementations of reasoners, that are both scalable and that can deal with uncertainty. 

In light of recent developments, a very important aspect to consider here is the case of Generative Models and their relationship to KR. Software tools such as ChatGPT offer a mapping from the vast amounts of information used to train them to well-structured, understandable natural language responses to prompts (queries) also presented to it in natural language. The scope of such systems is quite broad with no topic being off-limits. They provide responses to prompts concerning all aspect of human daily life, respond to questions concerning science and solve technical problems. These responses are often quite accurate and detailed. They also demonstrate a good level of reasoning ability. While it is not hard to find prompts that result in incorrect or confusing answers, the question is whether these systems are (or ought to be viewed as) knowledge-based systems. If they are, what are their weaknesses, and how can they be improved? 
That is, how can they be analysed from a knowledge-based perspective?
And if they aren't, why not? Such questions are now critically important as they could well determine the future direction of KR.

Finally, there are a number of unfortunate barriers between the KR and ML communities, going from differences in terminology to differences in expectations for publications (for example, in terms of the balance between theoretical development and empirical validation).

\subsection{Explainable AI (XAI)}
It has been argued that KR can address the central problem of explainable AI (XAI), which is to develop tools and methods to interpret predictions or recommendations made by models developed using machine learning and to present them in a human-understandable form. 
Indeed, ``interpreters'' of machine learning models implemented as declarative, rule-based knowledge representation systems would arguably allow us to build explanations of decisions or recommendation, by employing the abductive reasoning capabilities of such systems. 
Recent work in KR, especially in the area of argumentation (see Section~\ref{sec:areas:Argumentation}), supports the view that KR is potentially a viable approach to XAI. XAI is, however, still in its infancy and providing compelling evidence of applicability of KR for XAI remains an open problem. One point of concern is whether one can build KR-based interpreters to machine learning models that are accurate enough to provide a basis for building explanations. 
If one could, they could be used instead of machine learning models in the first place.

Even for the more modest goal of explaining predictions and recommendations made using symbolic KR methods, many challenges remain to be solved. In fact, recently there has been a revival of interest in ``explainable KR'', witnessed for instance by the workshop series Explainable Logic-Based Knowledge Representation (XLoKR). In principle, if standard symbolic KR methods are used, predictions and decisions should be easy to explain. For example, if knowledge is represented in (some fragment of) first-order logic, and a decision is made based on the result of a first-order reasoning process, then one can use a formal proof to explain a positive reasoning result, and a counter-model to explain a negative result. Despite many years of research, however, in practice things are not so easy. For example, proofs and counter-models may be very large or very hard to comprehend for a non-specialist users. To come up with good explanations, even in the symbolic case, many challenges remain to be solved.



\section{Promoting KR}

Perhaps the single most important ingredient necessary for success of any field of science is to have an exciting, relevant and potentially high-impact research program that promises to advance broad societal and scientific needs.
We have argued in previous sections that KR is an active and important field, both as a subarea of AI, and as a contributor or partner to other areas.
Moreover we have suggested that KR is fundamental for the development of any form of general artificial intelligence.  
Thus, the field of KR as outlined in this manifesto is clearly in a position to continue its vibrant development. 

Nevertheless, KR faces significant challenges. 
It has often been viewed by researchers in AI and, more broadly, CS, as being highly theoretical, perhaps esoteric, and lacking in practical applications. More recently, significant recent successes of machine learning methods, showcased by compelling applications, have overshadowed work in KR. This has resulted in a decrease in open academic positions in KR, as well as in the number of students and researchers attracted to the field. It might also be the reason for a geographic imbalance in KR research activity, with a stronger research contingent to be found in Europe than in other parts of the world.
In this section, we suggest a number of measures to promote KR, so that it remains an attractive, relevant, and thriving research area.

\subsection{Broadening the Scope of the KR Conference Series} 
The International Conference on Principles of Knowledge Representation and Reasoning, that is, the KR conference series, has been one of the most significant vehicles driving research in the area of knowledge representation. 
Consequently, it provides a major means and opportunity for promoting KR research and development in the future.
Here we make some suggestions as to how this may be accomplished.

While previous conferences in the series have welcomed and encouraged papers on applications,
the conference should nonetheless strive to make more room for research focusing on practical applications relevant to current societal needs and goals.
It should become a forum to discuss and showcase not just theoretical advances, but also successful deployments of KR systems. 
As part of this effort we recommend that program committees make a distinction between actual applications deployed and in use from work that is mostly theoretical (although of course theoretical work should also be motivated by possible future application). 
Further, as a step towards practical applications, the conference should promote experimental studies by seeking submissions presenting data sets, and collections of benchmark problems, and should sponsor KR system competitions as a mechanism for achieving performance improvements necessary for moving proof-of-concept solutions to practical and effective implementations. 

As we argued, it is important to see KR in the broader context of other areas of AI and beyond AI.
To promote these synergies and strengthen research cutting across different fields, the KR conference should seek productive collocations with other conferences such as ISWC, ICAPS, TARK, and ICLP, as well as with prominent conferences in machine learning. Further, the KR conference should continue to be open to participate in federated conferences such as FLoC. 
Starting a federated conference series centred around knowledge representation and with the KR conference as its centrepiece should also be explored. 
The program of KR conferences should reflect the importance of promoting cross-area interactions by the strategic selections of invited speakers and tutorials from bordering areas, and by holding panels and special topic sessions covering areas most promising for successful integration with KR.

A related topic concerns the location of KR conferences. 
We suggest that, in general, KR conferences should be located where they may benefit and have an impact on the local KR community.
While esoteric locations may make for interesting travel destinations, they are often expensive, thus making it difficult for researchers, particularly students, to attend.
As well, it is essential that any conference location allows for low-cost accommodation.
It should be a goal of the conference to keep registration fees affordable;
in selecting a conference venue, it may be worthwhile considering less expensive settings, such as those that may be provided by a university, in place of resorts or high-end conference facilities.

From its inception, the KR conference was distinguished by its emphasis on more mature and in-depth research than typically found in major AI conferences.
Consequently the page limit for submissions has generally been 9-10 pages, as opposed to the more usual 5-6 pages.
This has arguably resulted in a higher portion of archival-quality papers.
However, this higher threshold comes at a price.
Authors eager to present the most recent results of their research may find the technical depth expectation hurdle hard to overcome, particularly in the presence of deadlines and with other highly competitive venues with less stringent requirements for a submission.
As well, papers that are accepted for a KR conference may be already developed to the degree that makes extensions required by a subsequent journal publication much harder. 
This has been addressed to some extent by allowing short (4 page) submissions, where these short papers are otherwise subject to the same reviewing criteria.
We recommend that the KR steering committee review the current full paper submission requirements, with an eye to increasing submissions while not compromising technical quality.
Here are a few suggestions to this end:
First, the KR submission requirements could be brought into line with other conferences such as AAAI, ECAI, or IJCAI;
however this might be seen as erasing part of KR's unique character.
Second, the call for papers could be rephrased to emphasise that regular submissions need not be of maximum length, stating something like 
``contributions are welcome for both regular papers (in the range of 6 - 9 pages) and short papers (up to 4 pages). Regular papers that are under 9 pages will not be penalised in the reviewing process''.
(It is ironic that when a conference talks about a page limit of $n$, while this is technically an upper limit, in practice it is treated as a lower limit by authors and reviewers alike.
This suggested change tries to emphasis that shorter regular papers are welcome and encouraged.)

A third, intriguing possibility follows from the observation that accepted KR papers are already at, or close to, the requirements for a journal submission.
Consequently, the steering committee should investigate whether there may be a simple process for having regular accepted papers appear as journal articles with minimal additional effort.
This is not a new suggestions, and other conferences such as ICLP, SIGGRAPH, and VLBD have been doing this for years.
However, given KR's emphasis on longer, mature submissions, KR papers would be uniquely suited for such treatment.
\emph{Which} journal may be suitable for this is open to debate, but it may be that JAIR or some other forum would be open to having a special ``Transactions in KR'', or some such stream.

\subsection{Broadening the KR Community} 
Another essential goal for the KR community is to ensure fresh cohorts of researchers who consider KR an attractive, inspiring and exciting area. 
It is important to educate students in KR related topics, and to engage them early on in research efforts. It is similarly important to support young researchers, promote collaborations, and recognise research accomplishments. Finally, it is critical to develop programs bringing awareness of KR and its central role in AI research, especially outside of the geographic areas of current KR strength, now primarily Europe and, to a lesser degree, North America.
We believe, KR Inc.\ through its steering committee is in a position to develop initiatives aiming to address these goals.

Beyond some notable exceptions, such as answer set programming and description logics, a general issue concerns a lack of introductory or overview material for areas of KR:
the main KR textbook~\cite{BrachmanLevesque04} is nearly 20 years old, and the more technical Handbook of KR~\cite{DBLP:reference/fai/3} is over 15 years old.
A new textbook would be valuable as would an updated handbook, although this would require substantial effort and it is not clear where such effort would come from.
One option would be to develop a repository of KR materials as has been done for example for description logics.\footnote{\url{https://dl.kr.org/}}
Regardless, instructional materials should be developed introducing KR to undergraduate students and more advanced materials aimed at graduate students. The KR steering committee should promote undergraduate research projects in KR and showcase the best of them. It should encourage the KR community to develop tutorials and summer schools on KR topics that would bring the area of KR closer to students and young researchers outside of the regular academic curriculum. 
KR-themed tutorials at general AI conferences would be particularly well suited for reaching a broader audience.
When these programs are offered in person, KR Inc.\ should offer travel support to students and young researchers, especially from places with limited resources.

An in-person conference is unquestionably valuable, allowing one to meet and interact with researchers, students, and other practitioners, and as an incubator of new research and collaborations.
However, recent years have shown that virtual events can be effective in supporting research interactions and the exchange of ideas. 
These can be especially important in bringing KR to research communities facing economic barriers, and in reaching researchers from underrepresented groups and researchers with family conditions limiting their ability to travel. 
Tutorials, summer schools, panels, and seminar series are examples of events that lend themselves to virtual offerings. 
Further, even the KR conference might benefit by incorporating some online events into its program.
Most simply, conference presentations can be recorded and made generally available.
More radically the conference could switch to a virtual event every two or three years. This is something that can be investigated by the KR community through the KR steering committee.


\section{Conclusions}
\label{sec:conclusions}

This manifesto is based on the presentations, panels, working
groups, and discussions at the Dagstuhl Perspectives workshop 22282 ``Current and Future Challenges in Knowledge Representation and Reasoning''. 
To avoid an overly lengthy introduction, we deliberately kept our introduction in Section~\ref{sec:intro} short. Interested readers may, however, continue with the appendix, which covers the history of KR in more depth. The previous sections highlight that KR is a central, longstanding, and active area of Artificial Intelligence. While some sub-areas of KR have a long tradition, others, such as argumentation or hybrid systems, evolved more recently as core KR research areas. In Section~\ref{sec:areas}, we reviewed these different areas within and related to KR, presented an account of the state of the art as well as challenges within the sub-areas. Section~\ref{sec:challenges} then took a higher-level perspective and introduced general and major research challenges for KR as a whole. Finally, we concluded with recommendations for promoting the area.


\newpage

\begin{appendix}

\section{Appendix: History of Knowledge Representation and Reasoning}

Today it is broadly acknowledged that knowledge is a fundamental aspect of
intelligence.
This was arguably first stated by McCarthy and Hayes
in their seminal paper ``Some Philosophical Problems from the Standpoint of
AI~\cite{McCarthyHayes69}. They wrote there:
``\emph{[...] intelligence has two parts, which we shall call the \emph{epistemological} and the \emph{heuristic}. The epistemological part is the representation of the world in such a form that the solution of problems follows from the facts expressed in the representation. The heuristic part is the mechanism that on the basis of the information solves the problem and decides what to do.}''
The \emph{epistemological} component corresponds to what we understand today
as the problem of \emph{representation}, and the \emph{heuristic} part is the
\emph{reasoning} problem.
This stance was later restated by Brian Smith~\cite{Smith82} in the
\emph{Knowledge Representation Hypothesis}, discussed in the introduction.

It seems appropriate then to take the year 1969, the year when the paper by McCarthy and Hayes was published, as marking the birth of KR. That paper not only pinpointed the role of KR in AI but also identified many of the
key concerns of this area, all of which remain pertinent today. They are (quoting from the paper):
\emph{\begin{enumerate}
\item What kind of general representation of the world will allow the incorporation of specific observations and new scientific laws as they are discovered? 
\item Besides the representation of the physical world what other kinds of entities have to be provided for? For example, mathematical systems, goals, states of knowledge.
\item How are observations to be used to get knowledge about the world, and how are the other kinds of knowledge to be obtained? In particular what kinds of knowledge about the system’s own state of mind are to be provided for?
\item In what kind of internal notation is the system’s knowledge to be expressed?
\end{enumerate} }
While some earlier developments anticipated KR concerns, such as Newell and
Simon's \emph{General Problem Solver}~\cite{NewellS1956,NewellSS1957}, and
the development of the \emph{Resolution Principle} by Robinson~\cite{Robinson1965}, McCarthy and Hayes first explicitly stated the challenges defining the field.\footnote{
Other early work in the area is summarised in Brachman and Smith's 1980
survey~\cite{BrachmanSmith80} of KR.}
That said, research in KR as a mature area of AI is commonly taken as being marked by an
Artificial Intelligence Journal Special Issue on Non-monotonic Reasoning in
1980~\cite{Bobrow80}.
In 1989 the Principles of Knowledge Representation and Reasoning Conference
was founded, providing a dedicated, specialised forum for research in the
area\footnote{\url{https://kr.org/pastconfs.html}}.
Over the years KR has grown into a mature field of cross-disciplinary ideas,
with theoretical results and implemented systems inspired by foundational
concerns, as well as by practical considerations stemming from the needs of
concrete applications.
The primary focus was on developing languages for knowledge representation
based on first-order logic, sometimes extended by non-logical connectives
and modal operators and, in some cases, assigned a non-classical semantics.
However, driven to a large degree by the need for efficient reasoning, 
important early alternative approaches to knowledge representation emerged as well, most notably frames \cite{Minsky1974}, scripts \cite{SchankA1975}, production rule systems, and graphical formalisms such as conceptual graphs, inheritance networks, semantic networks, and Bayesian networks, the latter to model uncertainty. Today, knowledge representation languages rooted in these early proposals form the core of KR.


The choice of logic as a formalism for knowledge representation was not arbitrary.
We have already mentioned McCarthy and Hayes's 1969 paper;
as well a series of papers by McCarthy~\cite{McCarthy77,McCarthy79} and
Hayes~\cite{Hayes77,Hayes79,Hayes85,Hayes85b} implicitly or explicitly argued
for this position.
A large part of knowledge is declarative and consists of statements describing entities and relationships between them.
And incorporating new knowledge is often as simple as adding a new set of statements to the existing one.
The language of logic, first-order logic in particular, is well suited for modelling declarative knowledge.
Statements describing what is known can be modelled as logical formulas.
The Tarskian semantics of first-order logic provides a connection between a collection of formulas --- a body of knowledge about an application domain --- and structures, which could be interpreted as abstractions of that application domain with all its relevant entities and relationships.
First-order logic is modular and its statements can be arranged in hierarchies.
Further, the concept of a formal proof in first-order logic offers a way to derive new formulas from those given or established earlier. And logic already proved its mettle helping establish formal foundations for the rich and rigor-demanding edifice of mathematics.

Not surprisingly then, the influence of first-order logic can be seen in the \emph{Logic Theorist} by Newell, Simon and Shaw \cite{NewellS1956,NewellSS1957}, sometimes referred to as the first AI program, and the specification of the \emph{Advice Taker} by McCarthy \cite{McCarthy1959}. The discovery of the resolution proof method by Alan Robinson \cite{Robinson1965} and subsequent development of automated reasoning further strengthened the position of logic as the formalism of choice for KR. That being said, the early work on logic-based approaches to KR already pointed to limitations of first-order logic. First, as it was already known since G\"odel, first-order logic is undecidable. 
Moreover, even under restrictions on the language that result in
decidability, determining provability (or entailment) is often inherently hard.
Logic-based AI programs could serve as proofs of concept or as a theoretical
specification, but not as deployed systems working under time constraints. Second, some aspects of knowledge proved stubbornly difficult to formalise in first-order logic. In particular, commonsense reasoning, which McCarthy identified early on as a critical capability of intelligent systems \cite{McCarthy1959}, turned out to be a source of several challenging problems hard to capture within the first-order framework. Third, as argued by several researchers, Minsky most notable among them, logic and its automated reasoning machinery simply was not a good model of how the human brain works.

We will now consider these objections in more detail, as each led to major research directions within KR. The undecidability of first-order logic naturally pushed researchers to look for fragments of first-order logic in which effective reasoning was possible. Propositional logic is one such fragment. For its most commonly considered clausal version, resolution refutation was proved to be complete and sound. However, the problem of deciding whether a proof exists is coNP-complete while the dual problem of satisfiability is NP-complete. Thus, there is little hope for the existence of fast (polynomial-time bounded) algorithms for this problem. Nevertheless, because of impressive advances made in propositional satisfiability, the role of propositional logic in KR has been growing. It has left a strong mark among others in diagnosis, abductive reasoning, belief revision and update, as well as in planning. It has suggested or directly provided reasoning algorithms for answer-set programming, and spawned generalisations such as constraint satisfaction.

Production rules are another formalism proposed for KR that is closely
related to a fragment of first-order logic known as Horn logic. Because of
their restricted form, reasoning with production rules can be highly
efficient. They were used as the basis of expert systems, one of the most
successful class of AI programs built in the first golden era of AI in the 1970s and 1980s \cite{DavisBS1977}. This line of research started with expert systems such as MYCIN \cite{Shortliffe1976,BuchananS1984}, DENDRAL \cite{BuchananSF1969,BuchananF1981}, PROSPECTOR \cite{DudaGH1981}, and R1 (Xcon) \cite{JMcDermott80}, and evolved into building general purpose expert system shells. Important outcomes of this research include identifying knowledge acquisition as a major bottleneck and a challenge for KR, and establishing knowledge engineering as a subarea of KR.

Production rules emphasised the importance of the Horn fragment of first-order logic. Specialised resolution proof methods such as the SLD resolution \cite{KowalskiK1971} made it possible to think about logic as a declarative programming language \cite{Kowalski1974}. On the practical side, it led to the development of programming languages such as PLANNER \cite{Hewitt71} and PROLOG \cite{ColmerauerR1993}. On the theoretical side it opened the key question of the semantics of programs (especially programs with negation, which was allowed by the designers of those languages but only given a procedural interpretation), and helped establish the field of logic programming. Originally, logic programming and KR were developing side-by-side with few interactions between the two communities. Eventually, however, strong connections between the two fields were discovered. One of the outcomes of this coming together is answer set programming \cite{BrewkaET2011}, one of the most successful computational KR systems.

Another line of research emerged from the realization of a trade-off between the efficiency of reasoning and expressiveness of the language. It gave rise to an area in KR known as description logics \cite{BaaderHS2008}. Its origins can be traced to graphical models such as semantic networks \cite{Quillian1967} and structured languages such as KL-ONE \cite{BrachmanS1985}. However, it truly flourished once these earlier efforts were mapped onto fragments of first-order logic. This led to a careful and systematic analysis of the complexity and expressiveness and resulted in a broad range of description logics differing from each other in these respects. A major push behind the development of description logics came from the needs of the semantic web --- the world wide web with its content converted to machine interpretable formats. It resulted in the semantic web languages DAML, DAML+OIL \cite{Horrocks2002} and OWL \cite{McGuinnessVH2004}. Another important influence of description logics can be found in the areas of databases and information systems.

Finally, we note that much of the research in KR was driven by specialised
application domains.
One of the most influential ones concerns modelling and reasoning about
systems that evolve as a result of actions.
It was identified as  an area of interest in AI by McCarthy in a technical
report published by the Stanford University in 1963 \cite{McCarthy1963}.
In that report McCarthy introduced the \emph{situation calculus}, a first-order
logic language designed to model actions and their effects.
This language was further elaborated by McCarthy and Hayes~\cite{McCarthyHayes69} and
developed into a full theory by Reiter~\cite{Reiter01}.
The first two papers also introduced the key concerns to be addressed: in
particular, the famous \emph{frame problem} as well as two other fundamental
problems for KR, the \emph{ramification problem} and the \emph{qualification
problem}.
Addressing these problems, especially the frame problem, constituted major milestones in the development of KR. Elegant solutions were proposed not only for the original framework of the situation calculus but also in a related \emph{event calculus} \cite{KowalskiS1986}, and in non-classical formalisms for reasoning about action based on logic programming with the stable model semantics \cite{GelfondL1993}. 

Systems that change as actions are executed subsume planning problems.
Automated planning is as old as AI and some of the most influential early AI work concerns planning (for example, the STRIPS language \cite{FikesN1971} and the robot Shakey \cite{Nilsson1984,KuipersFHN2017}). Planning in general, and planning to address the needs of autonomous agents (robots), has been one of the core areas of AI. Many concerns in planning are those of representing knowledge about the world, and reasoning about that world as it changes. Hence, there are significant overlaps between KR and planning. Nevertheless, today planning and KR are two separate areas of AI.

The second major objection to first-order logic as a KR formalism stems from difficulties researchers discovered when attempting to use 
first-order logic to formalise commonsense reasoning. Humans can naturally handle rules with exceptions, reason without full information, represent and use normative statements, revise or update their knowledge (belief) base when new information becomes available, they can reason qualitatively about physical phenomena in the context of space and time, and they can represent and reason about uncertainty. Not surprisingly, since the beginning of AI, researchers argued that formalising and automating common sense is necessary for artificial intelligence. Yet, doing so within the confines of first-order logic proved a difficult task. This realization motivated several subareas of KR.

The types of reasoning mentioned above often involve certain non-logical reasoning principles such as adopting wherever possible a view that the world about which we reason is ``normal,'' avoiding making unnecessary assumptions when we do not know whether a property holds or not, or making assumptions based on their likelihood. These non-logical principles make it possible that valid inferences may become invalid in view of new information the reasoner may gain. This is clearly not the case for inferences supported by first-order logic.
The need then arose for building these non-logical principles into an
inference mechanism in a logical formalism. 

The first such principle was the \emph{closed world assumption} identified
and studied by Reiter \cite{Reiter1978}.
Further developments came in 1980, when McCarthy introduced
\emph{circumscription} \cite{McCarthy80}, which incorporated the principle of reasoning
in the context of minimal models only, 
Reiter introduced default logic \cite{Reiter80}, which integrated non-classical inference rules
(defaults) conditioned upon what could reasonably be assumed as possible,
while McDermott and Doyle introduced a modal logic allowing for introspection
\cite{McDermottD1980}.
This last approach was found lacking.
However, soon thereafter improved proposals for non-standard reasoning with
modal theories were made by McDermott \cite{McDermott1982}, who introduced a single fixpoint
principle to capture non-classical inferences for all normal modal logics,
and Moore, who proposed \emph{autoepistemic logic} \cite{Moore1985}.
Circumscription, default logic and modal logics of McDermott, Doyle and Moore
gave rise to the area of non-monotonic logics that  soon became one of the
main areas in KR.
A significant research effort expanded in that area explained the
relationships between different non-monotonic logics, and established a
comprehensive picture of their expressive power and complexity \cite{MarekT1993}.
It also demonstrated the effectiveness of non-monotonic logics in modelling abductive and diagnostic reasoning tasks \cite{deKleerEA1992}. An important contribution of default logic is the semantics it suggested for logic programs with negation. That semantics, known as the stable-model semantics, has been broadly accepted and is the foundation of answer-set programming mentioned above. 

A different attempt to provide an overview of non-monotonic formalisms was provided by Kraus, Lehmann, and Magidor \cite{KrausLehmannMagidor90}. In what has become known as the \emph{KLM approach} to non-monotonic reasoning, the authors proposed sets of axioms against which any non-monotonic consequence relation can be tested. Although this work was initially focused on non-monotonic consequence for classical propositional logic, it was subsequently extended to study non-monotonic consequence over conditional propositional logics \cite{LehmannMagidor1992}. This extension has given rise to a flurry of activity in designing non-monotonic reasoning systems for versions of conditional logics going beyond propositional logic.  

While logic was the most extensively studied framework to capture non-monotonic reasoning, a significant effort was expanded in 1980s to extend semantic networks to handle exceptions and defaults. This research effort evolved into the area of inheritance networks \cite{Touretzky1986}. As inheritance networks could be mapped into (non-monotonic) logics, this line of research gradually waned. It may receive soon renewed attention with the growing role of knowledge graphs \cite{JiPCMY2022} in KR. 

Human commonsense reasoning allows us to gracefully update or revise our knowledge and beliefs once new information becomes available. Formalising this capability is essential for intelligent behavior in a changing world, when a knowledge (belief) base has to be updated to reflect a new situation, or revised, when the situation does not change but ``better'' (more accurate) information is gained. The field that grew out of these considerations is called belief revision but it includes also considerations that are more accurately referred to as belief update. The major milestone for the area is the paper by Alchourr\'on, G\"ardenfors and Makinson \cite{AGM1985}. That paper proposed a set of postulates, now known as the AGM postulates, that any belief revision operator should satisfy. It was a way to address the unsettling realization that belief revision could not be reduced to a single revision operator. In fact, numerous operators were proposed that, arguably, adequately handled the task of revising a knowledge base by a formula representing a new piece of information. The paper was followed by papers characterizing, often in a constructive way, classes of operators satisfying all (or some) of the AGM postulates. In another interesting development, G{\"a}rdenfors and Makinson showed that KLM-style non-monotonic reasoning and AGM belief revision are two sides of the same coin \cite{GardenforsMakinson1994}. Informally, the link between the two formalisms is easy to explain: B is a non-monotonic consequence of A whenever B is a classical consequence of the knowledge base obtained from a revision of A.

Additional concerns addressed or partially addressed over the years included iterated revision, a major extension of the original framework that focused on a single act of revising \cite{Spohn88,DarwichePearl1997}, and a more general problem of knowledge merging \cite{Konieczny2000}. The field of belief update was identified as separate from belief revision by Katsuno and
Mendelzon~\cite{KatsunoMendelzon92}, who started it by proposing a set of postulates for belief update operators. In many respects its development followed that of belief revision. 
	
Qualitative reasoning is another type of commonsense reasoning humans are good at. It concerns reasoning about ``continuous'' domains such as space and time. While phenomena in such domains are most effectively modelled by systems of (partial) differential equations, few of us maintain such systems in our mind and solve them when the need arises. Instead, we have developed ``commonsense'' representations of such domains that allow us to reason efficiently about time, space, motion and actions that might affect them, and with sufficient accuracy to function \cite{Forbus2008,Davis2008}. The main thrust of this work is to develop ontologies of relevant concepts and develop reasoning techniques based on these ontologies. In this way, the area can be mapped onto (a fragment) of first-order logic and can be developed exploiting logical means. Two important subareas of qualitative reasoning focus on spatial \cite{CohnR2008} and temporal \cite{Fisher2008} reasoning, respectively.

Much of what we discussed falls firmly in the domain of qualitative approaches based on the language of logic. However, quantitative approaches also played a major role in KR research. One of the most prominent examples is the formalism of \emph{Bayesian networks} \cite{Pearl1985,Pearl1988,Darwiche2008}. 
A Bayesian network consists of a directed acyclic graph representing variables of the problem as nodes and direct influences among variables as edges, and of a collection of conditional probability tables associated with each variable. The power of Bayesian networks comes from an observation that valid inferences about probabilities of events or contributions of certain events to cause events that occurred can be derived without constructing explicitly joint distribution functions on all events involved. To make Bayesian networks an effective KR tool, researchers developed methods to represent Bayesian networks, to compile them into arithmetic circuits and propositional formulas, as well as algorithms to perform exact and approximate inference. Another key research theme concerned methods to construct Bayesian networks by standard knowledge engineering methods involving domain experts, by deriving them from specifications exploring their structural regularity by learning. Today, Bayesian networks form a dominant tool for reasoning with uncertain and incomplete information. An important aspect of Bayesian networks has been their ability to model causality.

We conclude this overview by noting again that not all knowledge is declarative. Much of what we know is procedural or algorithmic, and stored as such procedures in our brains. Using this knowledge involves little if any reasoning, just recalling from memory the right procedure for the occasion. Knowledge needed to accomplish routine, frequently repeated tasks, in particular, scene and speech understanding fall into this category. The ways human brains act to accomplish these tasks are not fully understood. However, they inspired the so called \emph{connectionist} models, best known as neural networks, as a possible framework for not only representing algorithmic knowledge but also for learning these representations from observations (examples). This approach has been outside of the scope of KR for much of its history. This is beginning to change. On the one hand, machine learning and neural network representations form that part of AI that admittedly ignited imagination of both research communities and the community at large. It is now by far the most vibrant branch of AI dwarfing all others. On the other hand, it is not by itself sufficient to reach the point when machines will act intelligently in ways humans do. For that these systems need knowledge and it is now evident that further progress in AI depends on successful integration of machine learning and knowledge representation.

\end{appendix}

\begin{participants}
\participant  Michael Beetz\\ Universit\"{a}t Bremen, DE
\participant  Meghyn Bienvenu\\ University of Bordeaux, FR
\participant  Piero Andrea Bonatti\\ University of Naples, IT
\participant  Diego Calvanese\\ Free University of Bozen-Bolzano, IT
\participant  Anthony Cohn\\ University of Leeds, GB
\participant  James P. Delgrande\\ Simon Fraser University – Burnaby, CA
\participant  Marc Denecker\\ KU Leuven, BE
\participant  Thomas Eiter\\ TU Wien, AT
\participant  Esra Erdem\\ Sabanci University – Istanbul, TR
\participant  Birte Glimm\\ Universit\"{a}t Ulm, DE
\participant  Andreas Herzig\\ Paul Sabatier University – Toulouse, FR
\participant  Ian Horrocks\\ University of Oxford, GB
\participant  Jean Christoph Jung\\ Universit\"{a}t Hildesheim, DE
\participant  Sebastien Konieczny\\ University of Artois/CNRS – Lens, FR
\participant  Gerhard Lakemeyer\\ RWTH Aachen University, DE
\participant  Thomas Meyer\\ University of Cape Town, ZA
\participant  Magdalena Ortiz\\ TU Wien, AT
\participant  Ana Ozaki\\ University of Bergen, NO
\participant  Milene Santos Teixeira\\ Universit\"{a}t Ulm, DE
\participant  Torsten Schaub\\ Universit\"{a}t Potsdam, DE
\participant  Steven Schockaert\\ Cardiff University, GB
\participant  Michael Thielscher\\ UNSW – Sydney, AU
\participant  Francesca Toni\\ Imperial College London, GB
\participant  Renata Wassermann\\ University of Sao Paulo, BR
\participant  Frank Wolter\\ University of Liverpool, GB
\end{participants}

\section{Acknowledgements}

We thank the Dagstuhl staff for their outstanding administrative and operational support in the organisation and running of the workshop.
We thank the participants of the workshop, without whose involvement this manifesto, of course, would not be possible.
We also thank those individuals who helped us with parts of this manifesto:
Vaishak Belle (Reasoning under Uncertainty),
Diego Calvanese (Information Systems),
Jens Cla{\ss}en (Reasoning about Action),
Giuseppe de Giacomo (Reasoning about Action),
Wolfgang Dvo\v{r}\'ak (Argumentation),
Anthony Hunter (Argumentation),
Gerhard Lakemeyer (KR and Robotics),
Marco Montali (Information Systems),
Ana Ozaki (KR and ML), 
Francesco Ricca (ASP),
Steven Schokaert (KR and ML),
Francesca Toni (Argumentation),
Johannes Wallner (Argumentation),
Stefan Woltran (Argumentation).

\begin{thebibliography}{100}
	
	\bibitem{potassco}
	\emph{Potassco} project, 2011+.
	\newblock \url{https://github.com/potassco/}, \url{https://potassco.org}.
	
	\bibitem{wasp}
	\emph{Wasp} project, 2013+.
	\newblock \url{http://alviano.github.io/wasp/}.
	
	\bibitem{dlv}
	\emph{Dlv} project, 2017+.
	\newblock \url{https://dlv.demacs.unical.it/home}.
	
	\bibitem{DBLP:conf/aaai/AbboudCL20}
	Ralph Abboud, {\.I}smail~{\.I}lkan Ceylan, and Thomas Lukasiewicz.
	\newblock Learning to reason: Leveraging neural networks for approximate {DNF}
	counting.
	\newblock In {\em The Thirty-Fourth {AAAI} Conference on Artificial
		Intelligence, {AAAI} 2020, The Thirty-Second Innovative Applications of
		Artificial Intelligence Conference, {IAAI} 2020, The Tenth {AAAI} Symposium
		on Educational Advances in Artificial Intelligence, {EAAI} 2020, New York,
		NY, USA, February 7-12, 2020}, pages 3097--3104. {AAAI} Press, 2020.
	
	\bibitem{AbelsJOSTW2021}
	Dirk Abels, Julian Jordi, Max Ostrowski, Torsten Schaub, Ambra Toletti, and
	Philipp Wanko.
	\newblock Train scheduling with hybrid answer set programming.
	\newblock {\em Theory Pract. Log. Program.}, 21(3):317--347, 2021.
	
	\bibitem{DBLP:journals/access/AlamRNMAVL22}
	Mirza~Mohtashim Alam, Md. Rashad Al~Hasan Rony, Mojtaba Nayyeri, Karishma
	Mohiuddin, M.~S. T.~Mahfuja Akter, Sahar Vahdati, and Jens Lehmann.
	\newblock Language model guided knowledge graph embeddings.
	\newblock {\em {IEEE} Access}, 10:76008--76020, 2022.
	
	\bibitem{AGM1985}
	Carlos~E Alchourr{\'o}n, Peter G{\"a}rdenfors, and David Makinson.
	\newblock {On the Logic of Theory Change: Partial Meet Contraction and Revision
		Functions}.
	\newblock {\em The Journal of Symbolic Logic}, 50(2):510--530, 1985.
	
	\bibitem{AlvianoFG2021}
	Mario Alviano, Wolfgang Faber, and Martin Gebser.
	\newblock Aggregate semantics for propositional answer set programs.
	\newblock {\em CoRR}, abs/2109.08662, 2021.
	
	\bibitem{DBLP:conf/www/AndreselCORSS20}
	Medina Andresel, Julien Corman, Magdalena Ortiz, Juan~L. Reutter, Ognjen
	Savkovic, and Mantas Simkus.
	\newblock Stable model semantics for recursive {SHACL}.
	\newblock In {\em {WWW} '20: The Web Conference 2020, Taipei, Taiwan, April
		20-24, 2020}, pages 1570--1580, 2020.
	
	\bibitem{DBLP:journals/csur/AnglesABHRV17}
	Renzo Angles, Marcelo Arenas, Pablo Barcel{\'{o}}, Aidan Hogan, Juan~L.
	Reutter, and Domagoj Vrgoc.
	\newblock Foundations of modern query languages for graph databases.
	\newblock {\em {ACM} Comput. Surv.}, 50(5):68:1--68:40, 2017.
	
	\bibitem{AntakiL1992}
	Charles Antaki and Ivan Leudar.
	\newblock Explaining in conversation: Towards an argument model.
	\newblock {\em European Journal of Social Psychology}, 22(2):181--194, 1992.
	
	\bibitem{DBLP:books/cu/ArenasBLM2014}
	Marcelo Arenas, Pablo Barcel{\'{o}}, Leonid Libkin, and Filip Murlak.
	\newblock {\em Foundations of Data Exchange}.
	\newblock Cambridge University Press, 2014.
	
	\bibitem{DBLP:conf/www/ArenasDK16}
	Marcelo Arenas, Gonzalo~I. Diaz, and Egor~V. Kostylev.
	\newblock Reverse engineering {SPARQL} queries.
	\newblock In Jacqueline Bourdeau, Jim Hendler, Roger Nkambou, Ian Horrocks, and
	Ben~Y. Zhao, editors, {\em Proceedings of the 25th International Conference
		on World Wide Web, {WWW} 2016, Montreal, Canada, April 11 - 15, 2016}, pages
	239--249. {ACM}, 2016.
	
	\bibitem{DBLP:conf/birthday/ArtaleF09}
	Alessandro Artale and Enrico Franconi.
	\newblock Foundations of temporal conceptual data models.
	\newblock In Alexander Borgida, Vinay~K. Chaudhri, Paolo Giorgini, and Eric
	S.~K. Yu, editors, {\em Conceptual Modeling: Foundations and Applications -
		Essays in Honor of John Mylopoulos}, volume 5600 of {\em Lecture Notes in
		Computer Science}, pages 10--35. Springer, 2009.
	
	\bibitem{DBLP:journals/jair/ArtaleKKRWZ22}
	Alessandro Artale, Roman Kontchakov, Alisa Kovtunova, Vladislav Ryzhikov, Frank
	Wolter, and Michael Zakharyaschev.
	\newblock First-order rewritability and complexity of two-dimensional temporal
	ontology-mediated queries.
	\newblock {\em J. Artif. Intell. Res.}, 75:1223--1291, 2022.
	
	\bibitem{DBLP:conf/ijcai/AsaiM20}
	Masataro Asai and Christian Muise.
	\newblock Learning neural-symbolic descriptive planning models via cube-space
	priors: The voyage home (to {STRIPS)}.
	\newblock In {\em Proceedings of the Twenty-Ninth International Joint
		Conference on Artificial Intelligence, {IJCAI} 2020}, pages 2676--2682.
	ijcai.org, 2020.
	
	\bibitem{AtkinsonB21}
	Katie Atkinson and Trevor J~M Bench{-}Capon.
	\newblock Argumentation schemes in {AI} and law.
	\newblock {\em Argument Comput.}, 12(3):417--434, 2021.
	
	\bibitem{DBLP:conf/iclr/AwasthiGGS20}
	Abhijeet Awasthi, Sabyasachi Ghosh, Rasna Goyal, and Sunita Sarawagi.
	\newblock Learning from rules generalizing labeled exemplars.
	\newblock In {\em 8th International Conference on Learning Representations,
		{ICLR} 2020, Addis Ababa, Ethiopia, April 26-30, 2020}. OpenReview.net, 2020.
	
	\bibitem{BaaderEtAl07}
	F.~Baader, D.~Calvanese, D.~McGuinness, D.~Nardi, and P~Patel-Schneider,
	editors.
	\newblock {\em The Description Logic Handbook}.
	\newblock Cambridge University Press, Cambridge, second edition, 2007.
	
	\bibitem{DBLP:journals/tocl/BaaderGL12}
	Franz Baader, Silvio Ghilardi, and Carsten Lutz.
	\newblock {LTL} over description logic axioms.
	\newblock {\em {ACM} Trans. Comput. Log.}, 13(3):21:1--21:32, 2012.
	
	\bibitem{BaaderHS2008}
	Franz Baader, Ian Horrocks, and Ulrike Sattler.
	\newblock Description logics.
	\newblock In van Harmelen et~al. \cite{DBLP:reference/fai/3}, pages 135--179.
	
	\bibitem{DBLP:conf/kr/BaaderKNP18}
	Franz Baader, Francesco Kriegel, Adrian Nuradiansyah, and Rafael
	Pe{\~{n}}aloza.
	\newblock Making repairs in description logics more gentle.
	\newblock In Michael Thielscher, Francesca Toni, and Frank Wolter, editors,
	{\em Principles of Knowledge Representation and Reasoning: Proceedings of the
		Sixteenth International Conference, {KR} 2018, Tempe, Arizona, 30 October - 2
		November 2018}, pages 319--328. {AAAI} Press, 2018.
	
	\bibitem{DBLP:conf/emnlp/BalazevicAH19}
	Ivana Balazevic, Carl Allen, and Timothy~M. Hospedales.
	\newblock Tucker: Tensor factorization for knowledge graph completion.
	\newblock In Kentaro Inui, Jing Jiang, Vincent Ng, and Xiaojun Wan, editors,
	{\em Proceedings of the 2019 Conference on Empirical Methods in Natural
		Language Processing and the 9th International Joint Conference on Natural
		Language Processing, {EMNLP-IJCNLP} 2019, Hong Kong, China, November 3-7,
		2019}, pages 5184--5193. Association for Computational Linguistics, 2019.
	
	\bibitem{BanbaraIKOSSTW2019}
	Mutsunori Banbara, Katsumi Inoue, Benjamin Kaufmann, Tenda Okimoto, Torsten
	Schaub, Takehide Soh, Naoyuki Tamura, and Philipp Wanko.
	\newblock teaspoon: solving the curriculum-based course timetabling problems
	with answer set programming.
	\newblock {\em Ann. Oper. Res.}, 275(1):3--37, 2019.
	
	\bibitem{DBLP:conf/ijcai/BanihashemiGL18}
	Bita Banihashemi, Giuseppe~De Giacomo, and Yves Lesp{\'{e}}rance.
	\newblock Abstraction of agents executing online and their abilities in the
	situation calculus.
	\newblock In J{\'{e}}r{\^{o}}me Lang, editor, {\em Proceedings of the
		Twenty-Seventh International Joint Conference on Artificial Intelligence,
		{IJCAI} 2018, July 13-19, 2018, Stockholm, Sweden}, pages 1699--1706.
	ijcai.org, 2018.
	
	\bibitem{Belle2017}
	Vaishak Belle.
	\newblock Logic meets probability: Towards explainable {AI} systems for
	uncertain worlds.
	\newblock In {\em Proceedings of the Twenty-Sixth International Joint
		Conference on Artificial Intelligence, {IJCAI-17}}, pages 5116--5120, 2017.
	
	\bibitem{Belle2022}
	Vaishak Belle.
	\newblock Logic meets learning: From aristotle to neural networks.
	\newblock In Pascal Hitzler and {Md Kamruzzaman } Sarker, editors, {\em
		Neuro-Symbolic Artificial Intelligence - The State of the Art}, Frontiers in
	Artificial Intelligence and Applications, pages 78 -- 102. IOS Press,
	Netherlands, January 2022.
	
	\bibitem{DBLP:journals/pvldb/BellomariniSG18}
	Luigi Bellomarini, Emanuel Sallinger, and Georg Gottlob.
	\newblock The vadalog system: Datalog-based reasoning for knowledge graphs.
	\newblock {\em Proc. {VLDB} Endow.}, 11(9):975--987, 2018.
	
	\bibitem{benedikt_et_al:DR:2020:11842}
	Michael Benedikt, Kristian Kersting, Phokion~G. Kolaitis, and Daniel Neider.
	\newblock {Logic and Learning (Dagstuhl Seminar 19361)}.
	\newblock {\em Dagstuhl Reports}, 9(9):1--22, 2020.
	
	\bibitem{BesnardH2008}
	Philippe Besnard and Anthony Hunter.
	\newblock {\em Elements of argumentation}, volume~47.
	\newblock MIT press Cambridge, 2008.
	
	\bibitem{BesnardH2018}
	Philippe Besnard and Anthony Hunter.
	\newblock A review of argumentation based on deductive arguments.
	\newblock In Pietro Baroni, Dov Gabbay, Massimilino Giacomin, and Leendert
	Van~der Torre, editors, {\em {Handbook of Formal Argumentation}}, pages
	437--484. {College Publications}, 2018.
	
	\bibitem{BidoitF87}
	N.~Bidoit and C.~Froidevaux.
	\newblock Minimalism subsumes default logic and circumscription.
	\newblock In {\em Proceedings of IEEE Symposium on Logic in Computer Science,
		LICS-87}, pages 89--97. IEEE Press, 1987.
	
	\bibitem{BidoitF91}
	N.~Bidoit and C.~Froidevaux.
	\newblock Negation by default and unstratifiable logic programs.
	\newblock {\em Theoretical Computer Science}, 78(1, (Part B)):85--112, 1991.
	
	\bibitem{DBLP:journals/ki/Bienvenu20}
	Meghyn Bienvenu.
	\newblock A short survey on inconsistency handling in ontology-mediated query
	answering.
	\newblock {\em K{\"{u}}nstliche Intell.}, 34(4):443--451, 2020.
	
	\bibitem{DBLP:conf/rweb/BienvenuO15}
	Meghyn Bienvenu and Magdalena Ortiz.
	\newblock Ontology-mediated query answering with data-tractable description
	logics.
	\newblock In Wolfgang Faber and Adrian Paschke, editors, {\em Reasoning Web.
		Web Logic Rules - 11th International Summer School 2015, Berlin, Germany,
		July 31 - August 4, 2015, Tutorial Lectures}, volume 9203 of {\em Lecture
		Notes in Computer Science}, pages 218--307. Springer, 2015.
	
	\bibitem{DBLP:journals/tods/BienvenuCLW14}
	Meghyn Bienvenu, Balder ten Cate, Carsten Lutz, and Frank Wolter.
	\newblock {Ontology-Based Data Access: A Study through Disjunctive Datalog,
		CSP, and MMSNP}.
	\newblock {\em {ACM} Trans. Database Syst.}, 39(4):33:1--33:44, 2014.
	
	\bibitem{Bobrow80}
	D.~Bobrow, editor.
	\newblock {\em Special Issue on Nonmonotonic Reasoning}, volume 13(1-2).
	\newblock Elsevier, 1980.
	
	\bibitem{DBLP:conf/kr/BolanderDH21}
	Thomas Bolander, Lasse Dissing, and Nicolai Herrmann.
	\newblock Del-based epistemic planning for human-robot collaboration: Theory
	and implementation.
	\newblock In Meghyn Bienvenu, Gerhard Lakemeyer, and Esra Erdem, editors, {\em
		Proceedings of the 18th International Conference on Principles of Knowledge
		Representation and Reasoning, {KR} 2021, Online event, November 3-12, 2021},
	pages 120--129, 2021.
	
	\bibitem{DBLP:journals/computing/BoltenhagenCC21}
	Mathilde Boltenhagen, Thomas Chatain, and Josep Carmona.
	\newblock Optimized {SAT} encoding of conformance checking artefacts.
	\newblock {\em Computing}, 103(1):29--50, 2021.
	
	\bibitem{Bonatti2019}
	Piero~A. Bonatti.
	\newblock {Rational closure for all description logics}.
	\newblock {\em Artificial Intelligence}, 274:197--223, 2019.
	
	\bibitem{DBLP:series/ssw/BonattiPS20}
	Piero~Andrea Bonatti, Iliana~Mineva Petrova, and Luigi Sauro.
	\newblock Defeasible reasoning in description logics: An overview on {DLN}.
	\newblock In Giuseppe Cota, Marilena Daquino, and Gian~Luca Pozzato, editors,
	{\em Applications and Practices in Ontology Design, Extraction, and
		Reasoning}, volume~49 of {\em Studies on the Semantic Web}, pages 178--193.
	{IOS} Press, 2020.
	
	\bibitem{BondarenkoDKT1997}
	Andrei Bondarenko, Phan~Minh Dung, Robert~A Kowalski, and Francesca Toni.
	\newblock An abstract, argumentation-theoretic approach to default reasoning.
	\newblock {\em Artificial intelligence}, 93(1-2):63--101, 1997.
	
	\bibitem{DBLP:journals/jair/BoothM06}
	Richard Booth and Thomas~Andreas Meyer.
	\newblock Admissible and restrained revision.
	\newblock {\em J. Artif. Intell. Res.}, 26:127--151, 2006.
	
	\bibitem{DBLP:journals/jair/BoothMVW11}
	Richard Booth, Thomas~Andreas Meyer, Ivan~Jos{\'{e}} Varzinczak, and Renata
	Wassermann.
	\newblock On the link between partial meet, kernel, and infra contraction and
	its application to horn logic.
	\newblock {\em J. Artif. Intell. Res.}, 42:31--53, 2011.
	
	\bibitem{DBLP:conf/nips/BordesUGWY13}
	Antoine Bordes, Nicolas Usunier, Alberto Garc{\'{\i}}a{-}Dur{\'{a}}n, Jason
	Weston, and Oksana Yakhnenko.
	\newblock Translating embeddings for modeling multi-relational data.
	\newblock In Christopher J.~C. Burges, L{\'{e}}on Bottou, Zoubin Ghahramani,
	and Kilian~Q. Weinberger, editors, {\em Advances in Neural Information
		Processing Systems 26: 27th Annual Conference on Neural Information
		Processing Systems 2013. Proceedings of a meeting held December 5-8, 2013,
		Lake Tahoe, Nevada, United States}, pages 2787--2795, 2013.
	
	\bibitem{DBLP:conf/aaai/Borgwardt0KKNS22}
	Stefan Borgwardt, J{\"{o}}rg Hoffmann, Alisa Kovtunova, Markus Kr{\"{o}}tzsch,
	Bernhard Nebel, and Marcel Steinmetz.
	\newblock Expressivity of planning with horn description logic ontologies.
	\newblock In {\em Thirty-Sixth {AAAI} Conference on Artificial Intelligence,
		{AAAI} 2022, Thirty-Fourth Conference on Innovative Applications of
		Artificial Intelligence, {IAAI} 2022, The Twelveth Symposium on Educational
		Advances in Artificial Intelligence, {EAAI} 2022 Virtual Event, February 22 -
		March 1, 2022}, pages 5503--5511. {AAAI} Press, 2022.
	
	\bibitem{DBLP:conf/rweb/BotoevaKLRWZ16}
	Elena Botoeva, Boris Konev, Carsten Lutz, Vladislav Ryzhikov, Frank Wolter, and
	Michael Zakharyaschev.
	\newblock Inseparability and conservative extensions of description logic
	ontologies: {A} survey.
	\newblock In Jeff~Z. Pan, Diego Calvanese, Thomas Eiter, Ian Horrocks, Michael
	Kifer, Fangzhen Lin, and Yuting Zhao, editors, {\em Reasoning Web: Logical
		Foundation of Knowledge Graph Construction and Query Answering - 12th
		International Summer School 2016, Aberdeen, UK, September 5-9, 2016, Tutorial
		Lectures}, volume 9885 of {\em Lecture Notes in Computer Science}, pages
	27--89. Springer, 2016.
	
	\bibitem{BrachmanLevesque04}
	R.J. Brachman and H.J. Levesque.
	\newblock {\em Knowledge Representation and Reasoning}.
	\newblock Morgan-Kaufmann, 2004.
	
	\bibitem{BrachmanLevesque22}
	R.J. Brachman and H.J. Levesque.
	\newblock {\em Machines Like Us: Toward {AI} with Common Sense}.
	\newblock The MIT Press, Cambridge, MA, 2022.
	
	\bibitem{BrachmanSmith80}
	R.J. Brachman and B.C. Smith.
	\newblock Special issue on knowledge representation.
	\newblock {\em SIGART Newsletter}, 70, 1980.
	
	\bibitem{BrachmanS1985}
	Ronald~J. Brachman and James~G. Schmolze.
	\newblock {An Overview of the KL-ONE Knowledge Representation System}.
	\newblock {\em Cognitive Science}, 9:171--216, 1985.
	
	\bibitem{DBLP:journals/aamas/BrennerN09}
	Michael Brenner and Bernhard Nebel.
	\newblock Continual planning and acting in dynamic multiagent environments.
	\newblock {\em Auton. Agents Multi Agent Syst.}, 19(3):297--331, 2009.
	
	\bibitem{BrewkaEtAl08}
	G.~Brewka, I.~Niemel{\"a}, and M.~Truszczynski.
	\newblock Nonmonotonic reasoning.
	\newblock In F.~{van Harmelen}, V.~Lifschitz, and B.~Porter, editors, {\em
		Handbook of Knowledge Representation}, pages 239--284. Elsevier Science, San
	Diego, USA, 2008.
	
	\bibitem{BrewkaET2011}
	Gerhard Brewka, Thomas Eiter, and Miros{\l}aw Truszczy{\'n}ski.
	\newblock {Answer Set Programming at a Glance}.
	\newblock {\em Communications of the ACM}, 54(12):92--103, 2011.
	
	\bibitem{BrewkaET2016}
	Gerhard Brewka, Thomas Eiter, and Miroslaw Truszczynski.
	\newblock Answer set programming: An introduction to the special issue.
	\newblock {\em {AI} Mag.}, 37(3):5--6, 2016.
	
	\bibitem{BritzEtAL2020}
	Katarina Britz, Giovanni Casini, Thomas Meyer, Kody Moodley, Uli Sattler, and
	Ivan Varzinczak.
	\newblock Principles of {KLM}-style defeasible description logics.
	\newblock {\em ACM Transactions on Computational Logic}, 22(1), 2020.
	
	\bibitem{BuchananF1981}
	Bruce Buchanan and Edward~A. Feigenbaum.
	\newblock {Dendral and Meta-dendral: Their Applications Dimension}.
	\newblock In {\em Readings in artificial intelligence}, pages 313--322.
	Elsevier, 1981.
	
	\bibitem{BuchananS1984}
	Bruce Buchanan and Edward Shortliffe.
	\newblock {\em {Rule-Based Expert Systems: The MYCIN Experiments of the
			Stanford Heuristic Programming Project}}.
	\newblock Addison Wesley, 1984.
	
	\bibitem{BuchananSF1969}
	Bruce Buchanan, Georgia Sutherland, and Edward A.Feigenbaum.
	\newblock {Heuristic DENDRAL: A Program for Generating Explanatory Hypotheses}.
	\newblock {\em Organic Chemistry}, 1969.
	
	\bibitem{Rhino99}
	Wolfram Burgard, Armin~B. Cremers, Dieter Fox, Dirk H{\"{a}}hnel, Gerhard
	Lakemeyer, Dirk Schulz, Walter Steiner, and Sebastian Thrun.
	\newblock Experiences with an interactive museum tour-guide robot.
	\newblock {\em Artif. Intell.}, 114(1-2):3--55, 1999.
	
	\bibitem{BusoniuOPST2013}
	Paula{-}Andra Busoniu, Johannes Oetsch, J{\"{o}}rg P{\"{u}}hrer, Peter
	Skocovsky, and Hans Tompits.
	\newblock Sealion: An eclipse-based {IDE} for answer-set programming with
	advanced debugging support.
	\newblock {\em Theory Pract. Log. Program.}, 13(4-5):657--673, 2013.
	
	\bibitem{DBLP:conf/lpnmr/CabalarDSS22}
	Pedro Cabalar, Mart{\'{\i}}n Di{\'{e}}guez, Torsten Schaub, and Anna Schuhmann.
	\newblock Metric temporal answer set programming over timed traces.
	\newblock In Georg Gottlob, Daniela Inclezan, and Marco Maratea, editors, {\em
		Logic Programming and Nonmonotonic Reasoning - 16th International Conference,
		{LPNMR} 2022, Genova, Italy, September 5-9, 2022, Proceedings}, volume 13416
	of {\em Lecture Notes in Computer Science}, pages 117--130. Springer, 2022.
	
	\bibitem{DBLP:journals/ai/CaliGP12}
	Andrea Cal{\`{\i}}, Georg Gottlob, and Andreas Pieris.
	\newblock Towards more expressive ontology languages: The query answering
	problem.
	\newblock {\em Artif. Intell.}, 193:87--128, 2012.
	
	\bibitem{CalimeriFGIKKLM2020}
	Francesco Calimeri, Wolfgang Faber, Martin Gebser, Giovambattista Ianni, Roland
	Kaminski, Thomas Krennwallner, Nicola Leone, Marco Maratea, Francesco Ricca,
	and Torsten Schaub.
	\newblock {ASP}-{C}ore-2 input language format.
	\newblock {\em Theory Pract. Log. Program.}, 20(2):294--309, 2020.
	
	\bibitem{CalimeriFPZ16}
	Francesco Calimeri, Davide Fusc{\`{a}}, Simona Perri, and Jessica Zangari.
	\newblock \emph{I}-{DLV}: The new intelligent grounder of {DLV}.
	\newblock In Giovanni Adorni, Stefano Cagnoni, Marco Gori, and Marco Maratea,
	editors, {\em AI*IA 2016: XVth International Conference of the Italian
		Association for Artificial Intelligence}, volume 10037 of {\em Lecture Notes
		in Computer Science}, pages 192--207. Springer, 2016.
	
	\bibitem{DBLP:journals/jar/CalvaneseGLLR07}
	Diego Calvanese, Giuseppe~De Giacomo, Domenico Lembo, Maurizio Lenzerini, and
	Riccardo Rosati.
	\newblock Tractable reasoning and efficient query answering in description
	logics: The \emph{DL-Lite} family.
	\newblock {\em J. Autom. Reason.}, 39(3):385--429, 2007.
	
	\bibitem{DBLP:journals/ai/CalvaneseGLLR13}
	Diego Calvanese, Giuseppe~De Giacomo, Domenico Lembo, Maurizio Lenzerini, and
	Riccardo Rosati.
	\newblock Data complexity of query answering in description logics.
	\newblock {\em Artif. Intell.}, 195:335--360, 2013.
	
	\bibitem{DBLP:conf/pods/CalvaneseGM13}
	Diego Calvanese, Giuseppe~De Giacomo, and Marco Montali.
	\newblock Foundations of data-aware process analysis: a database theory
	perspective.
	\newblock In Richard Hull and Wenfei Fan, editors, {\em Proceedings of the 32nd
		{ACM} {SIGMOD-SIGACT-SIGART} Symposium on Principles of Database Systems,
		{PODS} 2013, New York, NY, {USA} - June 22 - 27, 2013}, pages 1--12. {ACM},
	2013.
	
	\bibitem{DBLP:journals/iandc/CalvaneseGMP18}
	Diego Calvanese, Giuseppe~De Giacomo, Marco Montali, and Fabio Patrizi.
	\newblock First-order \emph{{\(\mu\)}}-calculus over generic transition systems
	and applications to the situation calculus.
	\newblock {\em Inf. Comput.}, 259(3):328--347, 2018.
	
	\bibitem{DBLP:conf/acl/CamburuSMLB20}
	Oana{-}Maria Camburu, Brendan Shillingford, Pasquale Minervini, Thomas
	Lukasiewicz, and Phil Blunsom.
	\newblock Make up your mind! adversarial generation of inconsistent natural
	language explanations.
	\newblock In Dan Jurafsky, Joyce Chai, Natalie Schluter, and Joel~R. Tetreault,
	editors, {\em Proceedings of the 58th Annual Meeting of the Association for
		Computational Linguistics, {ACL} 2020, Online, July 5-10, 2020}, pages
	4157--4165. Association for Computational Linguistics, 2020.
	
	\bibitem{CangelosiM22}
	Angelo Cangelosi and Minoru Asada.
	\newblock {\em {Cognitive Robotics}}.
	\newblock The MIT Press, 05 2022.
	
	\bibitem{DBLP:conf/aaai/CarlsonBKSHM10}
	Andrew Carlson, Justin Betteridge, Bryan Kisiel, Burr Settles, Estevam
	R.~Hruschka Jr., and Tom~M. Mitchell.
	\newblock Toward an architecture for never-ending language learning.
	\newblock In Maria Fox and David Poole, editors, {\em Proceedings of the
		Twenty-Fourth {AAAI} Conference on Artificial Intelligence, {AAAI} 2010,
		Atlanta, Georgia, USA, July 11-15, 2010}. {AAAI} Press, 2010.
	
	\bibitem{DBLP:conf/semweb/CarralDGJKU19}
	David Carral, Irina Dragoste, Larry Gonz{\'{a}}lez, Ceriel J.~H. Jacobs, Markus
	Kr{\"{o}}tzsch, and Jacopo Urbani.
	\newblock Vlog: {A} rule engine for knowledge graphs.
	\newblock In Chiara Ghidini, Olaf Hartig, Maria Maleshkova, Vojtech
	Sv{\'{a}}tek, Isabel~F. Cruz, Aidan Hogan, Jie Song, Maxime Lefran{\c{c}}ois,
	and Fabien Gandon, editors, {\em The Semantic Web - {ISWC} 2019 - 18th
		International Semantic Web Conference, Auckland, New Zealand, October 26-30,
		2019, Proceedings, Part {II}}, volume 11779 of {\em Lecture Notes in Computer
		Science}, pages 19--35. Springer, 2019.
	
	\bibitem{ROSPlan15}
	Michael Cashmore, Maria Fox, Derek Long, Daniele Magazzeni, Bram Ridder, Arnau
	Carrera, Narc{\'{\i}}s Palomeras, Nat{\`{a}}lia Hurt{\'{o}}s, and Marc
	Carreras.
	\newblock Rosplan: Planning in the robot operating system.
	\newblock In {\em Proceedings of the Twenty-Fifth International Conference on
		Automated Planning and Scheduling, {ICAPS} 2015, Jerusalem, Israel, June
		7-11, 2015}, pages 333--341, 2015.
	
	\bibitem{DBLP:conf/kr/CasiniMV20}
	Giovanni Casini, Thomas Meyer, and Ivan Varzinczak.
	\newblock Rational defeasible belief change.
	\newblock In Diego Calvanese, Esra Erdem, and Michael Thielscher, editors, {\em
		Proceedings of the 17th International Conference on Principles of Knowledge
		Representation and Reasoning, {KR} 2020, Rhodes, Greece, September 12-18,
		2020}, pages 213--222, 2020.
	
	\bibitem{CeruttiGTW2018}
	Federico Cerutti, Sarah~A Gaggl, Matthias Thimm, and Johannes Wallner.
	\newblock Foundations of implementations for formal argumentation.
	\newblock In Pietro Baroni, Dov Gabbay, Massimilino Giacomin, and Leendert
	Van~der Torre, editors, {\em {Handbook of Formal Argumentation}}, pages
	688--767. {College Publications}, 2018.
	
	\bibitem{CharwatDGWW2015}
	G{\"u}nther Charwat, Wolfgang Dvo{\v{r}}{\'a}k, Sarah~A Gaggl, Johannes~P
	Wallner, and Stefan Woltran.
	\newblock Methods for solving reasoning problems in abstract argumentation--a
	survey.
	\newblock {\em Artificial intelligence}, 220:28--63, 2015.
	
	\bibitem{KGs}
	Vinay Chaudhri, Chaitanya Baru, Naren Chittar, Xin Dong, Michael Genesereth,
	James Hendler, Aditya Kalyanpur, Douglas Lenat, Juan Sequeda, Denny
	Vrandečić, and Kuansan Wang.
	\newblock Knowledge graphs: Introduction, history and, perspectives.
	\newblock {\em AI Magazine}, 43(1):17--29, 2022.
	
	\bibitem{DBLP:journals/corr/abs-2202-09791}
	Jiaoyan Chen, Yuan He, Ernesto Jim{\'{e}}nez{-}Ruiz, Hang Dong, and Ian
	Horrocks.
	\newblock Contextual semantic embeddings for ontology subsumption prediction.
	\newblock {\em CoRR}, abs/2202.09791, 2022.
	
	\bibitem{DBLP:journals/ker/ChenCLWOY15}
	Juan Chen, Anthony~G. Cohn, Dayou Liu, Sheng{-}sheng Wang, Jihong Ouyang, and
	Qiangyuan Yu.
	\newblock A survey of qualitative spatial representations.
	\newblock {\em Knowl. Eng. Rev.}, 30(1):106--136, 2015.
	
	\bibitem{chopra2018handbook}
	Amit Chopra, Leendert van~der Torre, Harko Verhagen, and Serena Villata.
	\newblock {\em Handbook of normative multiagent systems}.
	\newblock College Publications, 2018.
	
	\bibitem{DBLP:books/sp/22/CiccioM22}
	Claudio~Di Ciccio and Marco Montali.
	\newblock Declarative process specifications: Reasoning, discovery, monitoring.
	\newblock In Wil M.~P. van~der Aalst and Josep Carmona, editors, {\em Process
		Mining Handbook}, volume 448 of {\em Lecture Notes in Business Information
		Processing}, pages 108--152. Springer, 2022.
	
	\bibitem{ClassenDelgrande21}
	Jens Cla{\ss}en and James~P. Delgrande.
	\newblock An account of intensional and extensional actions, and its
	application to belief, nondeterministic actions and fallible sensors.
	\newblock In Meghyn Bienvenu, Gerhard Lakemeyer, and Esra Erdem, editors, {\em
		Proceedings of the 18th International Conference on Principles of Knowledge
		Representation and Reasoning, {KR 2021}}, pages 194--204, 2021.
	
	\bibitem{DBLP:conf/birthday/ClassenLZ19}
	Jens Cla{\ss}en, Gerhard Lakemeyer, and Benjamin Zarrie{\ss}.
	\newblock Situation calculus meets description logics.
	\newblock In Carsten Lutz, Uli Sattler, Cesare Tinelli, Anni{-}Yasmin Turhan,
	and Frank Wolter, editors, {\em Description Logic, Theory Combination, and
		All That - Essays Dedicated to Franz Baader on the Occasion of His 60th
		Birthday}, volume 11560 of {\em Lecture Notes in Computer Science}, pages
	240--265. Springer, 2019.
	
	\bibitem{ClassenRLN12}
	Jens Cla{\ss}en, Gabriele R{\"{o}}ger, Gerhard Lakemeyer, and Bernhard Nebel.
	\newblock Platas - integrating planning and the action language golog.
	\newblock {\em K{\"{u}}nstliche Intell.}, 26(1):61--67, 2012.
	
	\bibitem{CocarascuSCT2020}
	Oana Cocarascu, Andria Stylianou, Kristijonas {\v{C}}yras, and Francesca Toni.
	\newblock Data-empowered argumentation for dialectically explainable
	predictions.
	\newblock In {\em {Proceedings of ECAI 2020}}, pages 2449--2456. IOS Press,
	2020.
	
	\bibitem{DBLP:journals/corr/abs-2301-12810}
	Roi Cohen, Mor Geva, Jonathan Berant, and Amir Globerson.
	\newblock Crawling the internal knowledge-base of language models.
	\newblock {\em CoRR}, abs/2301.12810, 2023.
	
	\bibitem{CohnR2008}
	Anthony~G. Cohn and Jochen Renz.
	\newblock Qualitative spatial representation and reasoning.
	\newblock In van Harmelen et~al. \cite{DBLP:reference/fai/3}, pages 551--596.
	
	\bibitem{ColmerauerR1993}
	Alain Colmerauer and Philippe Roussel.
	\newblock {The Birth of Prolog}.
	\newblock In John A.~N. Lee and Jean~E. Sammet, editors, {\em {History of
			Programming Languages Conference, HOPL-II}}, pages 37--52. {ACM}, 1993.
	
	\bibitem{10.1613/jair.1.13507}
	Andrew Cropper and Sebastijan Duman\v{c}i\'{c}.
	\newblock Inductive logic programming at 30: A new introduction.
	\newblock {\em J. Artif. Int. Res.}, 74, sep 2022.
	
	\bibitem{DBLP:conf/ijcai/Cui0L21}
	Zhenhe Cui, Yongmei Liu, and Kailun Luo.
	\newblock A uniform abstraction framework for generalized planning.
	\newblock In Zhi{-}Hua Zhou, editor, {\em Proceedings of the Thirtieth
		International Joint Conference on Artificial Intelligence, {IJCAI} 2021,
		Virtual Event / Montreal, Canada, 19-27 August 2021}, pages 1837--1844.
	ijcai.org, 2021.
	
	\bibitem{CyrasOKT2021}
	Kristijonas {\v{C}}yras, Tiago Oliveira, Amin Karamlou, and Francesca Toni.
	\newblock Assumption-based argumentation with preferences and goals for
	patient-centric reasoning with interacting clinical guidelines.
	\newblock {\em Argument \& Computation}, 12(2):149--189, 2021.
	
	\bibitem{CyrasRABT2021}
	Kristijonas {\v{C}}yras, Antonio Rago, Emanuele Albini, Pietro Baroni, and
	Francesca Toni.
	\newblock Argumentative xai: a survey.
	\newblock {\em arXiv preprint arXiv:2105.11266}, 2021.
	
	\bibitem{DBLP:conf/ijcai/Cyras0ABT21}
	Kristijonas Cyras, Antonio Rago, Emanuele Albini, Pietro Baroni, and Francesca
	Toni.
	\newblock Argumentative {XAI:} {A} survey.
	\newblock In Zhi{-}Hua Zhou, editor, {\em Proceedings of the Thirtieth
		International Joint Conference on Artificial Intelligence, {IJCAI} 2021,
		Virtual Event / Montreal, Canada, 19-27 August 2021}, pages 4392--4399.
	ijcai.org, 2021.
	
	\bibitem{DBLP:conf/emnlp/DalviJTXSPC21}
	Bhavana Dalvi, Peter Jansen, Oyvind Tafjord, Zhengnan Xie, Hannah Smith,
	Leighanna Pipatanangkura, and Peter Clark.
	\newblock Explaining answers with entailment trees.
	\newblock In Marie{-}Francine Moens, Xuanjing Huang, Lucia Specia, and
	Scott~Wen{-}tau Yih, editors, {\em Proceedings of the 2021 Conference on
		Empirical Methods in Natural Language Processing, {EMNLP} 2021, Virtual Event
		/ Punta Cana, Dominican Republic, 7-11 November, 2021}, pages 7358--7370.
	Association for Computational Linguistics, 2021.
	
	\bibitem{Darwiche2008}
	Adnan Darwiche.
	\newblock {Bayesian Networks}.
	\newblock In Frank~Van Harmelen, Vladimir Lifschitz, and Bruce Porter, editors,
	{\em {Handbook of Knowledge Representation}}, pages 467--509. Elsevier, 2008.
	
	\bibitem{DarwichePearl1997}
	Adnan Darwiche and Judea Pearl.
	\newblock {On the Logic of Iterated Belief Revision}.
	\newblock {\em Artificial intelligence}, 89(1-2):1--29, 1997.
	
	\bibitem{Davis2008}
	Ernest Davis.
	\newblock Physical reasoning.
	\newblock In van Harmelen et~al. \cite{DBLP:reference/fai/3}, pages 597--620.
	
	\bibitem{Davis17}
	Ernest Davis.
	\newblock Logical formalizations of commonsense reasoning: {A} survey.
	\newblock {\em J. Artif. Intell. Res.}, 59:651--723, 2017.
	
	\bibitem{10.1145/2701413}
	Ernest Davis and Gary Marcus.
	\newblock Commonsense reasoning and commonsense knowledge in artificial
	intelligence.
	\newblock {\em Commun. ACM}, 58(9):92–103, aug 2015.
	
	\bibitem{DavisBS1977}
	Randall Davis, Bruce Buchanan, and Edward Shortliffe.
	\newblock {Production Rules as a Representation for a Knowledge-Based
		Consultation Program}.
	\newblock {\em Artificial intelligence}, 8(1):15--45, 1977.
	
	\bibitem{DeGiacomoEtAl22}
	Giuseppe De~Giacomo, Riccardo De~Masellis, Fabrizio~Maria Maggi, and Marco
	Montali.
	\newblock Monitoring constraints and metaconstraints with temporal logics on
	finite traces.
	\newblock {\em ACM Trans. Softw. Eng. Methodol.}, 31(4), jul 2022.
	
	\bibitem{DeGiacomoRubin18}
	Giuseppe De~Giacomo and Sasha Rubin.
	\newblock Automata-theoretic foundations of fond planning for ltlf and ldlf
	goals.
	\newblock In {\em Proceedings of the 27th International Joint Conference on
		Artificial Intelligence}, IJCAI'18, page 4729–4735. AAAI Press, 2018.
	
	\bibitem{deKleerEA1992}
	Johan de~Kleer, Alan Mackworth, and Raymond Reiter.
	\newblock Characterizing diagnoses and systems.
	\newblock {\em Journal of Artificial Intelligence Research}, 52(2-3):197--222,
	1992.
	
	\bibitem{Delgrande11}
	James~P. Delgrande.
	\newblock What's in a default? {T}houghts on the nature and role of defaults in
	nonmonotonic reasoning.
	\newblock In Gerhard Brewka, Victor~W. Marek, and Miroslaw Truszczynski,
	editors, {\em Nonmonotonic Reasoning: Essays Celebrating its 30th
		Anniversary}. College Publications, 2011.
	
	\bibitem{Delgrandepeppas15}
	James~P. Delgrande and P.~Peppas.
	\newblock Belief revision in {H}orn theories.
	\newblock {\em Artificial Intelligence Journal}, 218:1 -- 22, 2015.
	
	\bibitem{DelgrandePeppasWoltran18}
	James~P. Delgrande, Pavlos Peppas, and Stefan Woltran.
	\newblock General belief revision.
	\newblock {\em Journal of the Association of Computing Machinery}, 65(5):1--34,
	2018.
	
	\bibitem{DelgrandeWassermann13}
	James~P. Delgrande and R.~Wassermann.
	\newblock Horn clause contraction functions.
	\newblock {\em Journal of Artificial Intelligence Research}, 48:475--511,
	November 2013.
	
	\bibitem{DBLP:conf/emnlp/DemeesterRR16}
	Thomas Demeester, Tim Rockt{\"{a}}schel, and Sebastian Riedel.
	\newblock Lifted rule injection for relation embeddings.
	\newblock In Jian Su, Xavier Carreras, and Kevin Duh, editors, {\em Proceedings
		of the 2016 Conference on Empirical Methods in Natural Language Processing,
		{EMNLP} 2016, Austin, Texas, USA, November 1-4, 2016}, pages 1389--1399. The
	Association for Computational Linguistics, 2016.
	
	\bibitem{DeneckerMarekTruszczynski03}
	Marc Denecker, Victor~W. Marek, and Miroslaw Truszczy\'nski.
	\newblock Uniform semantic treatment of default and autoepistemic logics.
	\newblock {\em Artificial Intelligence Journal}, 143(1):79--122, 2003.
	
	\bibitem{DBLP:journals/siglog/DeutschHLV18}
	Alin Deutsch, Richard Hull, Yuliang Li, and Victor Vianu.
	\newblock Automatic verification of database-centric systems.
	\newblock {\em {ACM} {SIGLOG} News}, 5(2):37--56, 2018.
	
	\bibitem{DodaroGRRS2019}
	Carmine Dodaro, Philip Gasteiger, Kristian Reale, Francesco Ricca, and
	Konstantin Schekotihin.
	\newblock Debugging non-ground {ASP} programs: Technique and graphical tools.
	\newblock {\em Theory Pract. Log. Program.}, 19(2):290--316, 2019.
	
	\bibitem{DoxiadisPapadimitriou09}
	A.~Doxiadis and C.H. Papadimitriou.
	\newblock {\em Logicomix}.
	\newblock Bloomsbury, New York, US, 2009.
	
	\bibitem{DudaGH1981}
	Richard Duda, John Gaschnig, and Peter Hart.
	\newblock {Model Design in the PROSPECTOR Consultant System for Mineral
		Exploration}.
	\newblock In {\em Readings in Artificial Intelligence}, pages 334--348.
	Elsevier, 1981.
	
	\bibitem{Dung1995}
	Phan~Minh Dung.
	\newblock On the acceptability of arguments and its fundamental role in
	nonmonotonic reasoning, logic programming and n-person games.
	\newblock {\em Artif. Intell.}, 77(2):321--358, 1995.
	
	\bibitem{DvorakGRW2021}
	Wolfgang Dvo{\v{r}}{\'a}k, Alexander Gre{\ss}ler, Anna Rapberger, and Stefan
	Woltran.
	\newblock The complexity landscape of claim-augmented argumentation frameworks.
	\newblock In {\em Proceedings of the AAAI Conference on Artificial
		Intelligence}, volume~35, pages 6296--6303, 2021.
	
	\bibitem{EiterG1993}
	Thomas Eiter and Georg Gottlob.
	\newblock Propositional circumscription and extended closed-world reasoning are
	$\pi_2^p$-complete.
	\newblock {\em Theor. Comput. Sci.}, 114(2):231--245, 1993.
	
	\bibitem{ErdemFMP2020}
	Esra Erdem, Muge Fidan, David~F. Manlove, and Patrick Prosser.
	\newblock A general framework for stable roommates problems using answer set
	programming.
	\newblock {\em Theory Pract. Log. Program.}, 20(6):911--925, 2020.
	
	\bibitem{ErdemGL2016}
	Esra Erdem, Michael Gelfond, and Nicola Leone.
	\newblock Applications of answer set programming.
	\newblock {\em {AI} Mag.}, 37(3):53--68, 2016.
	
	\bibitem{TFD12}
	Patrick Eyerich, Robert Mattm{\"{u}}ller, and Gabriele R{\"{o}}ger.
	\newblock Using the context-enhanced additive heuristic for temporal and
	numeric planning.
	\newblock In Erwin Prassler, Johann~Marius Z{\"{o}}llner, Rainer Bischoff,
	Wolfram Burgard, Robert Haschke, Martin H{\"{a}}gele, Gisbert Lawitzky,
	Bernhard Nebel, Paul{-}Gerhard Pl{\"{o}}ger, and Ulrich Reiser, editors, {\em
		Towards Service Robots for Everyday Environments - Recent Advances in
		Designing Service Robots for Complex Tasks in Everyday Environments},
	volume~76 of {\em Springer Tracts in Advanced Robotics}, pages 49--64.
	Springer, 2012.
	
	\bibitem{DBLP:journals/tcs/FaginKMP05}
	Ronald Fagin, Phokion~G. Kolaitis, Ren{\'{e}}e~J. Miller, and Lucian Popa.
	\newblock Data exchange: semantics and query answering.
	\newblock {\em Theor. Comput. Sci.}, 336(1):89--124, 2005.
	
	\bibitem{FalknerFSTT2018}
	Andreas~A. Falkner, Gerhard Friedrich, Konstantin Schekotihin, Richard Taupe,
	and Erich~Christian Teppan.
	\newblock Industrial applications of answer set programming.
	\newblock {\em K{\"{u}}nstliche Intell.}, 32(2-3):165--176, 2018.
	
	\bibitem{FandinnoH2021}
	Jorge Fandinno and Markus Hecher.
	\newblock Treewidth-aware complexity in {ASP:} not all positive cycles are
	equally hard.
	\newblock In {\em Thirty-Fifth {AAAI} Conference on Artificial Intelligence},
	pages 6312--6320. {AAAI} Press, 2021.
	
	\bibitem{FebbraroRR2011}
	Onofrio Febbraro, Kristian Reale, and Francesco Ricca.
	\newblock {ASPIDE:} integrated development environment for answer set
	programming.
	\newblock In James~P. Delgrande and Wolfgang Faber, editors, {\em Logic
		Programming and Nonmonotonic Reasoning - 11th International Conference,
		{LPNMR} 2011, Vancouver, Canada, May 16-19, 2011. Proceedings}, volume 6645
	of {\em Lecture Notes in Computer Science}, pages 317--330. Springer, 2011.
	
	\bibitem{DBLP:conf/bpm/FelliGMRW21}
	Paolo Felli, Alessandro Gianola, Marco Montali, Andrey Rivkin, and Sarah
	Winkler.
	\newblock Cocomot: Conformance checking of multi-perspective processes via
	{SMT}.
	\newblock In Artem Polyvyanyy, Moe~Thandar Wynn, Amy~Van Looy, and Manfred
	Reichert, editors, {\em Business Process Management - 19th International
		Conference, {BPM} 2021, Rome, Italy, September 06-10, 2021, Proceedings},
	volume 12875 of {\em Lecture Notes in Computer Science}, pages 217--234.
	Springer, 2021.
	
	\bibitem{FichteHN2022}
	Johannes~Klaus Fichte, Markus Hecher, and Mohamed~A. Nadeem.
	\newblock Plausibility reasoning via projected answer set counting - {A} hybrid
	approach.
	\newblock In Luc~De Raedt, editor, {\em Proceedings of the Thirty-First
		International Joint Conference on Artificial Intelligence, {IJCAI} 2022,
		Vienna, Austria, 23-29 July 2022}, pages 2620--2626. ijcai.org, 2022.
	
	\bibitem{FijalkowEtAl22}
	Nathanaël Fijalkow, Bastien Maubert, Aniello Murano, Sasha Rubin, and Moshe
	Vardi.
	\newblock {Public and Private Affairs in Strategic Reasoning}.
	\newblock In {\em {Proceedings of the 19th International Conference on
			Principles of Knowledge Representation and Reasoning}}, pages 132--140, 8
	2022.
	
	\bibitem{FikesN1971}
	Richard~E. Fikes and Nils~J. Nilsson.
	\newblock {STRIPS: A New Approach to the Application of Theorem Proving to
		Problem Solving}.
	\newblock {\em Artificial intelligence}, 2(3-4):189--208, 1971.
	
	\bibitem{Fisher2008}
	Michael Fisher.
	\newblock Temporal representation and reasoning.
	\newblock In van Harmelen et~al. \cite{DBLP:reference/fai/3}, pages 513--550.
	
	\bibitem{Forbus2008}
	Kenneth~D. Forbus.
	\newblock Qualitative modeling.
	\newblock In van Harmelen et~al. \cite{DBLP:reference/fai/3}, pages 361--393.
	
	\bibitem{DBLP:conf/krdb/FranconiN00}
	Enrico Franconi and Gary Ng.
	\newblock The i.com tool for intelligent conceptual modeling.
	\newblock In {\em Proceedings of the 7th International Workshop on Knowledge
		Representation meets Databases {(KRDB} 2000), Berlin, Germany, August 21,
		2000}, pages 45--53, 2000.
	
	\bibitem{GabbayEtAl13}
	Dov Gabbay, John Horty, Xavier Parent, Ron {van der Meyden}, and Leendert {van
		der Torre}, editors.
	\newblock {\em Handbook of Deontic Logic and Normative Systems}.
	\newblock College Publications, 2013.
	
	\bibitem{gabbay2003many}
	Dov Gabbay, Agi Kurucz, Frank Wolter, and Michael Zakharyachev.
	\newblock {\em Many-dimensional modal logics: theory and applications}.
	\newblock Studies in Logic and the Foundations of Mathematics. Elsevier, 2003.
	
	\bibitem{GardenforsMakinson1994}
	P.~G{\"a}rdenfors and D.~Makinson.
	\newblock Nonmonotonic inference based on expectations.
	\newblock {\em Artificial Intelligence}, 65(2):197--245, 1994.
	
	\bibitem{TaskMotionPlanning20}
	Caelan~Reed Garrett, Rohan Chitnis, Rachel Holladay, Beomjoon Kim, Tom Silver,
	Leslie~Pack Kaelbling, and Tom{\'{a}}s Lozano{-}P{\'{e}}rez.
	\newblock Integrated task and motion planning.
	\newblock {\em CoRR}, abs/2010.01083, 2020.
	
	\bibitem{DBLP:conf/aaai/Geffner22}
	Hector Geffner.
	\newblock Target languages (vs. inductive biases) for learning to act and plan.
	\newblock In {\em Thirty-Sixth {AAAI} Conference on Artificial Intelligence,
		{AAAI} 2022, Thirty-Fourth Conference on Innovative Applications of
		Artificial Intelligence, {IAAI} 2022, The Twelveth Symposium on Educational
		Advances in Artificial Intelligence, {EAAI} 2022 Virtual Event, February 22 -
		March 1, 2022}, pages 12326--12333. {AAAI} Press, 2022.
	
	\bibitem{DBLP:journals/natmi/GeirhosJMZBBW20}
	Robert Geirhos, J{\"{o}}rn{-}Henrik Jacobsen, Claudio Michaelis, Richard~S.
	Zemel, Wieland Brendel, Matthias Bethge, and Felix~A. Wichmann.
	\newblock Shortcut learning in deep neural networks.
	\newblock {\em Nat. Mach. Intell.}, 2(11):665--673, 2020.
	
	\bibitem{GelfondL1988}
	Michael Gelfond and Vladimir Lifschitz.
	\newblock The stable model semantics for logic programming.
	\newblock In Robert~A. Kowalski and Kenneth~A. Bowen, editors, {\em Logic
		Programming, Proceedings of the Fifth International Conference and
		Symposium}, pages 1070--1080. {MIT} Press, 1988.
	
	\bibitem{GelfondL1993}
	Michael Gelfond and Vladimir Lifschitz.
	\newblock {Representing Action and Change by Logic Programs}.
	\newblock {\em The Journal of Logic Programming}, 17(2-4):301--321, 1993.
	
	\bibitem{GetoorTaskar:book07}
	Lise Getoor and Ben Taskar, editors.
	\newblock {\em Introduction to Statistical Relational Learning}.
	\newblock Adaptive Computation and Machine Learning. {MIT} Press, 2007.
	
	\bibitem{GhallabLaruelle94}
	Malik Ghallab and Herv{\'{e}} Laruelle.
	\newblock Representation and control in ixtet, a temporal planner.
	\newblock In {\em Proceedings of the Second International Conference on
		Artificial Intelligence Planning Systems, University of Chicago, Chicago,
		Illinois, USA, June 13-15, 1994}, pages 61--67, 1994.
	
	\bibitem{DBLP:journals/ai/GiacomoFLPS22}
	Giuseppe~De Giacomo, Paolo Felli, Brian Logan, Fabio Patrizi, and Sebastian
	Sardi{\~{n}}a.
	\newblock Situation calculus for controller synthesis in manufacturing systems
	with first-order state representation.
	\newblock {\em Artif. Intell.}, 302:103598, 2022.
	
	\bibitem{DBLP:conf/aaai/GiacomoIFP20}
	Giuseppe~De Giacomo, Luca Iocchi, Marco Favorito, and Fabio Patrizi.
	\newblock Restraining bolts for reinforcement learning agents.
	\newblock In {\em The Thirty-Fourth {AAAI} Conference on Artificial
		Intelligence, {AAAI} 2020, The Thirty-Second Innovative Applications of
		Artificial Intelligence Conference, {IAAI} 2020, The Tenth {AAAI} Symposium
		on Educational Advances in Artificial Intelligence, {EAAI} 2020, New York,
		NY, USA, February 7-12, 2020}, pages 13659--13662. {AAAI} Press, 2020.
	
	\bibitem{DBLP:books/sp/18/GiacomoLLPR18}
	Giuseppe~De Giacomo, Domenico Lembo, Maurizio Lenzerini, Antonella Poggi, and
	Riccardo Rosati.
	\newblock Using ontologies for semantic data integration.
	\newblock In {\em A Comprehensive Guide Through the Italian Database Research
		Over the Last 25 Years}, pages 187--202. 2018.
	
	\bibitem{DBLP:conf/atal/GiacomoL20}
	Giuseppe~De Giacomo and Yves Lesp{\'{e}}rance.
	\newblock Goal formation through interaction in the situation calculus: {A}
	formal account grounded in behavioral science.
	\newblock In Amal El~Fallah Seghrouchni, Gita Sukthankar, Bo~An, and Neil
	Yorke{-}Smith, editors, {\em Proceedings of the 19th International Conference
		on Autonomous Agents and Multiagent Systems, {AAMAS} '20, Auckland, New
		Zealand, May 9-13, 2020}, pages 294--302. International Foundation for
	Autonomous Agents and Multiagent Systems, 2020.
	
	\bibitem{DBLP:journals/ai/GiacomoLP16}
	Giuseppe~De Giacomo, Yves Lesp{\'{e}}rance, and Fabio Patrizi.
	\newblock Bounded situation calculus action theories.
	\newblock {\em Artif. Intell.}, 237:172--203, 2016.
	
	\bibitem{DBLP:conf/aaai/GiacomoLT20}
	Giuseppe~De Giacomo, Yves Lesp{\'{e}}rance, and Eugenia Ternovska.
	\newblock Elgolog: {A} high-level programming language with memory of the
	execution history.
	\newblock In {\em The Thirty-Fourth {AAAI} Conference on Artificial
		Intelligence, {AAAI} 2020, The Thirty-Second Innovative Applications of
		Artificial Intelligence Conference, {IAAI} 2020, The Tenth {AAAI} Symposium
		on Educational Advances in Artificial Intelligence, {EAAI} 2020, New York,
		NY, USA, February 7-12, 2020}, pages 2806--2813. {AAAI} Press, 2020.
	
	\bibitem{DBLP:conf/aaai/XuLZ17a}
	Giuseppe~De Giacomo, Fabrizio~Maria Maggi, Andrea Marrella, and Fabio Patrizi.
	\newblock On the disruptive effectiveness of automated planning for
	ltl\emph{f}-based trace alignment.
	\newblock In Satinder Singh and Shaul Markovitch, editors, {\em Proceedings of
		the Thirty-First {AAAI} Conference on Artificial Intelligence, February 4-9,
		2017, San Francisco, California, {USA}}, pages 3555--3561. {AAAI} Press,
	2017.
	
	\bibitem{DBLP:conf/kr/GiacomoRS98}
	Giuseppe~De Giacomo, Raymond Reiter, and Mikhail Soutchanski.
	\newblock Execution monitoring of high-level robot programs.
	\newblock In Anthony~G. Cohn, Lenhart~K. Schubert, and Stuart~C. Shapiro,
	editors, {\em Proceedings of the Sixth International Conference on Principles
		of Knowledge Representation and Reasoning (KR'98), Trento, Italy, June 2-5,
		1998}, pages 453--465. Morgan Kaufmann, 1998.
	
	\bibitem{DBLP:conf/ijcai/GiacomoV13}
	Giuseppe~De Giacomo and Moshe~Y. Vardi.
	\newblock Linear temporal logic and linear dynamic logic on finite traces.
	\newblock In Francesca Rossi, editor, {\em {IJCAI} 2013, Proceedings of the
		23rd International Joint Conference on Artificial Intelligence, Beijing,
		China, August 3-9, 2013}, pages 854--860. {IJCAI/AAAI}, 2013.
	
	\bibitem{DBLP:conf/ijcai/GiacomoV15}
	Giuseppe~De Giacomo and Moshe~Y. Vardi.
	\newblock Synthesis for {LTL} and {LDL} on finite traces.
	\newblock In Qiang Yang and Michael~J. Wooldridge, editors, {\em Proceedings of
		the Twenty-Fourth International Joint Conference on Artificial Intelligence,
		{IJCAI} 2015, Buenos Aires, Argentina, July 25-31, 2015}, pages 1558--1564.
	{AAAI} Press, 2015.
	
	\bibitem{GiordanoEtAl2015}
	L.~Giordano, V.~Gliozzi, N.~Olivetti, and G.L. Pozzato.
	\newblock Semantic characterization of rational closure: From propositional
	logic to description logics.
	\newblock {\em Artificial Intelligence}, 226:1--33, 2015.
	
	\bibitem{GHMS14a}
	Birte Glimm, Ian Horrocks, Boris Motik, Giorgos Stoilos, and Zhe Wang.
	\newblock Hermit: An {OWL} 2 reasoner.
	\newblock {\em Journal of Automated Reasoning}, 2014.
	\newblock submitted.
	
	\bibitem{DBLP:conf/rweb/GlimmK19}
	Birte Glimm and Yevgeny Kazakov.
	\newblock Classical algorithms for reasoning and explanation in description
	logics.
	\newblock In Markus Kr{\"{o}}tzsch and Daria Stepanova, editors, {\em Reasoning
		Web. Explainable Artificial Intelligence - 15th International Summer School
		2019}, volume 11810 of {\em Lecture Notes in Computer Science}, pages 1--64.
	Springer, 2019.
	
	\bibitem{GoffredoHVCV2022}
	Pierpaolo Goffredo, Shohreh Haddadan, Vorakit Vorakitphan, Elena Cabrio, and
	Serena Villata.
	\newblock Fallacious argument classification in political debates.
	\newblock In {\em {Proceedings of the 31st International Joint Conference on
			Artificial Intelligence, IJCAI 2022}}, pages 4143--4149, 2022.
	
	\bibitem{GorogiannisH2011}
	Nikos Gorogiannis and Anthony Hunter.
	\newblock Instantiating abstract argumentation with classical logic arguments:
	Postulates and properties.
	\newblock {\em Artificial Intelligence}, 175(9-10):1479--1497, 2011.
	
	\bibitem{Gottlob1992}
	Georg Gottlob.
	\newblock Complexity results for nonmonotonic logics.
	\newblock {\em J. Log. Comput.}, 2(3):397--425, 1992.
	
	\bibitem{HahnSSS2022}
	Susana Hahn, Orkunt Sabuncu, Torsten Schaub, and Tobias Stolzmann.
	\newblock Clingraph: {ASP}-based visualization.
	\newblock In Georg Gottlob, Daniela Inclezan, and Marco Maratea, editors, {\em
		Logic Programming and Nonmonotonic Reasoning - 16th International Conference,
		{LPNMR} 2022, Genova, Italy, September 5-9, 2022, Proceedings}, volume 13416
	of {\em Lecture Notes in Computer Science}, pages 401--414. Springer, 2022.
	
	\bibitem{DBLP:conf/sat/HaimW09}
	Shai Haim and Toby Walsh.
	\newblock Restart strategy selection using machine learning techniques.
	\newblock In Oliver Kullmann, editor, {\em Theory and Applications of
		Satisfiability Testing - {SAT} 2009, 12th International Conference, {SAT}
		2009, Swansea, UK, June 30 - July 3, 2009. Proceedings}, volume 5584 of {\em
		Lecture Notes in Computer Science}, pages 312--325. Springer, 2009.
	
	\bibitem{DBLP:journals/ai/Halpern90}
	Joseph~Y. Halpern.
	\newblock An analysis of first-order logics of probability.
	\newblock {\em Artif. Intell.}, 46(3):311--350, 1990.
	
	\bibitem{DBLP:journals/corr/abs-2206-14268}
	Shibo Hao, Bowen Tan, Kaiwen Tang, Hengzhe Zhang, Eric~P. Xing, and Zhiting Hu.
	\newblock Bertnet: Harvesting knowledge graphs from pretrained language models.
	\newblock {\em CoRR}, abs/2206.14268, 2022.
	
	\bibitem{DBLP:journals/jair/HaririCMGMF13}
	Babak~Bagheri Hariri, Diego Calvanese, Marco Montali, Giuseppe~De Giacomo,
	Riccardo~De Masellis, and Paolo Felli.
	\newblock Description logic knowledge and action bases.
	\newblock {\em J. Artif. Intell. Res.}, 46:651--686, 2013.
	
	\bibitem{HarrisonLY2014}
	Amelia~J. Harrison, Vladimir Lifschitz, and Fangkai Yang.
	\newblock The semantics of gringo and infinitary propositional formulas.
	\newblock In Chitta Baral, Giuseppe~De Giacomo, and Thomas Eiter, editors, {\em
		Principles of Knowledge Representation and Reasoning: Proceedings of the
		Fourteenth International Conference, {KR} 2014, Vienna, Austria, July 20-24,
		2014}. {AAAI} Press, 2014.
	
	\bibitem{rdf11-semantics}
	Patrick~J. Hayes and Peter~F. Patel-Schneider, editors.
	\newblock {\em {RDF 1.1 Semantics}}.
	\newblock W3C Recommendation, 25 February 2014.
	\newblock Available at \url{https://www.w3.org/TR/rdf11-mt/}.
	
	\bibitem{Hayes77}
	P.J. Hayes.
	\newblock In defense of logic.
	\newblock In {\em Proceedings of the International Joint Conference on
		Artificial Intelligence}, pages 559--565, Cambridge, MA, 1977.
	
	\bibitem{Hayes79}
	P.J. Hayes.
	\newblock The logic of frames.
	\newblock In D.~Metzing, editor, {\em Frame Conceptions and Text
		Understanding}, pages 46--61. Walter de Gruyter and Co., 1979.
	
	\bibitem{Hayes85b}
	P.J. Hayes.
	\newblock Naive physics {I}: Ontology for liquids.
	\newblock In J.R. Hobbs and R.C. Moore, editors, {\em Formal Theories of the
		Commonsense World}, pages 71--108. Ablex, 1985.
	
	\bibitem{Hayes85}
	P.J. Hayes.
	\newblock The second naive physics manifesto.
	\newblock In J.R. Hobbs and R.C. Moore, editors, {\em Formal Theories of the
		Commonsense World}, pages 1--36. Ablex, 1985.
	
	\bibitem{Hecher2022}
	Markus Hecher.
	\newblock Treewidth-aware reductions of normal {ASP} to {SAT} - is normal {ASP}
	harder than {SAT} after all?
	\newblock {\em Artif. Intell.}, 304:103651, 2022.
	
	\bibitem{HeintzLM23}
	Fredrik Heintz, Gerhard Lakemeyer, and Sheila McIlraith.
	\newblock {Cognitive Robotics (Dagstuhl Seminar 22391)}.
	\newblock {\em Dagstuhl Reports}, 12(9):200--219, 2023.
	
	\bibitem{Hewitt71}
	C.~Hewitt.
	\newblock Planner: A language for proving theorems in robots.
	\newblock In {\em International Joint Conference on Artificial Intelligence},
	London, U.K., 1971.
	
	\bibitem{DBLP:series/faia/342}
	Pascal Hitzler and Md.~Kamruzzaman Sarker, editors.
	\newblock {\em Neuro-Symbolic Artificial Intelligence: The State of the Art},
	volume 342 of {\em Frontiers in Artificial Intelligence and Applications}.
	\newblock {IOS} Press, 2021.
	
	\bibitem{HoffmannNebel01}
	J{\"{o}}rg Hoffmann and Bernhard Nebel.
	\newblock The {FF} planning system: Fast plan generation through heuristic
	search.
	\newblock {\em J. Artif. Intell. Res.}, 14:253--302, 2001.
	
	\bibitem{DBLP:conf/aaai/HofmannNCL16}
	Till Hofmann, Tim Niemueller, Jens Cla{\ss}en, and Gerhard Lakemeyer.
	\newblock Continual planning in golog.
	\newblock In Dale Schuurmans and Michael~P. Wellman, editors, {\em Proceedings
		of the Thirtieth {AAAI} Conference on Artificial Intelligence, February
		12-17, 2016, Phoenix, Arizona, {USA}}, pages 3346--3353. {AAAI} Press, 2016.
	
	\bibitem{DBLP:conf/icaart/HofmannVGHNL21}
	Till Hofmann, Tarik Viehmann, Mostafa Gomaa, Daniel Habering, Tim Niemueller,
	and Gerhard Lakemeyer.
	\newblock Multi-agent goal reasoning with the {CLIPS} executive in the robocup
	logistics league.
	\newblock In Ana~Paula Rocha, Luc Steels, and H.~Jaap van~den Herik, editors,
	{\em Proceedings of the 13th International Conference on Agents and
		Artificial Intelligence, {ICAART} 2021, Volume 1, Online Streaming, February
		4-6, 2021}, pages 80--91. {SCITEPRESS}, 2021.
	
	\bibitem{10.1145/3447772}
	Aidan Hogan, Eva Blomqvist, Michael Cochez, Claudia D’amato, Gerard~De Melo,
	Claudio Gutierrez, Sabrina Kirrane, Jos\'{e} Emilio~Labra Gayo, Roberto
	Navigli, Sebastian Neumaier, Axel-Cyrille~Ngonga Ngomo, Axel Polleres,
	Sabbir~M. Rashid, Anisa Rula, Lukas Schmelzeisen, Juan Sequeda, Steffen
	Staab, and Antoine Zimmermann.
	\newblock Knowledge graphs.
	\newblock {\em ACM Comput. Surv.}, 54(4), jul 2021.
	
	\bibitem{Horridge:PhD}
	Matthew Horridge.
	\newblock {\em Justification based explanation in ontologies}.
	\newblock PhD thesis, University of Manchester, {UK}, 2011.
	
	\bibitem{Horrocks2002}
	Ian Horrocks.
	\newblock {DAML+OIL}: A description logic for the semantic web.
	\newblock {\em IEEE Data Eng. Bull.}, 25(1):4--9, 2002.
	
	\bibitem{horty2019epistemic}
	John Horty.
	\newblock Epistemic oughts in stit semantics.
	\newblock {\em Ergo, an Open Access Journal of Philosophy}, 6, 2019.
	
	\bibitem{HughesCresswell68}
	G.E. Hughes and M.J. Cresswell.
	\newblock {\em An Introduction to Modal Logic}.
	\newblock Methuen and Co. Ltd., 1968.
	
	\bibitem{Hunter2018}
	Anthony Hunter.
	\newblock Towards a framework for computational persuasion with applications in
	behaviour change.
	\newblock {\em Argument \& Computation}, 9(1):15--40, 2018.
	
	\bibitem{HunterEtAl19}
	Anthony Hunter, Gabriele Kern{-}Isberner, Thomas Meyer, and Renata Wassermann.
	\newblock The role of non-monotonic reasoning in future development of
	artificial intelligence ({D}agstuhl {P}erspectives workshop 19072).
	\newblock {\em Dagstuhl Reports}, 9(2):73--90, 2019.
	
	\bibitem{DBLP:journals/jair/HupkesDMB20}
	Dieuwke Hupkes, Verna Dankers, Mathijs Mul, and Elia Bruni.
	\newblock Compositionality decomposed: How do neural networks generalise?
	\newblock {\em J. Artif. Intell. Res.}, 67:757--795, 2020.
	
	\bibitem{DBLP:journals/corr/abs-2010-05953}
	Jena~D. Hwang, Chandra Bhagavatula, Ronan~Le Bras, Jeff Da, Keisuke Sakaguchi,
	Antoine Bosselut, and Yejin Choi.
	\newblock {COMET-ATOMIC} 2020: On symbolic and neural commonsense knowledge
	graphs.
	\newblock {\em CoRR}, abs/2010.05953, 2020.
	
	\bibitem{DBLP:journals/jair/IcarteKVM22}
	Rodrigo~Toro Icarte, Toryn~Q. Klassen, Richard~Anthony Valenzano, and Sheila~A.
	McIlraith.
	\newblock Reward machines: Exploiting reward function structure in
	reinforcement learning.
	\newblock {\em J. Artif. Intell. Res.}, 73:173--208, 2022.
	
	\bibitem{IrwinRT2022}
	Benjamin Irwin, Antonio Rago, and Francesca Toni.
	\newblock Forecasting argumentation frameworks.
	\newblock {\em arXiv preprint arXiv:2205.11590}, 2022.
	
	\bibitem{JiPCMY2022}
	Shaoxiong Ji, Shirui Pan, Erik Cambria, Pekka Marttinen, and Philip~S. Yu.
	\newblock {A Survey on Knowledge Graphs: Representation, Acquisition, and
		Applications}.
	\newblock {\em IEEE Transactions on Neural Networks and Learning Systems},
	33(2):494--514, 2022.
	
	\bibitem{10.5555/3495724.3495875}
	Kai Jia and Martin Rinard.
	\newblock Efficient exact verification of binarized neural networks.
	\newblock In {\em Proceedings of the 34th International Conference on Neural
		Information Processing Systems}, NIPS'20. Curran Associates Inc., 2020.
	
	\bibitem{JIMENEZRUIZ2011146}
	E.~{Jiménez Ruiz}, B.~Cuenca Grau, I.~Horrocks, and R.~Berlanga.
	\newblock Supporting concurrent ontology development: Framework, algorithms and
	tool.
	\newblock {\em Data and Knowledge Engineering}, 70(1):146--164, 2011.
	
	\bibitem{JinThielscher07}
	Y.~Jin and M.~Thielscher.
	\newblock Iterated belief revision, revised.
	\newblock {\em Artificial Intelligence}, 171(1):1--18, 2007.
	
	\bibitem{DBLP:journals/corr/abs-2204-10176}
	Zijian Jin, Xingyu Zhang, Mo~Yu, and Lifu Huang.
	\newblock Probing script knowledge from pre-trained models.
	\newblock {\em CoRR}, abs/2204.10176, 2022.
	
	\bibitem{DBLP:journals/ai/JungLPW22}
	Jean~Christoph Jung, Carsten Lutz, Hadrien Pulcini, and Frank Wolter.
	\newblock Logical separability of labeled data examples under ontologies.
	\newblock {\em Artif. Intell.}, 313:103785, 2022.
	
	\bibitem{KabirESHFM2022}
	Mohimenul Kabir, Flavio~O. Everardo, Ankit~K. Shukla, Markus Hecher,
	Johannes~Klaus Fichte, and Kuldeep~S. Meel.
	\newblock Approx{ASP} - a scalable approximate answer set counter.
	\newblock In {\em Thirty-Sixth {AAAI} Conference on Artificial Intelligence},
	pages 5755--5764. {AAAI} Press, 2022.
	
	\bibitem{KaminskiRSW2020}
	Roland Kaminski, Javier Romero, Torsten Schaub, and Philipp Wanko.
	\newblock How to build your own {ASP}-based system?!
	\newblock {\em CoRR}, abs/2008.06692, 2020.
	
	\bibitem{KaminskiS2021}
	Roland Kaminski and Torsten Schaub.
	\newblock On the foundations of grounding in answer set programming.
	\newblock {\em CoRR}, abs/2108.04769, 2021.
	
	\bibitem{KatsunoMendelzon92}
	H.~Katsuno and A.~Mendelzon.
	\newblock On the difference between updating a knowledge base and revising it.
	\newblock In P.~G{\"{a}}rdenfors, editor, {\em Belief Revision}, pages
	183--203, Cambridge, 1992. Cambridge University Press.
	
	\bibitem{ELK-JAR}
	Yevgeny Kazakov, Markus Kr{\"o}tzsch, and Franti\v{s}ek Siman\v{c}\'{i}k.
	\newblock The incredible {ELK}: From polynomial procedures to efficient
	reasoning with $\mathcal{EL}$ ontologies.
	\newblock {\em Journal of Automated Reasoning}, 53(1):1--61, 2014.
	
	\bibitem{DBLP:conf/icml/KohNTMPKL20}
	Pang~Wei Koh, Thao Nguyen, Yew~Siang Tang, Stephen Mussmann, Emma Pierson, Been
	Kim, and Percy Liang.
	\newblock Concept bottleneck models.
	\newblock In {\em Proceedings of the 37th International Conference on Machine
		Learning, {ICML} 2020, 13-18 July 2020, Virtual Event}, volume 119 of {\em
		Proceedings of Machine Learning Research}, pages 5338--5348. {PMLR}, 2020.
	
	\bibitem{KonevWW09}
	Boris Konev, Dirk Walther, and Frank Wolter.
	\newblock Forgetting and uniform interpolation in large-scale description logic
	terminologies.
	\newblock In Craig Boutilier, editor, {\em IJCAI}, pages 830--835, 2009.
	
	\bibitem{KoniecznyPinoPerez02}
	S.~Konieczny and R.~{Pino P\'erez}.
	\newblock Merging information under constraints: A logical framework.
	\newblock {\em Journal of Logic and Computation}, 12(5):773--808, 2002.
	
	\bibitem{Konieczny2000}
	S{\'{e}}bastien Konieczny.
	\newblock {On the Difference between Merging Knowledge Bases and Combining
		them}.
	\newblock In Anthony~G. Cohn, Fausto Giunchiglia, and Bart Selman, editors,
	{\em {Proceedings od the 7th International Conference on Principles of
			Knowledge Representation and Reasoning, KR 2000}}, pages 135--144. Morgan
	Kaufmann, 2000.
	
	\bibitem{Kowalski1974}
	Robert Kowalski.
	\newblock {Predicate Logic as Programming Language}.
	\newblock In {\em IFIP congress}, volume~74, pages 569--544, 1974.
	
	\bibitem{KowalskiK1971}
	Robert Kowalski and Donald Kuehner.
	\newblock {Linear Resolution with Selection Function}.
	\newblock {\em Artificial Intelligence}, 2(3-4):227--260, 1971.
	
	\bibitem{KowalskiS1986}
	Robert~A. Kowalski and Marek~J. Sergot.
	\newblock {A Logic-based Calculus of Events}.
	\newblock {\em New Gener. Comput.}, 4(1):67--95, 1986.
	
	\bibitem{KrausLehmannMagidor90}
	S.~Kraus, D.~Lehmann, and M.~Magidor.
	\newblock Nonmonotonic reasoning, preferential models and cumulative logics.
	\newblock {\em Artificial Intelligence Journal}, 44(1-2):167--207, 1990.
	
	\bibitem{Kripke80}
	S.A. Kripke.
	\newblock {\em Naming and Necessity}.
	\newblock Harvard University Press, Cambridge, Mass., 1980.
	
	\bibitem{DBLP:conf/ijcai/Krotzsch0OT18}
	Markus Kr{\"{o}}tzsch, Maximilian Marx, Ana Ozaki, and Veronika Thost.
	\newblock Attributed description logics: Reasoning on knowledge graphs.
	\newblock In J{\'{e}}r{\^{o}}me Lang, editor, {\em Proceedings of the
		Twenty-Seventh International Joint Conference on Artificial Intelligence,
		{IJCAI} 2018, July 13-19, 2018, Stockholm, Sweden}, pages 5309--5313.
	ijcai.org, 2018.
	
	\bibitem{KuipersFHN2017}
	Benjamin Kuipers, Edward Feigenbaum, Peter~E. Hart, and Nils~J. Nilsson.
	\newblock Shakey: from conception to history.
	\newblock {\em Ai Magazine}, 38(1):88--103, 2017.
	
	\bibitem{DBLP:conf/ijcai/KuzelkaDS16}
	Ondrej Kuzelka, Jesse Davis, and Steven Schockaert.
	\newblock Learning possibilistic logic theories from default rules.
	\newblock In Subbarao Kambhampati, editor, {\em Proceedings of the Twenty-Fifth
		International Joint Conference on Artificial Intelligence, {IJCAI} 2016, New
		York, NY, USA, 9-15 July 2016}, pages 1167--1173. {IJCAI/AAAI} Press, 2016.
	
	\bibitem{KvarnstromDoherty00}
	Jonas Kvarnstr{\"{o}}m and Patrick Doherty.
	\newblock Talplanner: {A} temporal logic based forward chaining planner.
	\newblock {\em Ann. Math. Artif. Intell.}, 30(1-4):119--169, 2000.
	
	\bibitem{LeauteWilliams05}
	Thomas L{\'{e}}aut{\'{e}} and Brian~C. Williams.
	\newblock Coordinating agile systems through the model-based execution of
	temporal plans.
	\newblock In {\em Proceedings, The Twentieth National Conference on Artificial
		Intelligence}, pages 114--120, 2005.
	
	\bibitem{LeeY2017}
	Joohyung Lee and Zhun Yang.
	\newblock {LPMLN}, weak constraints, and p-log.
	\newblock In Satinder Singh and Shaul Markovitch, editors, {\em Proceedings of
		the Thirty-First {AAAI} Conference on Artificial Intelligence}, pages
	1170--1177. {AAAI} Press, 2017.
	
	\bibitem{DBLP:conf/dlog/LeeMPB06}
	Kevin Lee, Thomas~Andreas Meyer, Jeff~Z. Pan, and Richard Booth.
	\newblock Computing maximally satisfiable terminologies for the description
	logic \emph{ALC} with cyclic definitions.
	\newblock In Bijan Parsia, Ulrike Sattler, and David Toman, editors, {\em
		Proceedings of the 2006 International Workshop on Description Logics
		(DL2006), Windermere, Lake District, UK, May 30 - June 1, 2006}, volume 189
	of {\em {CEUR} Workshop Proceedings}. CEUR-WS.org, 2006.
	
	\bibitem{LehmannMagidor1992}
	Daniel Lehmann and Menachem Magidor.
	\newblock {What does a conditional knowledge base entail?}
	\newblock {\em Art. Intell.}, 55:1--60, 1992.
	
	\bibitem{LehtonenWJ2021}
	Tuomo Lehtonen, Johannes~P Wallner, and Matti J{\"a}rvisalo.
	\newblock Declarative algorithms and complexity results for assumption-based
	argumentation.
	\newblock {\em Journal of Artificial Intelligence Research}, 71:265--318, 2021.
	
	\bibitem{DBLP:conf/pods/Lenzerini02}
	Maurizio Lenzerini.
	\newblock Data integration: {A} theoretical perspective.
	\newblock In {\em Proceedings of the Twenty-first {ACM} {SIGACT-SIGMOD-SIGART}
		Symposium on Principles of Database Systems, June 3-5, Madison, Wisconsin,
		{USA}}, pages 233--246, 2002.
	
	\bibitem{LakemeyerLevesque08}
	Hector~J. Levesque and Gerhard Lakemeyer.
	\newblock Cognitive robotics.
	\newblock In van Harmelen et~al. \cite{DBLP:reference/fai/3}, pages 869--886.
	
	\bibitem{Golog97}
	Hector~J. Levesque, Raymond Reiter, Yves Lesp{\'{e}}rance, Fangzhen Lin, and
	Richard~B. Scherl.
	\newblock {GOLOG:} {A} logic programming language for dynamic domains.
	\newblock {\em J. Log. Program.}, 31(1-3):59--83, 1997.
	
	\bibitem{DBLP:conf/semweb/LiBS19}
	Na~Li, Zied Bouraoui, and Steven Schockaert.
	\newblock Ontology completion using graph convolutional networks.
	\newblock In Chiara Ghidini, Olaf Hartig, Maria Maleshkova, Vojtech
	Sv{\'{a}}tek, Isabel~F. Cruz, Aidan Hogan, Jie Song, Maxime Lefran{\c{c}}ois,
	and Fabien Gandon, editors, {\em The Semantic Web - {ISWC} 2019 - 18th
		International Semantic Web Conference, Auckland, New Zealand, October 26-30,
		2019, Proceedings, Part {I}}, volume 11778 of {\em Lecture Notes in Computer
		Science}, pages 435--452. Springer, 2019.
	
	\bibitem{DBLP:conf/aaai/LiLCR18}
	Oscar Li, Hao Liu, Chaofan Chen, and Cynthia Rudin.
	\newblock Deep learning for case-based reasoning through prototypes: {A} neural
	network that explains its predictions.
	\newblock In Sheila~A. McIlraith and Kilian~Q. Weinberger, editors, {\em
		Proceedings of the Thirty-Second {AAAI} Conference on Artificial
		Intelligence, (AAAI-18), the 30th innovative Applications of Artificial
		Intelligence (IAAI-18), and the 8th {AAAI} Symposium on Educational Advances
		in Artificial Intelligence (EAAI-18), New Orleans, Louisiana, USA, February
		2-7, 2018}, pages 3530--3537. {AAAI} Press, 2018.
	
	\bibitem{DBLP:conf/acl/LiS19}
	Tao Li and Vivek Srikumar.
	\newblock Augmenting neural networks with first-order logic.
	\newblock In Anna Korhonen, David~R. Traum, and Llu{\'{\i}}s M{\`{a}}rquez,
	editors, {\em Proceedings of the 57th Conference of the Association for
		Computational Linguistics, {ACL} 2019, Florence, Italy, July 28- August 2,
		2019, Volume 1: Long Papers}, pages 292--302. Association for Computational
	Linguistics, 2019.
	
	\bibitem{DBLP:conf/sat/LiangGPC16}
	Jia~Hui Liang, Vijay Ganesh, Pascal Poupart, and Krzysztof Czarnecki.
	\newblock Learning rate based branching heuristic for {SAT} solvers.
	\newblock In Nadia Creignou and Daniel~Le Berre, editors, {\em Theory and
		Applications of Satisfiability Testing - {SAT} 2016 - 19th International
		Conference, Bordeaux, France, July 5-8, 2016, Proceedings}, volume 9710 of
	{\em Lecture Notes in Computer Science}, pages 123--140. Springer, 2016.
	
	\bibitem{Lierler2021}
	Yuliya Lierler.
	\newblock Constraint answer set programming: Integrational and translational
	(or smt-based) approaches.
	\newblock {\em CoRR}, abs/2107.08252, 2021.
	
	\bibitem{LierlerMR2016}
	Yuliya Lierler, Marco Maratea, and Francesco Ricca.
	\newblock Systems, engineering environments, and competitions.
	\newblock {\em {AI} Mag.}, 37(3):45--52, 2016.
	
	\bibitem{Lin2023RepairingClassicalModels}
	Songtuan Lin, Alban Grastien, and Pascal Bercher.
	\newblock Towards automated modeling assistance: An efficient approach for
	repairing flawed planning domains.
	\newblock In {\em Proceedings of the 37th AAAI Conference on Artificial
		Intelligence (AAAI 2023)}. AAAI Press, 2023.
	
	\bibitem{LindnerMattmuellerNebel2020}
	Felix Lindner, Robert Mattmüller, and Bernhard Nebel.
	\newblock Evaluation of the moral permissibility of action plans.
	\newblock {\em Artif. Intell.}, 287, 2020.
	
	\bibitem{DBLP:conf/ijcai/LiuL21}
	Daxin Liu and Gerhard Lakemeyer.
	\newblock Reasoning about beliefs and meta-beliefs by regression in an
	expressive probabilistic action logic.
	\newblock In Zhi{-}Hua Zhou, editor, {\em Proceedings of the Thirtieth
		International Joint Conference on Artificial Intelligence, {IJCAI} 2021,
		Virtual Event / Montreal, Canada, 19-27 August 2021}, pages 1951--1958.
	ijcai.org, 2021.
	
	\bibitem{LuckcuckFDDF19}
	Matt Luckcuck, Marie Farrell, Louise~A. Dennis, Clare Dixon, and Michael
	Fisher.
	\newblock Formal specification and verification of autonomous robotic systems:
	{A} survey.
	\newblock {\em {ACM} Comput. Surv.}, 52(5):100:1--100:41, 2019.
	
	\bibitem{LUKASIEWICZ2008852}
	Thomas Lukasiewicz.
	\newblock Expressive probabilistic description logics.
	\newblock {\em Artificial Intelligence}, 172(6):852--883, 2008.
	
	\bibitem{DBLP:conf/time/LutzWZ08}
	Carsten Lutz, Frank Wolter, and Michael Zakharyaschev.
	\newblock Temporal description logics: {A} survey.
	\newblock In St{\'{e}}phane Demri and Christian~S. Jensen, editors, {\em 15th
		International Symposium on Temporal Representation and Reasoning, {TIME}
		2008, Universit{\'{e}} du Qu{\'{e}}bec {\`{a}} Montr{\'{e}}al, Canada, 16-18
		June 2008}, pages 3--14. {IEEE} Computer Society, 2008.
	
	\bibitem{DBLP:journals/ai/ManhaeveDKDR21}
	Robin Manhaeve, Sebastijan Dumancic, Angelika Kimmig, Thomas Demeester, and
	Luc~De Raedt.
	\newblock Neural probabilistic logic programming in {DeepProbLog}.
	\newblock {\em Artif. Intell.}, 298:103504, 2021.
	
	\bibitem{MarekT89}
	V.~Marek and M.~Truszczynski.
	\newblock {Stable Semantics for Logic Programs and Default Theories}.
	\newblock In Ewing~L. Lusk and Ross~A. Overbeek, editors, {\em {Proceedings of
			the North American Conference on Logic Programming, NACLP 1989}}, pages
	243--256. {MIT} Press, 1989.
	
	\bibitem{MarekT1993}
	Victor~W. Marek and Miroslaw Truszczynski.
	\newblock {\em {Nonmonotonic Logic: Context-dependent Reasoning}}.
	\newblock Springer, 1993.
	
	\bibitem{MarekT1999}
	Victor~W. Marek and Miroslaw Truszczynski.
	\newblock Stable models and an alternative logic programming paradigm.
	\newblock In Krzysztof~R. Apt, Victor~W. Marek, Mirek Truszczynski, and
	David~Scott Warren, editors, {\em The Logic Programming Paradigm - {A}
		25-Year Perspective}, Artificial Intelligence, pages 375--398. Springer,
	1999.
	
	\bibitem{DBLP:conf/uai/MarraK21}
	Giuseppe Marra and Ondrej Kuzelka.
	\newblock Neural markov logic networks.
	\newblock In Cassio~P. de~Campos, Marloes~H. Maathuis, and Erik Quaeghebeur,
	editors, {\em Proceedings of the Thirty-Seventh Conference on Uncertainty in
		Artificial Intelligence, {UAI} 2021, Virtual Event, 27-30 July 2021}, volume
	161 of {\em Proceedings of Machine Learning Research}, pages 908--917. {AUAI}
	Press, 2021.
	
	\bibitem{McCarthy77}
	J.~McCarthy.
	\newblock Epistemological problems in artificial intelligence.
	\newblock In {\em Proceedings of the International Joint Conference on
		Artificial Intelligence}, pages 1038--1044, Cambridge, MA, 1977.
	
	\bibitem{McCarthy79}
	J.~McCarthy.
	\newblock First order theories of individual concepts and propositions.
	\newblock In D.~Michie, editor, {\em Machine Intelligence 9}, pages 129--147.
	Edinburgh University Press, 1979.
	
	\bibitem{McCarthy80}
	J.~McCarthy.
	\newblock Circumscription -- a form of non-monotonic reasoning.
	\newblock {\em Artificial Intelligence Journal}, 13:27--39, 1980.
	
	\bibitem{McCarthyHayes69}
	J.~McCarthy and P.J. Hayes.
	\newblock Some philosophical problems from the standpoint of artificial
	intelligence.
	\newblock In D.~Michie and B.~Meltzer, editors, {\em Machine Intelligence 4},
	pages 463--502. Edinburgh University Press, 1969.
	
	\bibitem{McCarthy1959}
	John McCarthy.
	\newblock {Programs with Common Sense}, 1959.
	\newblock \url{http://jmc.stanford.edu/articles/mcc59/mcc59.pdf}.
	
	\bibitem{McCarthy1963}
	John McCarthy.
	\newblock Situations, actions, and causal laws.
	\newblock Technical report, Stanford University, Department of Computer
	Science, 1963.
	
	\bibitem{McCHay69}
	John McCarthy and Patrick~J. Hayes.
	\newblock Some philosophical problems from the standpoint of artificial
	intelligence.
	\newblock In B.~Meltzer and D.~Michie, editors, {\em Machine Intelligence 4},
	pages 463--502. Edinburgh University Press, 1969.
	\newblock reprinted in McC90.
	
	\bibitem{McDermott1982}
	Drew McDermott.
	\newblock {Nonmonotonic Logic II: Nonmonotonic Modal Theories}.
	\newblock {\em Journal of the ACM (JACM)}, 29(1):33--57, 1982.
	
	\bibitem{McDermottD1980}
	Drew McDermott and Jon Doyle.
	\newblock {Non-monotonic Logic I}.
	\newblock {\em Artificial intelligence}, 13(1-2):41--72, 1980.
	
	\bibitem{JMcDermott80}
	John~P. McDermott.
	\newblock {R1: an Expert in the Computer Systems Domain}.
	\newblock In Robert Balzer, editor, {\em Proceedings of the 1st Annual National
		Conference on Artificial Intelligence, AAAI 1980}, pages 269--271. {AAAI}
	Press/MIT Press, 1980.
	
	\bibitem{McGuinnessVH2004}
	Deborah~L. McGuinness and Frank~Van Harmelen.
	\newblock {OWL Web Ontology Language Overview}.
	\newblock {\em W3C recommendation}, 10(10):2004, 2004.
	
	\bibitem{DBLP:conf/aaai/MeyerLB05}
	Thomas~Andreas Meyer, Kevin Lee, and Richard Booth.
	\newblock Knowledge integration for description logics.
	\newblock In Manuela~M. Veloso and Subbarao Kambhampati, editors, {\em
		Proceedings, The Twentieth National Conference on Artificial Intelligence and
		the Seventeenth Innovative Applications of Artificial Intelligence
		Conference, July 9-13, 2005, Pittsburgh, Pennsylvania, {USA}}, pages
	645--650. {AAAI} Press / The {MIT} Press, 2005.
	
	\bibitem{Minsky1974}
	Marvin Minsky.
	\newblock {A Framework for Representing Knowledge}.
	\newblock MIT-AI Laboratory Memo 306, June, 1974, Reprinted in \emph{The
		Psychology of Computer Vision}, P. Winston (Ed.), McGraw-Hill, 1975.
	
	\bibitem{MochalesM2011}
	Raquel Mochales and Marie-Francine Moens.
	\newblock Argumentation mining.
	\newblock {\em Artificial Intelligence and Law}, 19:1--22, 2011.
	
	\bibitem{ModgilP2013}
	Sanjay Modgil and Henry Prakken.
	\newblock A general account of argumentation with preferences.
	\newblock {\em Artificial Intelligence}, 195:361--397, 2013.
	
	\bibitem{ModgilP2018}
	Sanjay Modgil and Henry Prakken.
	\newblock Abstract rule-based argumentation.
	\newblock In Pietro Baroni, Dov Gabbay, Massimilino Giacomin, and Leendert
	Van~der Torre, editors, {\em {Handbook of Formal Argumentation}}, pages
	287--364. {College Publications}, 2018.
	
	\bibitem{DBLP:conf/dlog/MoodleyMS14}
	Kodylan Moodley, Thomas Meyer, and Uli Sattler.
	\newblock {DIP:} {A} defeasible-inference platform for {OWL} ontologies.
	\newblock In Meghyn Bienvenu, Magdalena Ortiz, Riccardo Rosati, and Mantas
	Simkus, editors, {\em Informal Proceedings of the 27th International Workshop
		on Description Logics, Vienna, Austria, July 17-20, 2014}, volume 1193 of
	{\em {CEUR} Workshop Proceedings}, pages 671--683. CEUR-WS.org, 2014.
	
	\bibitem{Moore1984}
	Robert~C. Moore.
	\newblock Possible-world semantics for autoepistemic logic.
	\newblock In {\em Proceedings of the Non-Monotonic Reasoning Workshop}, pages
	344--354. American Association for Artificial Intelligence {(AAAI)}, 1984.
	
	\bibitem{Moore1985}
	Robert~C Moore.
	\newblock {Semantical Considerations on Nonmonotonic Logic}.
	\newblock {\em Artificial Intelligence}, 25(1):75--94, 1985.
	
	\bibitem{owl2-profiles}
	Boris Motik, Bernardo {Cuenca Grau}, Ian Horrocks, Zhe Wu, Achille Fokoue, and
	Carsten Lutz, editors.
	\newblock {\em {OWL~2 Web Ontology Language: Profiles}}.
	\newblock W3C Recommendation, 27 October 2009.
	\newblock Available at \url{http://www.w3.org/TR/owl2-profiles/}.
	
	\bibitem{ILP}
	Stephen Muggleton.
	\newblock Inductive logic programming.
	\newblock {\em New Generation Computing}, 8(4):295--318, 1991.
	
	\bibitem{DBLP:journals/jlp/MuggletonR94}
	Stephen~H. Muggleton and Luc~De Raedt.
	\newblock Inductive logic programming: Theory and methods.
	\newblock {\em J. Log. Program.}, 19/20:629--679, 1994.
	
	\bibitem{DBLP:conf/fmcad/Narodytska18}
	Nina Narodytska.
	\newblock Formal verification of deep neural networks.
	\newblock In Nikolaj~S. Bj{\o}rner and Arie Gurfinkel, editors, {\em 2018
		Formal Methods in Computer Aided Design, {FMCAD} 2018, Austin, TX, USA,
		October 30 - November 2, 2018}, page~1. {IEEE}, 2018.
	
	\bibitem{DBLP:conf/aaai/NarodytskaKRSW18}
	Nina Narodytska, Shiva~Prasad Kasiviswanathan, Leonid Ryzhyk, Mooly Sagiv, and
	Toby Walsh.
	\newblock Verifying properties of binarized deep neural networks.
	\newblock In Sheila~A. McIlraith and Kilian~Q. Weinberger, editors, {\em
		{AAAI}}, pages 6615--6624. {AAAI} Press, 2018.
	
	\bibitem{NayakPagnuccoPeppas03}
	A.C. Nayak, M.~Pagnucco, and P.~Peppas.
	\newblock Dynamic belief revision operators.
	\newblock {\em Artificial Intelligence}, 146(2):193--228, 2003.
	
	\bibitem{DBLP:journals/pami/NayyeriXALY23}
	Mojtaba Nayyeri, Chengjin Xu, Mirza~Mohtashim Alam, Jens Lehmann, and
	Hamed~Shariat Yazdi.
	\newblock Logicenn: {A} neural based knowledge graphs embedding model with
	logical rules.
	\newblock {\em {IEEE} Trans. Pattern Anal. Mach. Intell.}, 45(6):7050--7062,
	2023.
	
	\bibitem{DBLP:conf/semweb/NenovPMHWB15}
	Yavor Nenov, Robert Piro, Boris Motik, Ian Horrocks, Zhe Wu, and Jay Banerjee.
	\newblock Rdfox: {A} highly-scalable {RDF} store.
	\newblock In Marcelo Arenas, {\'{O}}scar Corcho, Elena Simperl, Markus
	Strohmaier, Mathieu d'Aquin, Kavitha Srinivas, Paul Groth, Michel Dumontier,
	Jeff Heflin, Krishnaprasad Thirunarayan, and Steffen Staab, editors, {\em The
		Semantic Web - {ISWC} 2015 - 14th International Semantic Web Conference,
		Bethlehem, PA, USA, October 11-15, 2015, Proceedings, Part {II}}, volume 9367
	of {\em Lecture Notes in Computer Science}, pages 3--20. Springer, 2015.
	
	\bibitem{NewellSS1957}
	Allen Newell, J.~C. Shaw, and Herbert~A. Simon.
	\newblock {Empirical Explorations of the Logic Theory Machine: a Case Study in
		Heuristic}.
	\newblock In Morton~M. Astrahan, editor, {\em {Western Joint Computer
			Conference: Techniques for Reliability, {IRE-AIEE-ACM} 1957 (Western)}},
	pages 218--230. {ACM}, 1957.
	
	\bibitem{NewellS1956}
	Allen Newell and Herbert~A. Simon.
	\newblock {The Logic Theory Machine-A Complex Information Processing System}.
	\newblock {\em {IRE} Trans. Inf. Theory}, 2(3):61--79, 1956.
	
	\bibitem{Niemela1999}
	Ilkka Niemel{\"{a}}.
	\newblock Logic programs with stable model semantics as a constraint
	programming paradigm.
	\newblock {\em Ann. Math. Artif. Intell.}, 25(3-4):241--273, 1999.
	
	\bibitem{NiemelaS1997}
	Ilkka Niemel{\"{a}} and Patrik Simons.
	\newblock {Smodels - An Implementation of the Stable Model and Well-Founded
		Semantics for Normal {LP}}.
	\newblock In {\em Proceedings of the 4th International Conference on Logic
		Programming and Nonmotonic Reasoning, {LPNMR 1997}}, volume 1265 of {\em
		Lecture Notes in Computer Science}, pages 421--430. Springer, 1997.
	
	\bibitem{Nilsson1984}
	Nils~J. Nilsson.
	\newblock Shakey the robot, 1984.
	\newblock {SRI International, Menlo Park, California}.
	
	\bibitem{DBLP:journals/ai/Nilsson86}
	Nils~J. Nilsson.
	\newblock Probabilistic logic.
	\newblock {\em Artif. Intell.}, 28(1):71--87, 1986.
	
	\bibitem{NoumanPE21}
	Ahmed Nouman, Volkan Patoglu, and Esra Erdem.
	\newblock Hybrid conditional planning for robotic applications.
	\newblock {\em Int. J. Robotics Res.}, 40(2-3), 2021.
	
	\bibitem{PajunenJ2021}
	Jukka Pajunen and Tomi Janhunen.
	\newblock Solution enumeration by optimality in answer set programming.
	\newblock {\em CoRR}, abs/2108.03474, 2021.
	
	\bibitem{DBLP:conf/rweb/Pareti021}
	Paolo Pareti and George Konstantinidis.
	\newblock A review of {SHACL:} from data validation to schema reasoning for
	{RDF} graphs.
	\newblock In {\em Reasoning Web. Declarative Artificial Intelligence - 17th
		International Summer School 2021, Leuven, Belgium, September 8-15, 2021,
		Tutorial Lectures}, pages 115--144, 2021.
	
	\bibitem{Pearl1985}
	Judea Pearl.
	\newblock {Bayesian networks: A model of self-activated memory for evidential
		reasoning}.
	\newblock In {\em {Proceedings of the 7th Conference of the Cognitive Science
			Society}}, pages 15--17, 1985.
	
	\bibitem{Pearl1988}
	Judea Pearl.
	\newblock {\em Probabilistic Reasoning in Intelligent Systems: Networks of
		Plausible Inference}.
	\newblock Morgan Kaufmann, 1988.
	
	\bibitem{DBLP:series/ssw/Penaloza20}
	Rafael Pe{\~{n}}aloza.
	\newblock Axiom pinpointing.
	\newblock In Giuseppe Cota, Marilena Daquino, and Gian~Luca Pozzato, editors,
	{\em Applications and Practices in Ontology Design, Extraction, and
		Reasoning}, volume~49 of {\em Studies on the Semantic Web}, pages 162--177.
	{IOS} Press, 2020.
	
	\bibitem{Peppas08}
	P.~Peppas.
	\newblock Belief revision.
	\newblock In F.~{van Harmelen}, V.~Lifschitz, and B.~Porter, editors, {\em
		Handbook of Knowledge Representation}, pages 317--359. Elsevier Science, San
	Diego, USA, 2008.
	
	\bibitem{DBLP:conf/emnlp/PetroniRRLBWM19}
	Fabio Petroni, Tim Rockt{\"{a}}schel, Sebastian Riedel, Patrick S.~H. Lewis,
	Anton Bakhtin, Yuxiang Wu, and Alexander~H. Miller.
	\newblock Language models as knowledge bases?
	\newblock In Kentaro Inui, Jing Jiang, Vincent Ng, and Xiaojun Wan, editors,
	{\em Proceedings of the 2019 Conference on Empirical Methods in Natural
		Language Processing and the 9th International Joint Conference on Natural
		Language Processing, {EMNLP-IJCNLP}}, pages 2463--2473. Association for
	Computational Linguistics, 2019.
	
	\bibitem{PETRUCCI201866}
	Giulio Petrucci, Marco Rospocher, and Chiara Ghidini.
	\newblock Expressive ontology learning as neural machine translation.
	\newblock {\em Journal of Web Semantics}, 52-53:66--82, 2018.
	
	\bibitem{PrakkenR2021}
	Henry Prakken and Rosa Ratsma.
	\newblock A top-level model of case-based argumentation for explanation:
	formalisation and experiments.
	\newblock {\em Argument \& Computation}, (Preprint):1--36, 2021.
	
	\bibitem{DBLP:conf/aaai/PratesALLV19}
	Marcelo O.~R. Prates, Pedro H.~C. Avelar, Henrique Lemos, Lu{\'{\i}}s~C. Lamb,
	and Moshe~Y. Vardi.
	\newblock Learning to solve np-complete problems: {A} graph neural network for
	decision {TSP}.
	\newblock In {\em The Thirty-Third {AAAI} Conference on Artificial
		Intelligence, {AAAI}}, pages 4731--4738. {AAAI} Press, 2019.
	
	\bibitem{DBLP:journals/dke/QueraltACT12}
	Anna Queralt, Alessandro Artale, Diego Calvanese, and Ernest Teniente.
	\newblock Ocl-lite: Finite reasoning on {UML/OCL} conceptual schemas.
	\newblock {\em Data Knowl. Eng.}, 73:1--22, 2012.
	
	\bibitem{Quillian1967}
	M~Ross Quillian.
	\newblock {Word Concepts: A Theory and Simulation of Some Basic Semantic
		Capabilities}.
	\newblock {\em Behavioral Science}, 12(5):410--430, 1967.
	
	\bibitem{RagniEtAl2020}
	Marco Ragni, Gabriele Kern{-}Isberner, Christoph Beierle, and Kai Sauerwald.
	\newblock Cognitive logics - features, formalisms, and challenges.
	\newblock In Giuseppe~De Giacomo and J{\'{e}}r{\^{o}}me Lang, editors, {\em
		{ECAI} 2020 - 24th European Conference on Artificial Intelligence}, volume
	325, pages 2931--2932. {IOS} Press, 2020.
	
	\bibitem{RagoCBT2020}
	Antonio Rago, Oana Cocarascu, Christos Bechlivanidis, and Francesca Toni.
	\newblock {Argumentation as a Framework for Interactive Explanations for
		Recommendations}.
	\newblock In {\em {Proceedings of the 17th International Conference on
			Principles of Knowledge Representation and Reasoning}}, pages 805--815, 9
	2020.
	
	\bibitem{DBLP:books/daglib/0030179}
	Manfred Reichert and Barbara Weber.
	\newblock {\em Enabling Flexibility in Process-Aware Information Systems -
		Challenges, Methods, Technologies}.
	\newblock Springer, 2012.
	
	\bibitem{Reiter78}
	R.~Reiter.
	\newblock On closed world data bases.
	\newblock In H.~Gallaire and J.~Minker, editors, {\em Logic and Databases},
	pages 55--76. Plenum Press, 1978.
	
	\bibitem{Reiter80}
	R.~Reiter.
	\newblock A logic for default reasoning.
	\newblock {\em Artificial Intelligence Journal}, 13(1-2):81--132, 1980.
	
	\bibitem{Reiter01}
	R.~Reiter.
	\newblock {\em Knowledge in Action: Logical Foundations for Specifying and
		Implementing Dynamical Systems}.
	\newblock The MIT Press, Cambridge, MA, 2001.
	
	\bibitem{Reiter1978}
	Raymond Reiter.
	\newblock {On Closed World Data Bases}.
	\newblock In Herve Gallaire and Jack Minker, editors, {\em {Logic and Data
			Bases}}, pages 55--76. Plenum Press, 1978.
	
	\bibitem{DBLP:conf/kdd/Ribeiro0G16}
	Marco~T{\'{u}}lio Ribeiro, Sameer Singh, and Carlos Guestrin.
	\newblock "why should {I} trust you?": Explaining the predictions of any
	classifier.
	\newblock In Balaji Krishnapuram, Mohak Shah, Alexander~J. Smola, Charu~C.
	Aggarwal, Dou Shen, and Rajeev Rastogi, editors, {\em Proceedings of the 22nd
		{ACM} {SIGKDD} International Conference on Knowledge Discovery and Data
		Mining, San Francisco, CA, USA, August 13-17, 2016}, pages 1135--1144. {ACM},
	2016.
	
	\bibitem{RizwanPE2020}
	Momina Rizwan, Volkan Patoglu, and Esra Erdem.
	\newblock Human robot collaborative assembly planning: An answer set
	programming approach.
	\newblock {\em Theory Pract. Log. Program.}, 20(6):1006--1020, 2020.
	
	\bibitem{DBLP:conf/aaaifs/RobertsHCCJA16}
	Mark Roberts, Laura~M. Hiatt, Alexandra Coman, Dongkyu Choi, Benjamin Johnson,
	and David~W. Aha.
	\newblock Actorsim, {A} toolkit for studying cross-disciplinary challenges in
	autonomy.
	\newblock In {\em 2016 {AAAI} Fall Symposia, Arlington, Virginia, USA, November
		17-19, 2016}. {AAAI} Press, 2016.
	
	\bibitem{Robinson1965}
	John~Alan Robinson.
	\newblock A machine-oriented logic based on the resolution principle.
	\newblock {\em J. {ACM}}, 12(1):23--41, 1965.
	
	\bibitem{RussellNorvig10}
	Stuart~J. Russell and Peter Norvig.
	\newblock {\em Artificial Intelligence: A Modern Approach}.
	\newblock Pearson Education, third edition, 2010.
	
	\bibitem{DBLP:journals/corr/abs-2210-01240}
	Abulhair Saparov and He~He.
	\newblock Language models are greedy reasoners: {A} systematic formal analysis
	of chain-of-thought.
	\newblock {\em CoRR}, abs/2210.01240, 2022.
	
	\bibitem{SchankA1975}
	Roger~C. Schank and Robert~P. Abelson.
	\newblock {Scripts, Plans and Knowledge}.
	\newblock In {\em The 4th International Joint Conference on Artificial
		Intelligence, IJCAI 1975}, pages 151--157, 1975.
	
	\bibitem{Schild1991}
	Klaus Schild.
	\newblock A correspondence theory for terminological logics: Preliminary
	report.
	\newblock In John Mylopoulos and Raymond Reiter, editors, {\em Proceedings of
		the 12th International Joint Conference on Artificial Intelligence}, pages
	466--471, Sydney, Australia, 1991.
	
	\bibitem{SchneiderGP2013}
	Jodi Schneider, Tudor Groza, and Alexandre Passant.
	\newblock A review of argumentation for the social semantic web.
	\newblock {\em Semantic Web}, 4(2):159--218, 2013.
	
	\bibitem{SchwarzT1994}
	Grigori Schwarz and Miroslaw Truszczynski.
	\newblock Minimal knowledge problem: {A} new approach.
	\newblock {\em Artif. Intell.}, 67(1):113--141, 1994.
	
	\bibitem{ShapiroEtAl11}
	Steven Shapiro, Maurice Pagnucco, Yves Lesp{\'e}rance, and Hector~J. Levesque.
	\newblock Iterated belief change in the situation calculus.
	\newblock {\em Artificial Intelligence}, 175(1):165--192, 2011.
	
	\bibitem{Shoham1987}
	Yoav Shoham.
	\newblock A semantical approach to nonmonotonic logics.
	\newblock In {\em Proceedings of Logics in Computer Science}, pages 275--279,
	Ithaca, New York, 1987.
	
	\bibitem{Shortliffe1976}
	Edward Shortliffe.
	\newblock {\em {Computer-based Medical Consultations: MYCIN}}.
	\newblock Addison-Wesley, 1976.
	
	\bibitem{SPGK06a}
	Evren Sirin, Bijan Parsia, Bernardo Cuenca~Grau, Aditya Kalyanpur, and Yarden
	Katz.
	\newblock Pellet: A practical {OWL-DL} reasoner.
	\newblock {\em Journal of Web Semantics}, 5(2):51--53, 2007.
	
	\bibitem{Smith82}
	B.C. Smith.
	\newblock {\em Reflections and Semantics in a Procedural Language}.
	\newblock PhD thesis, MIT, 1982.
	
	\bibitem{SpiegelEtAl2019}
	Lars{-}Phillip Spiegel, Gabriele Kern{-}Isberner, and Marco Ragni.
	\newblock Rational inference patterns.
	\newblock In Abhaya~C. Nayak and Alok Sharma, editors, {\em {PRICAI} 2019: 16th
		Pacific Rim International Conference on Artificial Intelligence}, volume
	11670 of {\em Lecture Notes in Computer Science}, pages 405--417. Springer,
	2019.
	
	\bibitem{Spohn88}
	W.~Spohn.
	\newblock Ordinal conditional functions: A dynamic theory of epistemic states.
	\newblock In W.L. Harper and B.~Skyrms, editors, {\em Causation in Decision,
		Belief Change, and Statistics}, volume~II, pages 105--134. Kluwer Academic
	Publishers, 1988.
	
	\bibitem{DBLP:journals/ws/SteigmillerLG14}
	Andreas Steigmiller, Thorsten Liebig, and Birte Glimm.
	\newblock Konclude: System description.
	\newblock {\em J. Web Sem.}, 27:78--85, 2014.
	
	\bibitem{DBLP:journals/kais/StrumbeljK14}
	Erik Strumbelj and Igor Kononenko.
	\newblock Explaining prediction models and individual predictions with feature
	contributions.
	\newblock {\em Knowl. Inf. Syst.}, 41(3):647--665, 2014.
	
	\bibitem{DBLP:series/lncs/5445}
	Heiner Stuckenschmidt, Christine Parent, and Stefano Spaccapietra, editors.
	\newblock {\em Modular Ontologies: Concepts, Theories and Techniques for
		Knowledge Modularization}, volume 5445 of {\em Lecture Notes in Computer
		Science}.
	\newblock Springer, 2009.
	
	\bibitem{STUDER1998161}
	Rudi Studer, V.Richard Benjamins, and Dieter Fensel.
	\newblock Knowledge engineering: Principles and methods.
	\newblock {\em Data \& Knowledge Engineering}, 25(1):161--197, 1998.
	
	\bibitem{DBLP:conf/iclr/SunDNT19}
	Zhiqing Sun, Zhi{-}Hong Deng, Jian{-}Yun Nie, and Jian Tang.
	\newblock Rotate: Knowledge graph embedding by relational rotation in complex
	space.
	\newblock In {\em 7th International Conference on Learning Representations,
		{ICLR} 2019, New Orleans, LA, USA, May 6-9, 2019}. OpenReview.net, 2019.
	
	\bibitem{SyrjanenN2001}
	Tommi Syrj{\"a}nen and Ilkka Niemel{\"a}.
	\newblock The smodels system.
	\newblock In {\em Proceedings of the 6th International Conference on Logic
		Programming and Nonmotonic Reasoning, {LPNMR 2001}}, volume 2173 of {\em
		Lecture Notes in Computer Science}, pages 434--438. Springer, 2001.
	
	\bibitem{BeetzTenorth17}
	Moritz Tenorth and Michael Beetz.
	\newblock Representations for robot knowledge in the knowrob framework.
	\newblock {\em Artif. Intell.}, 247:151--169, 2017.
	
	\bibitem{TeufelSB2009}
	Simone Teufel, Advaith Siddharthan, and Colin Batchelor.
	\newblock Towards domain-independent argumentative zoning: Evidence from
	chemistry and computational linguistics.
	\newblock In {\em {Proceedings of the 2009 Conference on Empirical Methods in
			Natural Language Processing}}, pages 1493--1502, 2009.
	
	\bibitem{Thielscher97}
	Michael Thielscher.
	\newblock Ramification and causality.
	\newblock {\em Artificial Intelligence Journal}, 89(1-2):317--364, 1997.
	
	\bibitem{Minerva00}
	Sebastian Thrun, Michael Beetz, Maren Bennewitz, Wolfram Burgard, Armin~B.
	Cremers, Frank Dellaert, Dieter Fox, Dirk H{\"{a}}hnel, Charles~R. Rosenberg,
	Nicholas Roy, Jamieson Schulte, and Dirk Schulz.
	\newblock Probabilistic algorithms and the interactive museum tour-guide robot
	minerva.
	\newblock {\em Int. J. Robotics Res.}, 19(11):972--999, 2000.
	
	\bibitem{ThrunBF05}
	Sebastian Thrun, Wolfram Burgard, and Dieter Fox.
	\newblock {\em Probabilistic robotics}.
	\newblock Intelligent robotics and autonomous agents. {MIT} Press, 2005.
	
	\bibitem{Touretzky1986}
	David~S. Touretzky.
	\newblock {\em {The Mathematics of Inheritance Systems}}.
	\newblock Morgan Kaufmann, 1986.
	
	\bibitem{DBLP:journals/jmlr/TrouillonDGWRB17}
	Th{\'{e}}o Trouillon, Christopher~R. Dance, {\'{E}}ric Gaussier, Johannes
	Welbl, Sebastian Riedel, and Guillaume Bouchard.
	\newblock Knowledge graph completion via complex tensor factorization.
	\newblock {\em J. Mach. Learn. Res.}, 18:130:1--130:38, 2017.
	
	\bibitem{10.2307/4321713}
	Johan van Benthem.
	\newblock Epistemic logic and epistemology: The state of their affairs.
	\newblock {\em Philosophical Studies: An International Journal for Philosophy
		in the Analytic Tradition}, 128(1):49--76, 2006.
	
	\bibitem{DBLP:books/sp/Aalst16}
	Wil M.~P. van~der Aalst.
	\newblock {\em Process Mining - Data Science in Action, Second Edition}.
	\newblock Springer, 2016.
	
	\bibitem{DBLP:reference/fai/3}
	Frank van Harmelen, Vladimir Lifschitz, and Bruce~W. Porter, editors.
	\newblock {\em Handbook of Knowledge Representation}, volume~3 of {\em
		Foundations of Artificial Intelligence}.
	\newblock Elsevier, 2008.
	
	\bibitem{vanHeijenoort1967}
	Jean van Heijenoort.
	\newblock {\em From Frege to G\"{o}del: A Source Book in Mathematical Logic,
		1879-1931}.
	\newblock Cambridge, MA, USA: Harvard University Press, 1967.
	
	\bibitem{VanbesienBD2022}
	Linde Vanbesien, Maurice Bruynooghe, and Marc Denecker.
	\newblock Analyzing semantics of aggregate answer set programming using
	approximation fixpoint theory.
	\newblock {\em Theory Pract. Log. Program.}, 22(4):523--537, 2022.
	
	\bibitem{VillataC2018}
	Serena Villata and Elena Cabrio.
	\newblock Five years of argument mining: a data-driven analysis.
	\newblock In {\em {Proceedings of the 27th International Joint Conference on
			Artificial Intelligence, IJCAI 2018}}. International Joint Conferences on
	Artificial Intelligence Organization, 2018.
	
	\bibitem{10.1145/2629489}
	Denny Vrande\v{c}i\'{c} and Markus Kr\"{o}tzsch.
	\newblock Wikidata: A free collaborative knowledgebase.
	\newblock {\em Commun. ACM}, 57(10):78–85, sep 2014.
	
	\bibitem{owl2}
	{W3C OWL Working Group}.
	\newblock {OWL} 2 web ontology language document overview.
	\newblock URL, 2009.
	\newblock http://www.w3.org/TR/owl2-overview/.
	
	\bibitem{DBLP:journals/jair/WalegaZG23}
	Przemyslaw~Andrzej Walega, Michal Zawidzki, and Bernardo~Cuenca Grau.
	\newblock Finite materialisability of datalog programs with metric temporal
	operators.
	\newblock {\em J. Artif. Intell. Res.}, 76, 2023.
	
	\bibitem{Wang2017KnowledgeGE}
	Quan Wang, Zhendong Mao, Bin Wang, and Li~Guo.
	\newblock Knowledge graph embedding: A survey of approaches and applications.
	\newblock {\em IEEE Transactions on Knowledge and Data Engineering},
	29:2724--2743, 2017.
	
	\bibitem{DBLP:journals/corr/abs-2201-11903}
	Jason Wei, Xuezhi Wang, Dale Schuurmans, Maarten Bosma, Ed~H. Chi, Quoc Le, and
	Denny Zhou.
	\newblock Chain of thought prompting elicits reasoning in large language
	models.
	\newblock {\em CoRR}, abs/2201.11903, 2022.
	
	\bibitem{WeinzierlTF2020}
	Antonius Weinzierl, Richard Taupe, and Gerhard Friedrich.
	\newblock Advancing lazy-grounding {ASP} solving techniques - restarts, phase
	saving, heuristics, and more.
	\newblock {\em Theory Pract. Log. Program.}, 20(5):609--624, 2020.
	
	\bibitem{WhiteheadRussell1927}
	Alfred~North Whitehead and Bertrand Arthur~William Russell.
	\newblock {\em {Principia {M}athematica; 2nd ed.}}
	\newblock Cambridge Univ. Press, Cambridge, 1927.
	
	\bibitem{DBLP:conf/nips/WiegreffeM21}
	Sarah Wiegreffe and Ana Marasovic.
	\newblock Teach me to explain: {A} review of datasets for explainable natural
	language processing.
	\newblock In Joaquin Vanschoren and Sai{-}Kit Yeung, editors, {\em Proceedings
		of the Neural Information Processing Systems Track on Datasets and Benchmarks
		1, NeurIPS Datasets and Benchmarks 2021, December 2021, virtual}, 2021.
	
	\bibitem{KR2022-51}
	Hong Wu, Zhe Wang, Kewen Wang, and Yi-Dong Shen.
	\newblock {Learning Typed Rules over Knowledge Graphs}.
	\newblock In {\em {Proceedings of the 19th International Conference on
			Principles of Knowledge Representation and Reasoning}}, pages 494--503, 8
	2022.
	
	\bibitem{DBLP:conf/ijcai/XiaoCKLPRZ18}
	Guohui Xiao, Diego Calvanese, Roman Kontchakov, Domenico Lembo, Antonella
	Poggi, Riccardo Rosati, and Michael Zakharyaschev.
	\newblock Ontology-based data access: {A} survey.
	\newblock In J{\'{e}}r{\^{o}}me Lang, editor, {\em Proceedings of the
		Twenty-Seventh International Joint Conference on Artificial Intelligence,
		{IJCAI} 2018, July 13-19, 2018, Stockholm, Sweden}, pages 5511--5519.
	ijcai.org, 2018.
	
	\bibitem{DBLP:conf/semweb/XiaoLKKKDCCCB20}
	Guohui Xiao, Davide Lanti, Roman Kontchakov, Sarah Komla{-}Ebri,
	Elem~G{\"{u}}zel Kalayci, Linfang Ding, Julien Corman, Benjamin Cogrel, Diego
	Calvanese, and Elena Botoeva.
	\newblock The virtual knowledge graph system ontop.
	\newblock In Jeff~Z. Pan, Valentina A.~M. Tamma, Claudia d'Amato, Krzysztof
	Janowicz, Bo~Fu, Axel Polleres, Oshani Seneviratne, and Lalana Kagal,
	editors, {\em The Semantic Web - {ISWC} 2020 - 19th International Semantic
		Web Conference, Athens, Greece, November 2-6, 2020, Proceedings, Part {II}},
	volume 12507 of {\em Lecture Notes in Computer Science}, pages 259--277.
	Springer, 2020.
	
	\bibitem{DBLP:conf/icml/XuZFLB18}
	Jingyi Xu, Zilu Zhang, Tal Friedman, Yitao Liang, and Guy~Van den Broeck.
	\newblock A semantic loss function for deep learning with symbolic knowledge.
	\newblock In Jennifer~G. Dy and Andreas Krause, editors, {\em Proceedings of
		the 35th International Conference on Machine Learning, {ICML} 2018,
		Stockholmsm{\"{a}}ssan, Stockholm, Sweden, July 10-15, 2018}, volume~80 of
	{\em Proceedings of Machine Learning Research}, pages 5498--5507. {PMLR},
	2018.
	
	\bibitem{DBLP:journals/corr/YangYHGD14a}
	Bishan Yang, Wen{-}tau Yih, Xiaodong He, Jianfeng Gao, and Li~Deng.
	\newblock Embedding entities and relations for learning and inference in
	knowledge bases.
	\newblock In Yoshua Bengio and Yann LeCun, editors, {\em 3rd International
		Conference on Learning Representations, {ICLR} 2015, San Diego, CA, USA, May
		7-9, 2015, Conference Track Proceedings}, 2015.
	
	\bibitem{YangIL2020}
	Zhun Yang, Adam Ishay, and Joohyung Lee.
	\newblock Neur{ASP}: Embracing neural networks into answer set programming.
	\newblock In Christian Bessiere, editor, {\em Proceedings of the Twenty-Ninth
		International Joint Conference on Artificial Intelligence, {IJCAI}}, pages
	1755--1762, 2020.
	
	\bibitem{ZamanSMLU13}
	Safdar Zaman, Gerald Steinbauer, Johannes Maurer, Peter Lepej, and Suzana Uran.
	\newblock An integrated model-based diagnosis and repair architecture for
	ros-based robot systems.
	\newblock In {\em 2013 {IEEE} International Conference on Robotics and
		Automation, Karlsruhe, Germany, May 6-10, 2013}, pages 482--489, 2013.
	
	\bibitem{DBLP:conf/kr/Zarriess18}
	Benjamin Zarrie{\ss}.
	\newblock Complexity of projection with stochastic actions in a probabilistic
	description logic.
	\newblock In Michael Thielscher, Francesca Toni, and Frank Wolter, editors,
	{\em Principles of Knowledge Representation and Reasoning: Proceedings of the
		Sixteenth International Conference, {KR} 2018, Tempe, Arizona, 30 October - 2
		November 2018}, pages 514--523. {AAAI} Press, 2018.
	
	\bibitem{DBLP:conf/cvpr/ZellersLLYZSKHF22}
	Rowan Zellers, Jiasen Lu, Ximing Lu, Youngjae Yu, Yanpeng Zhao, Mohammadreza
	Salehi, Aditya Kusupati, Jack Hessel, Ali Farhadi, and Yejin Choi.
	\newblock {MERLOT} {RESERVE:} neural script knowledge through vision and
	language and sound.
	\newblock In {\em {IEEE/CVF} Conference on Computer Vision and Pattern
		Recognition, {CVPR} 2022, New Orleans, LA, USA, June 18-24, 2022}, pages
	16354--16366. {IEEE}, 2022.
	
	\bibitem{ZellnerABG2021}
	Maximilian Zellner, Ali~E Abbas, David~V Budescu, and Aram Galstyan.
	\newblock A survey of human judgement and quantitative forecasting methods.
	\newblock {\em Royal Society Open Science}, 8(2):201187, 2021.
	
	\bibitem{DBLP:journals/corr/abs-2205-11502}
	Honghua Zhang, Liunian~Harold Li, Tao Meng, Kai{-}Wei Chang, and Guy {Van den
		Broeck}.
	\newblock On the paradox of learning to reason from data.
	\newblock {\em CoRR}, abs/2205.11502, 2022.
	
	\bibitem{DBLP:conf/cade/ZhaoS18}
	Yizheng Zhao and Renate~A. Schmidt.
	\newblock {FAME:} an automated tool for semantic forgetting in expressive
	description logics.
	\newblock In Didier Galmiche, Stephan Schulz, and Roberto Sebastiani, editors,
	{\em Automated Reasoning - 9th International Joint Conference, {IJCAR}},
	volume 10900 of {\em Lecture Notes in Computer Science}, pages 19--27.
	Springer, 2018.
	
	\bibitem{ZhuangWangWangQi16}
	Zhiqiang Zhuang, Zhe Wang, Kewen Wang, and Guilin Qi.
	\newblock {DL}-{L}ite contraction and revision.
	\newblock {\em Journal of Artificial Intelligence Research}, 56:328--378, 2016.
	
\end{thebibliography}

\end{document}